\documentclass[11pt,draftcls,onecolumn]{IEEEtran}
\usepackage{amsmath, amsthm, amssymb}
\usepackage{latexsym}
\usepackage{exscale}
\usepackage{fontenc}
\usepackage{graphicx}
\usepackage{epsfig}
\usepackage{nicefrac}
\usepackage{mathptmx}
\usepackage{xcolor,epic,eepic}
\usepackage[ruled,lined,linesnumbered]{algorithm2e}
\usepackage{subfigure}
\usepackage{fancyhdr}
\usepackage{hyperref}
\usepackage{multirow}
\usepackage{setspace}
\usepackage{authblk}

\begin{document}
\renewcommand{\baselinestretch}{1.1}
\newcommand{\nin}{\in\!\!\!/}
\newcommand{\rb}{\rangle}
\newcommand{\lb}{\langle}
\newcommand{\R}{{\bf R}}
\newcommand{\thTh}{{\theta \in \Theta}}
\newcommand{\Dip}{{D}}
\newcommand{\cG}{{\cal G}}
\newcommand{\cQ}{{\cal Q}}
\newcommand{\curvelet}{{c}}
\newcommand{\orient}{{\theta}}
\newcommand{\ERC}{{\rm ERC}}
\newcommand{\DU}{{{\cal D}_U}}
\newcommand{\ipidx}{{p}}
\newcommand{\ipatom}{{g}}
\newcommand{\Ga}{{\Gamma}}
\newcommand{\pGa}{{p \in \Gamma}}
\newcommand{\lu}{{\bf l}^1}
\newcommand{\Ld}{{\bf L}^2}
\newcommand{\lz}{{\bf l}^0}
\newcommand{\ld}{{\bf l}^2}
\newcommand{\V}{{\bf V}}
\newcommand{\cL}{{\cal L}}
\newcommand{\W}{{\bf W}}
\newcommand{\La}{{\Lambda}}
\newcommand{\tLa}{{\tilde \Lambda}}
\newcommand{\pLa}{{p \in \Lambda}}
\newcommand{\qO}{{q \in \Omega}}
\newcommand{\ptLa}{{p \in \tilde \Lambda}}
\newcommand{\supp}{{\Lambda}}
\newcommand{\om}{{\omega}}
\newcommand{\mq}{{m,q}}
\newcommand{\jn}{{j,n}}
\newcommand{\opt}{{\tilde}}
\newcommand{\IPop}{{U}}
\newcommand{\CC}{{\bf C}}
\newcommand{\Z}{{\bf Z}}
\newcommand{\Reg}{\Lambda}
\newtheorem{definition}{Definition}
\newtheorem{theorem}{Theorem}
\newtheorem{lemma}{Lemma}
\newtheorem{corollary}{Corollary}
\newtheorem{prop}{Proposition}
\providecommand{\argmin}{\mathop{\textup{argmin}}}
\def\truelabel{\label}
\newcommand{\Span}{\mathop{\textup{span}}}
\newcommand{\pen}{\text{pen}}
\newcommand{\ud}{\textup{d}}
\newcommand{\D}{\mathcal{D}}
\newcommand{\Mg}{\mathcal{M}_{\gamma}}
\newcommand{\Tde}{T^{\frac{2\alpha}{\alpha+1}}}
\newcommand{\B}{\mathcal{B}}
\newcommand{\wor}{k}
\newcommand{\Cal}{\mathbf{C}^{\alpha}}
\newcommand{\eqdef}{\overset{.def}{=}}
\newcommand{\tga}{\tilde g}
\newcommand{\ldeuxj}{l^2_j}
\newcommand{\ga}{g}
\newcommand{\nogeom}{\Xi}
\newcommand{\tS}{\tilde S}
\newcommand{\tb}{\tilde b}
\newcommand{\ldeux}{l^2}
\newcommand {\ImU} {{\bf ImU}}
\newcommand{\Proba}{\mathbb{P}}
\newcommand{\C} {{\bf C}}
\newcommand{\gammageom}{\upsilon}
\newcommand{\Ch}[1]{{\bf Ch: #1}}

\newcommand{\ba}{\mathbf{a}}
\newcommand{\bbf}{\mathbf{f}}
\newcommand{\by}{\mathbf{y}}
\newcommand{\bw}{\mathbf{w}}
\newcommand{\bz}{\mathbf{z}}

\newcommand{\bB}{\mathbf{B}}
\newcommand{\bD}{\mathbf{D}}
\newcommand{\bI}{\mathbf{I}}
\newcommand{\bM}{\mathbf{M}}
\newcommand{\bS}{\mathbf{S}}
\newcommand{\bT}{\mathbf{T}}
\newcommand{\bU}{\mathbf{U}}
\newcommand{\bW}{\mathbf{W}}

\linespread{0.97}

\title{Solving Inverse Problems with Piecewise Linear Estimators: From Gaussian Mixture Models to Structured Sparsity}
\author[1]{Guoshen \textsc{Yu}*}
\author[1]{Guillermo \textsc{Sapiro}}
\author[2]{St\'ephane \textsc{Mallat}}
\affil[1]{ECE, University of Minnesota, Minneapolis, Minnesota, 55414, USA}
\affil[2]{CMAP, Ecole Polytechnique,  91128 Palaiseau Cedex, France}

\maketitle

\begin{center}
\vspace{-8ex} 
\textit{EDICS: TEC-RST} \\
{\textbf{\today}} \vspace{3ex}
\end{center}

\begin{abstract}
A general framework for solving image inverse problems is introduced in this paper. The approach is based on Gaussian mixture models, estimated via a computationally efficient MAP-EM algorithm. A dual mathematical interpretation of the proposed framework with structured sparse estimation is described, which shows that the resulting piecewise linear estimate stabilizes the estimation when compared to traditional sparse inverse problem techniques. This interpretation also suggests an effective dictionary motivated initialization for the MAP-EM algorithm. We demonstrate that in a number of image inverse problems, including inpainting, zooming, and deblurring, the same algorithm produces either equal, often significantly better, or very small margin worse results than the best published ones, at a lower computational cost. 
\end{abstract}

\section{Introduction}
\label{sec:intro}

Image restoration often requires to solve an inverse problem. It amounts to estimate an image $\bbf$ from a measurement 
$$
\by = \bU \bbf + \bw,
$$
obtained through a non-invertible linear degradation operator $\bU$, and contaminated by an additive noise $\bw$. Typical degradation operators include masking, subsampling in a uniform grid and convolution, the corresponding inverse problems often named inpainting or interpolation, zooming and deblurring. Estimating $\bbf$ requires some prior information on the image, or equivalently image models. Finding good image models is therefore at the heart of image estimation. 

Mixture models are often used as image priors since they enjoy the flexibility of signal description by assuming that the signals are generated by a mixture of probability distributions~\cite{mclachlan1988mixture}. Gaussian mixture models (GMM) have been shown to provide powerful tools for data classification and segmentation applications (see for example~\cite{celeux1995gaussian, friedman1997image, permuter2003gaussian, stauffer1999adaptive}), however, they have not yet been shown to generate state-of-the-art in a general class of inverse problems. Ghahramani and Jordan have applied GMM for learning from incomplete data, i.e., images degraded by a masking operator, and have shown good classification results, however, it does not lead to state-of-the-art inpainting~\cite{ghahramani1994supervised}. Portilla et al. have shown image denoising impressive results by assuming Gaussian scale mixture models (deviating from GMM by assuming different scale factors in the mixture of Gaussians) on wavelet representations~\cite{portilla2003image}, and have recently extended its applications on image deblurring~\cite{guerrero2008image}. Recently, Zhou et al. have developed an nonparametric Bayesian approach using more elaborated models, such as beta and Dirichlet processes, which leads to excellent results in denoising and inpainting~\cite{zhou2010nonparametric}. 

The now popular sparse signal models, on the other hand, assume that the signals can be accurately represented with a few coefficients selecting atoms in some dictionary~\cite{mallat2008wts}.  Recently, very impressive image restoration results have been obtained with local patch-based sparse representations calculated with dictionaries learned from natural images~\cite{aharon2006k, elad2006image, mairal2008sparse}.  Relative to pre-fixed dictionaries such as wavelets~\cite{mallat2008wts},  curvelets~\cite{candes2004new}, and bandlets~\cite{mallat2008orthogonal}, learned dictionaries enjoy the advantage of being better adapted to the images, thereby enhancing the sparsity. However, dictionary learning is a large-scale and highly non-convex problem. It requires high computational complexity, and its mathematical behavior is not yet well understood. In the dictionaries aforementioned,  the actual sparse image representation is calculated with relatively expensive non-linear estimations, such as $l_1$ or matching pursuits~\cite{daubechies2004iterative, efron2004least,mallat1993matching}. More importantly, as will be reviewed in Section~\ref{sec:sparse:l1}, with a full degree of freedom in selecting the approximation space (atoms of the dictionary), non-linear sparse inverse problem estimation may be unstable and imprecise due to the coherence of the dictionary~\cite{mallat10SME}. 

Structured sparse image representation models further regularize the sparse 
estimation by assuming dependency on the selection of the active atoms. One simultaneously selects  blocks of approximation
atoms, thereby reducing the number of possible approximation spaces~\cite{baraniuk2008model, eldar2009compressed,eldar2008robust,huang2009structure,jenatton2009structured,Stojnic2009block}. 
These structured approximations
have been shown to improve the signal estimation in a compressive
sensing context for a random operator
$\bU$. However, for more unstable inverse problems such as zooming or deblurring,
 this regularization by itself is not sufficient to reach state-of-the-art results. 
 Recently some good image zooming results have been obtained with
 structured sparsity based on directional block structures in wavelet representations~\cite{mallat10SME}.
 However, this directional regularization is not general enough to be extended to solve other inverse problems.
 
This work shows that the Gaussian mixture models (GMM), estimated via an MAP-EM (maximum a posteriori expectation-maximization) algorithm, lead to results in the same ballpark as the state-of-the-art in a number of imaging inverse problems, at a lower computational cost. The MAP-EM algorithm is described in Section~\ref{sec:SSMS}. After briefly reviewing sparse inverse problem estimation approaches, a mathematical equivalence between the proposed piecewise linear estimation (PLE) from GMM/MAP-EM and structured sparse estimation is shown in Section~\ref{sec:MAP:EM:sparsity}. This connection shows that PLE stabilizes the sparse estimation with a structured learned overcomplete dictionary composed of a union of PCA (Principal Component Analysis) bases, and with collaborative prior information incorporated in the eigenvalues, that privileges in the estimation the atoms that are more likely to be important. This interpretation suggests also an effective dictionary motivated initialization for the MAP-EM algorithm. In Section~\ref{sec:initial:exp} we support the importance of different components of the proposed PLE via some initial experiments. Applications of the proposed PLE in image inpainting, zooming, and deblurring are presented in sections~\ref{sec:inpainting},~\ref{sec:zooming}, and~\ref{sec:deblurring} respectively, and are compared with previous state-of-the-art methods. Conclusions are drawn in Section~\ref{sec:conclusion}. 

\section{Piecewise Linear Estimation}
\label{sec:SSMS}

This section describes the Gaussian mixture models (GMM) and the MAP-EM algorithm, which lead to the proposed piecewise linear estimation (PLE). 

\subsection{Gaussian Mixture Model}
\label{subsec:GMM}

Natural images include rich and non-stationary content, whereas when restricted to local windows, image structures appear to be simpler and are therefore easier to model. Following some previous works~\cite{aharon2006k, buades2006review, mairal2008sparse}, an image is decomposed into overlapping $\sqrt{N} \times \sqrt{N}$ local patches 
\begin{equation}
\label{eqn:inverse:problem:patch}
\by_i = \bU_i \bbf_i + \bw_i,
\end{equation}
where $\bU_i$ is the degradation operator restricted to the patch $i$, $\by_i$, $\bbf_i$ and $\bw_i$ are respectively the degraded, original image patches and the noise restricted to the patch, with $1 \leq i \leq I$, $I$ being the total number of patches. Treated as a signal, each of the patches is estimated, and their corresponding estimates are finally combined and averaged, leading to the estimate of the image. 

GMM describes local image patches with a mixture of Gaussian distributions. Assume there exist $K$ Gaussian distributions $\{\mathcal{N} (\mu_k, \Sigma_k)\}_{1 \leq k \leq K}$ parametrized by their means $\mu_k$ and covariances $\Sigma_k$. Each image patch $\bbf_i$ is independently drawn from one of these Gaussians with an unknown index $k$, whose probability density function is
\begin{equation}
\label{eqn:multivariate:gaussian}
p(\bbf_i) = \frac{1}{(2\pi)^{N/2} |\Sigma_{k_i}|^{1/2}} \exp\left({-\frac{1}{2} (\bbf_i - \mu_k)^T \Sigma_{k_i}^{-1} (\bbf_i - \mu_k)}\right).
\end{equation}
Estimating $\{\bbf_i\}_{1 \leq i \leq I}$ from $\{\by_i\}_{1 \leq i \leq I}$ can then be casted into the following problems:
\begin{itemize}
\item Estimate the Gaussian parameters $\{(\mu_k, \Sigma_k)\}_{1 \leq k \leq K}$, from the degraded data $\{\by_i\}_{1 \leq i \leq I}$.
\item Identify the Gaussian distribution $k_i$ that generates the patch $i$, $\forall 1 \leq i \leq I$.
\item Estimate $\bbf_i$ from its corresponding Gaussian distribution $(\mu_{k_i}, \Sigma_{k_i})$, $\forall 1 \leq i \leq I$.
\end{itemize}
These problems are overall non-convex. The next section will present a maximum a posteriori expectation-maximization (MAP-EM) algorithm that calculates a local-minimum solution~\cite{allassonniere2007towards}. 

\subsection{MAP-EM Algorithm}
Following an initialization, addressed in Section~\ref{subsec:init}, the MAP-EM algorithm is an iterative procedure that alternates between two steps: 
\begin{itemize}
\item In the E-step, assuming that the estimates of the Gaussian parameters $\{(\tilde{\mu}_k, \tilde{\Sigma}_k)\}_{1 \leq k \leq K}$ are known (following the previous M-step), for each patch  one calculates the maximum a posteriori (MAP) estimates $\tilde{\bbf}_i^k$ with all the Gaussian models, and selects the best Gaussian model $k_i$ to obtain the estimate of the patch $\tilde{\bbf}_i = \tilde{\bbf}_i^{k_i}$.
\item In the M-step, assuming that the Gaussian model selection $k_i$ and the signal estimate $\tilde{\bbf}_i$, $\forall i$, are known (following the previous E-step), one estimates (updates) the Gaussian models $\{(\tilde{\mu}_k, \tilde{\Sigma}_k)\}_{1 \leq k \leq K}$. 
\end{itemize} 

\subsubsection{E-step: Signal Estimation and Model Selection}
In the E-step, the estimates of the Gaussian parameters $\{(\tilde{\mu}_k, \tilde{\Sigma}_k)\}_{1 \leq k \leq K}$ are assumed to be known. To simplify the notation, we assume without loss of generality that the Gaussians have zero means $\tilde{\mu}_k = \mathbf{0}$, as one can always center the image patches  with respect to the means. 

For each image patch $i$, the signal estimation and model selection is calculated to maximize the log a-posteriori probability $\log p(\bbf_i | \by_i, \tilde{\Sigma}_k)$:
\begin{eqnarray}
\label{eqn:MAP:model:signal}
(\tilde{\bbf}_i, k_i) & = & \arg \max_{\bbf, k} \log p(\bbf_i | \by_i, \tilde{\Sigma}_k) 
=  \arg \max_{\bbf, k} \left( \log p(\by_i|\bbf_i, \tilde{\Sigma}_k) +  \log p(\bbf_i|\tilde{\Sigma}_k) \right)\nonumber \\
& = & \arg \min_{\bbf_i, k} \left( \|\bU_i \bbf_i - \by_i\|^2 + \sigma^2 \bbf_i^T \tilde{\Sigma}_k^{-1} \bbf_i + \sigma^2 \log \left| \tilde{\Sigma}_k \right| \right),
\end{eqnarray}
where the second equality follows the Bayes rule and the third one is derived with the assumption that $\bw_i \sim \mathcal{N}(\mathbf{0}, \sigma^2 Id)$, with $Id$ the identity matrix, and $\bbf_i \sim \mathcal{N}(\mathbf{0}, \tilde{\Sigma}_k)$. 

The maximization is first calculated over $\bbf_i$ and then over $k$. Given a Gaussian signal model $\bbf_i \sim \mathcal{N}(\mathbf{0}, \tilde{\Sigma}_k)$, it is well known that the MAP estimate 
\begin{equation}
\label{eqn:MAP:estimate:gaussian}
\tilde{\bbf}_i^k= \arg \min_{{\bbf}_i}  \left( \|\bU_i {\bbf}_i - \by_i\|^2 + \sigma^2 \bbf_i^T \tilde{\Sigma}_{k}^{-1} \bbf_i \right)
\end{equation}
minimizes the risk $E[\|\tilde{\bbf}_i^k - \bbf_i\|^2]$~\cite{mallat2008wts}. One can verify that the solution to~\eqref{eqn:MAP:estimate:gaussian} can be calculated with a linear filtering 
\begin{equation}
\label{eqn:MAP:estimate:gaussian:solution}
\tilde{\bbf}_i^k = \bW_{{k_i}} \by_i,
\end{equation}
where
\begin{equation}
\label{eqn:MAP:wiener}
\bW_{k_i} = (\bU_i^T \bU_i + \sigma^2 \Sigma_{k_i}^{-1} )^{-1} \bU_i^T
\end{equation}
is a Wiener filter matrix.  Since $\bU_i^T \bU_i$ is semi-positive definite, $\bU_i^T \bU_i + \sigma^2 \Sigma_{k_i}^{-1}$ is positive definite and its inverse is well defined, if $\Sigma_k$ is full rank.

The best Gaussian model $k_i$ that generates the maximum MAP probability among all the models is then selected with the estimated $\tilde{\bbf}_i^k$
\begin{equation}
\label{eqn:MAP:model:selection}
k_i = \arg \min_{k} \left( \|\bU_i \tilde{\bbf}_i^k - \by\|^2 + \sigma^2 (\tilde{\bbf}_i^k)^T\Sigma_k^{-1} \tilde{\bbf}_i^k+  \sigma^2 \log \left| \tilde{\Sigma}_k \right| \right).
\end{equation}
The signal estimate is obtained by plugging in the best model $k_i$ in the MAP estimate~\eqref{eqn:MAP:estimate:gaussian}
\begin{equation}
\label{eqn:MAP:estimate:gaussian1:best}
\tilde{\bbf}_i= \tilde{\bbf}_i^{k_i}. \end{equation}

The whole E-step is basically calculated with a set of linear filters. For typical applications such as zooming and deblurring where the degradation operators $\bU_i$ are translation-invariant and do not depend on the patch index $i$, i.e., $\bU_i \equiv \bU$, the Wiener filter matrices $\bW_{k_i} \equiv \bW_k$~\eqref{eqn:MAP:wiener} can be precomputed for the $K$ Gaussian distributions. Calculating~\eqref{eqn:MAP:estimate:gaussian:solution}  thus requires only $2N^2$ floating-point operations (flops), where $N$ is the image patch size. For a translation-variant degradation $\bU_i$, random masking for example, $\bW_{k_i}$ needs to be calculated at each position where $\bU_i$ changes. Since $\bU_i^T \bU_i + \sigma^2 \Sigma_{k_i}^{-1}$ is positive definite, the matrix inversion can be implemented with $N^3/3 + 2N^2 \approx N^3/3$ flops through a Cholesky factorization~\cite{boyd2004convex}. All this makes the E-step computationally efficient. 

\subsubsection{M-step: Model Estimation}
In the M-step, the Gaussian model selection $k_i$ and the signal estimate $\tilde{\bbf}_i$ of all the patches are assumed to be known. Let 
$\mathcal{C}_k$ be the ensemble of the patch indices $i$ that are assigned to the $k$-th Gaussian model, i.e., $\mathcal{C}_k = \{i|k_i=k\}$, and let $|\mathcal{C}_k|$ be its cardinality. The parameters of each Gaussian model are estimated with the maximum likelihood estimate using all the patches assigned to that Gaussian cluster,
\begin{equation}
\label{eqn:ML:gaussian}
(\tilde{\mu}_k, \tilde{\Sigma}_k) = \arg \max_{\mu_k, \Sigma_k} \log p(\{\tilde{\bbf}_i\}_{i \in \mathcal{C}_k}|\mu_k, \Sigma_k). 
\end{equation}
With the Gaussian model~\eqref{eqn:multivariate:gaussian} , one can easily verify that the resulting estimate is the empirical estimate
\begin{equation}
\label{eqn:ML:covariacne}
\tilde{\mu}_k = \frac{1}{|\mathcal{C}_k|}\sum_{i \in \mathcal{C}_k} \tilde{\bbf}_i~~\textrm{and}~~
\tilde{\Sigma}_k = \frac{1}{|\mathcal{C}_k|} \sum_{i \in \mathcal{C}_k} (\tilde{\bbf}_i  - \tilde{\mu}_k) (\tilde{\bbf}_i  - \tilde{\mu}_k)^T. \end{equation}

The empirical covariance estimate may be improved through regularization when there is lack of data~\cite{schafer2005shrinkage} (for typical patch size $8 \times 8$, the dimension of the covariance matrix ${\Sigma}_k $ is $64 \times 64$, while the $|\mathcal{C}_k|$ is typically in the order of a few hundred). A simple and standard eigenvalue-based regularization is used here, $\tilde{\Sigma}_k \leftarrow \tilde{\Sigma}_k + \varepsilon Id$,  where $\epsilon$ is a small constant. The regularization also guarantees that the estimate $\tilde{\Sigma}_k$ of the covariance matrix is full-rank, so that the Wiener filter~\eqref{eqn:MAP:wiener} is always well defined. This is important for the Gaussian model selection~\eqref{eqn:MAP:model:selection} as well, since if $\tilde{\Sigma}_k$ is not full rank, then $\log \left| \tilde{\Sigma}_k \right| \rightarrow - \infty$, biasing the model selection. The computational complexity of the M-step is negligible with respect to the E-step. 

As the MAP-EM algorithm described above iterates, the MAP probability of the observed signals \\$p(\{\tilde{\bbf}_i\}_{1 \leq i \leq I} | \{\by_i\}_{1 \leq i \leq I}, \{\tilde{\mu}_k, \tilde{\Sigma}_k\}_{1 \leq k \leq K})$ always increases. This can be observed by interpreting the E- and M-steps as a coordinate descent optimization~\cite{hathaway1986another}.  In the experiments, the convergence of the patch clustering and resulting PSNR is always observed. 

\section{PLE and Structured Sparse Estimation}
\label{sec:MAP:EM:sparsity}

The MAP-EM algorithm described above requires an initialization. A good initialization is highly important for iterative algorithms that try to solve non-convex problems, and remains  an active research topic~\cite{baudry2010combining, fraley1998many}. This section describes a dual structured sparse interpretation of GMM and MAP-EM, which suggests an effective dictionary motivated initialization for the MAP-EM algorithm. Moreover, it shows that the resulting piecewise linear estimate stabilizes traditional sparse inverse problem estimation. 

The sparse inverse problem estimation approaches will be first reviewed. After describing the connection between MAP-EM and structured sparsity via estimation in PCA bases, an intuitive and effective initialization will be presented. 

\subsection{Sparse Inverse Problem Estimation}
\label{sec:sparse:l1}
Traditional sparse super-resolution estimation in dictionaries provides
effective non-parametric approaches to inverse problems, although the coherence of the dictionary and 
their large degree of freedom may become sources of instability and errors. These
algorithms are briefly reviewed in this section. ``Super-resolution'' is loosely used here 
as these approaches try to recover information that is lost after the degradation. 

A signal $\bbf \in \mathbb{R}^N$ is estimated
by taking advantage of prior information which specifies a
dictionary $\bD \in \R^{N \times |\Gamma|}$, having $ |\Gamma|$ columns corresponding to 
atoms $\{ \phi_m \}_{m \in  \Gamma}$, where $\bbf$ has
a sparse approximation. This dictionary may be a basis or some
redundant frame, with $ |\Gamma| \geq N$.
Sparsity means that $\bbf$ is well approximated by its orthogonal
projection $\bbf_\La$ over a subspace $\V_\La$ generated by a small
number $|\supp| \ll  |\Gamma|$ of column vectors $\{ \phi_m \}_{m \in \La}$
of $\bD$:
\begin{equation}
\label{eqn:sparse:f}
\bbf = \bbf_\La + \epsilon_\La = \bD (\ba \cdot \mathbf{1}_\La) +  \epsilon_\La,
\end{equation}
where $\ba \in \mathbb{R}^{|\Gamma|}$ is the transform coefficient vector, $\ba \cdot \mathbf{1}_\Lambda$ selects the coefficients in $\Lambda$ and sets the others to zero, $\bD \, (\ba \cdot \mathbf{1}_\Lambda)$ multiplies the matrix $\bD$ with the vector $\ba \cdot \mathbf{1}_\Lambda$, and $\|\epsilon_\La\|^2 \ll \|\bbf\|^2$ is a small
approximation error. 

Sparse inversion algorithms try to estimate from the degraded signal $\by = \bU \bbf + \bw$ the support $\Lambda$ and the coefficients $\ba$ in $\Lambda$ 
that specify the projection of $\bbf$ in the approximation space $\V_\Lambda$.
It results from~\eqref{eqn:sparse:f} that
\begin{equation}
\label{eqn:sparse:y}
\by = \bU \bD (\ba \cdot \mathbf{1}_\La)  + \epsilon',
~~\textrm{with}~~~\epsilon' = \bU \epsilon + \bw.
\end{equation}
This means that $\by$ is well approximated by the same sparse set $\Lambda$ of
 atoms and the same coefficients $\ba$ in the transformed dictionary $\bU\bD$, whose columns
are the transformed vectors $\{\bU \phi_m \}_{m \in \Gamma}$. 

Since $\bU$ is not an invertible operator,
the transformed dictionary $\bU \bD$ is redundant, with column vectors which are
linearly dependent. It results that $\by$
has an infinite number of possible decompositions in $\bU \bD$. 
A sparse approximation $\tilde \by  = \bU \bD \tilde{\ba}$ of $\by$ 
can be calculated with a basis pursuit algorithm which minimizes a
Lagrangian penalized by a sparse $l_1$
norm~\cite{chen1999adb, tibshirani1996regression}
\begin{equation}
\label{eq:sparse:le} 
\tilde{\ba} = \arg \min_{\ba} \| \bU \bD {\ba} - \by \|^2
   + \lambda\, \|\ba\|_1,
\end{equation}
or with faster greedy matching pursuit algorithms~\cite{mallat1993matching}. The resulting 
sparse estimation of $\bbf$ is 
\begin{equation}
\label{eqn:sparse:inverse:recons}
\tilde{\bbf} =  \bD \tilde{\ba}.
\end{equation}

As we explain next, this simple approach is not straightforward and often not as effective as it seems.
The {\it Restrictive Isometry Property} of Cand\`es and Tao
\cite{candes2006near} and Donoho \cite{donoho2006compressed} is a 
strong sufficient condition which guarantees the correctness of the penalized
$l_1$ estimation. This restrictive isometry property is valid for certain
classes of operators $\bU$, but not for important structured operators
such as subsampling on a uniform grid or convolution. For structured operators,
the precision and stability of this sparse inverse estimation
depends upon the ``geometry'' of the approximation support
$\La$ of $\bbf$, which is not well understood
mathematically, despite some sufficient exact recovery conditions
proved for example by Tropp~\cite{tropp2004greed}, and many others (mostly related to the coherence of the equivalent dictionary). Nevertheless, some necessary qualitative conditions 
for a precise and stable sparse super-resolution estimate~\eqref{eqn:sparse:inverse:recons} can be deduced as follows~\cite{mallat2008wts, mallat10SME}:
\begin{itemize}
\item \textbf{Sparsity.} $\bD$ provides a sparse representation for $\bbf$. 
\item \textbf{Recoverability.} The atoms have non negligible norms $\| \bU \phi_m\|^2 \gg 0$. If the degradation operator $\bU$ applied to
$\phi_m$ leaves no ``trace,'' the corresponding coefficient $\ba[m]$ can not be recovered from $\by$ with~\eqref{eq:sparse:le}. We will see in the next subsection that this recoverability property of transformed relevant atoms having sufficient energy is critical for the GMM/MAP-EM introduced in the previous section as well.
\item \textbf{Stability.}  The transformed dictionary $\bU \bD$ is incoherent enough. Sparse inverse problem estimation may be unstable if some columns $\{ \bU \phi_m\}_{m \in \Ga}$ in $\bU \bD$ are too similar. To see this, let us imagine a toy example, where a constant-value atom and a highly oscillatory atom (with values $-1,1,-1,1,\ldots$), after a $\times 2$ subsampling, become identical. The sparse estimation~\eqref{eq:sparse:le} can not distinguish between them, which results in an unstable inverse problem estimate~\eqref{eqn:sparse:inverse:recons}. The coherence of  $\bU \bD$ depends on $\bD$ as well as on the operator $\bU$. Regular operators $\bU$ such as subsampling on a uniform grid and  convolution, usually lead to a coherent $\bU \bD$, which makes accurate inverse problem estimation difficult. 
\end{itemize}


Several authors have applied this sparse super-resolution framework~\eqref{eq:sparse:le} and~\eqref{eqn:sparse:inverse:recons}
for image inverse problems. Sparse estimation in dictionaries
of curvelet frames and DCT have been applied successfully to image
inpainting~\cite{elad2005simultaneous, fadili2009inpainting, guleryuz2006nonlinear}.
However, for uniform grid interpolations, Section~\ref{sec:zooming}
shows that the resulting interpolation estimations are not as precise as simple
linear bicubic interpolations. A contourlet zooming algorithm~\cite{mueller2007iiu}
can provide a slightly better PSNR than a bicubic interpolation, but
the results are considerably below the state-of-the-art. Learned dictionaries of image
patches have generated good inpainting results~\cite{mairal2008sparse,
zhou2010nonparametric}. In some recent works sparse super-resolution algorithms 
with learned dictionary have been studied for zooming and deblurring~\cite{lou2009direct, yang2010SR}. 
As shown in sections~\ref{sec:zooming} and~\ref{sec:deblurring}, although they sometimes produce
good visual quality, they often generate artifacts and the resulting PSNRs are not as good as more standard methods.  

Another source of instability of these algorithms comes from their full degree of freedom. 
The non-linear approximation space $\V_\Lambda$ is estimated by selecting the approximation support $\Lambda$, with basically no constraint.
A selection of $|\La|$ atoms from a dictionary of size $|\Gamma|$ thus corresponds to a choice of an approximation space
among ${|\Gamma|} \choose{|\La|}$ possible subspaces. In a local patch-based sparse estimation with $8 \times 8$ patch size, typical values of $|\Gamma|=256$ and $|\La|=8$ lead to a huge degree of freedom ${{256} \choose{8}} \sim 10^{14}$, further stressing the inaccuracy of estimating $\ba$ from an $\bU \bD$.

These issues are addressed with the proposed PLE framework and its mathematical connection with structured sparse models described next. 

\subsection{Structured Sparse Estimation in PCA bases}
\label{subsubsec:PCA}

The PCA bases bridge the GMM/MAP-EM framework presented in Section~\ref{sec:SSMS} with the sparse estimation described above. For signals $\{\bbf_i\}$ following a statistical distribution, a PCA basis is defined as the matrix that diagonalizes the data covariance matrix $\Sigma_k = E[\bbf_i \bbf_i^T]$,
\begin{equation}
\label{eqn:covariance:diag}
\Sigma_{k} = \bB_k \bS_k \bB_k^T,
\end{equation}
where $\bB_k$ is the PCA basis and $\bS_k = \mathrm{diag}(\lambda_{1}^k, \ldots, \lambda_{N}^k)$ is a diagonal matrix, whose diagonal elements $\lambda_1^k \geq \lambda_2^k \geq \ldots \geq \lambda_N^k$ are the sorted eigenvalues. It can be shown that the PCA basis is orthonormal, i.e., $\bB_k \bB_k^T = Id$, each of its columns $\phi_k^m$, $1 \leq m \leq N$, being an atom that represents one principal direction. The eigenvalues are non-negative, $\lambda_m \geq 0$, and measure the energy of the signals $\{\bbf_i\}$ in each of the principal directions~\cite{mallat2008wts}.

Transforming $\bbf_i$ from the canonical basis to the PCA basis $\ba_i^k = \bB_k^T \bbf_i$, 
one can verify that the MAP estimate~\eqref{eqn:MAP:estimate:gaussian}-\eqref{eqn:MAP:wiener} can be equivalently calculated as
\begin{equation}
\label{eqn:est:reconstruct:k}
\tilde{\bbf}_i^k= \bB_k \tilde{\ba}_i^k ,
\end{equation}
where, following simple algebra and calculus, the MAP estimate of the PCA coefficients $\tilde{\ba}_i^k$ is obtained by 
\begin{equation}
\label{eqn:MAP:estimate:gaussian:PCA}
\tilde{\ba}_i^k = \arg \min_{{\ba}_i}  \left( \|\bU_i \bB_k {\ba}_i - \by_i\|^2 + \sigma^2 \sum_{m=1}^N \frac{|{\ba}_i[m]|^2}{\lambda_m^k} \right). 
\end{equation}

\begin{figure}[htbp]
\vspace{-5ex}
\begin{center}
\begin{tabular}{cc}
\epsfxsize=4cm \epsffile{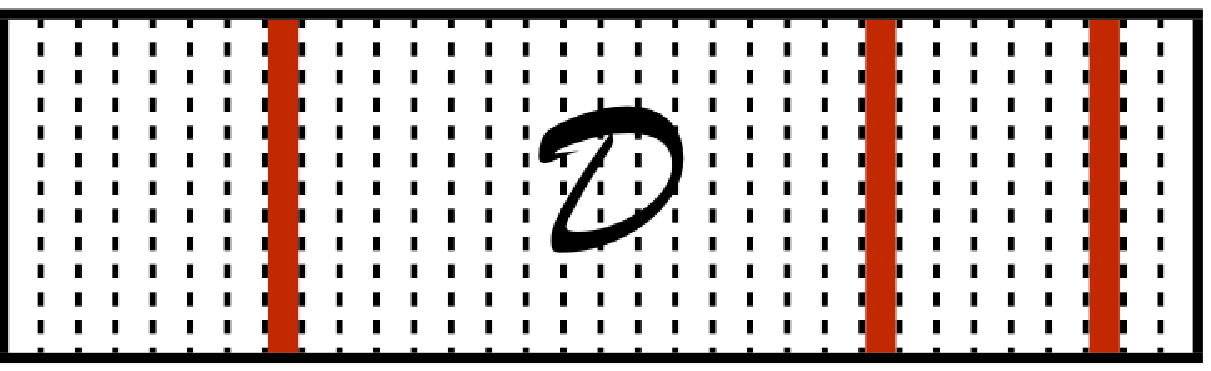} &
\epsfxsize=5.4cm \epsffile{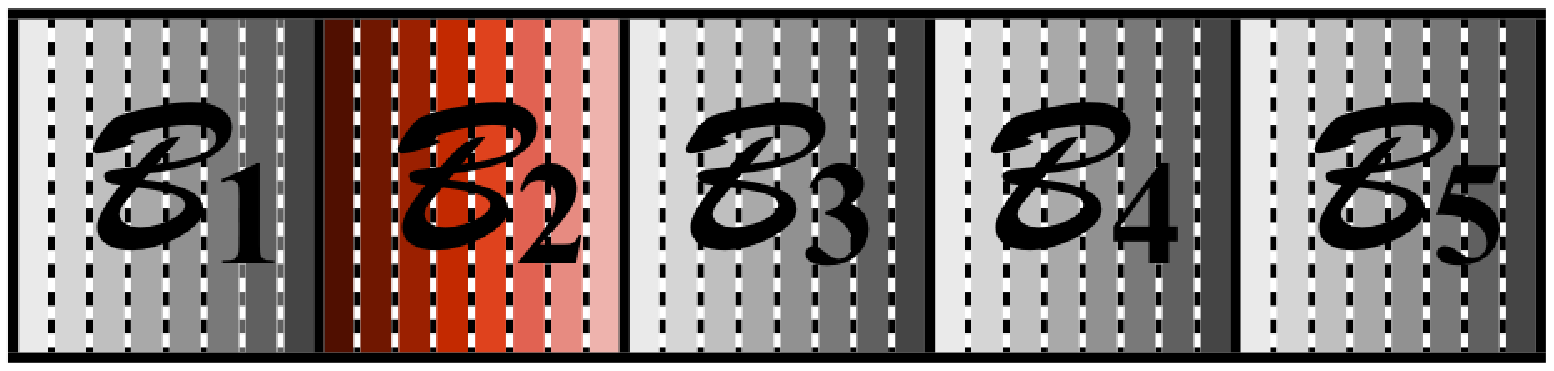} \end{tabular}
\end{center}
\vspace{-5ex}
\caption{\small Left: Traditional overcomplete dictionary. Each column represents an atom in the dictionary. Non-linear estimation has the full degree of freedom to select any combination of atoms (marked by the columns in red). Right: The underlying structured sparse piecewise linear dictionary of the proposed approach. The dictionary is composed of a family of PCA bases whose atoms are pre-ordered by their associated eigenvalues. For each image patch, an optimal \textit{linear} estimator is calculated in each PCA basis and the best linear estimate among the bases is selected (marked by the basis in red).} \label{fig:dict}
\vspace{-5ex}
\end{figure}

Comparing~\eqref{eqn:MAP:estimate:gaussian:PCA} with~\eqref{eq:sparse:le}, the MAP-EM estimation can thus be interpreted as a structured sparse estimation. As illustrated in Figure~\ref{fig:dict}, the proposed dictionary has the advantage of the traditional learned overcomplete dictionaries being overcomplete, and adapted to the image under test thanks to the Gaussian model estimation in the M-step (which is equivalent to updating the PCAs), but the resulting piecewise linear estimator (PLE) is more structured than the traditional nonlinear sparse estimation. PLE is calculated with a \textit{linear} estimation in each basis and a \textit{non-linear} best basis selection:
\begin{itemize}
\item \textbf{Nonlinear block sparsity.} The dictionary is composed of a union of $K$ PCA bases. To represent an image patch, the \textit{non-linear} model selection~\eqref{eqn:MAP:model:signal} in the E-step restricts the estimation to only one basis ($N$ atoms out of $KN$ selected in group), and has a degree of freedom equal to $K$, sharply reduced from that in the traditional sparse estimation which has the full degree of freedom in atom selection. 
\item \textbf{Linear collaborative filtering.} Inside each PCA basis, the atoms are pre-ordered by their associated eigenvalues (which decay very fast as we will later see, leading to sparsity inside the block as well). In contrast to the non-linear sparse $l_1$ estimation~\eqref{eq:sparse:le}, the MAP estimate~\eqref{eqn:MAP:estimate:gaussian:PCA} implements the regularization with the $l_2$ norm of the coefficients weighted by the eigenvalues $\{\lambda_m^k\}_{1 \leq m \leq N}$, and is calculated with a \textit{linear} filtering~\eqref{eqn:MAP:estimate:gaussian:solution}~\eqref{eqn:MAP:wiener}. The eigenvalues are computed from all the signals $\{\bbf_i\}$ in the same Gaussian distribution class. The resulting estimation therefore implements a collaborative filtering which incorporates the information from all the signals in the same cluster~\cite{abernethy2009new}. The weighting scheme privileges the coefficients ${\ba}_i[m]$ corresponding to the principal directions with large eigenvalues $\lambda_m$, where the energy is likely to be high, and penalizes the others. For the ill-posed inverse problems, the collaborative prior information incorporated in the eigenvalues $\{\lambda_m^k\}_{1 \leq m \leq N}$ further stabilizes the estimate. (Note that this collaborative weighting is fundamentally different than the standard one used in iterative weighted $l_2$ approaches to sparse coding~\cite{daubechies2009iteratively}.) 
\end{itemize}

As described in Section~\ref{sec:SSMS}, the complexity of the MAP-EM algorithm is dominated by the E-step. For an image patch size of $\sqrt{N} \times\sqrt{N}$ (typical value $8 \times 8$), it costs $2KN^2$ flops for translation-invariant degradation operators such as uniform subsampling and convolution, and $KN^3/3$ flops for translation-variant operators such as random masking, where $K$ is the number of PCA bases. The overall complexity is therefore tightly upper bounded by $\mathcal{O}(2LKN^2)$ or $\mathcal{O}(LKN^3/3)$, where $L$ is the number of iterations. As will be shown in Section~\ref{sec:initial:exp}, the algorithm converges fast for image inverse problems, typically in $L=3$ to 5 iterations. On the other hand, the complexity of the $l_1$ minimization with the same dictionary is $\mathcal{O}(KN^3)$, with typically a large factor in front as the $l_1$ converges slowly in practice. The MAP-EM algorithm is thus typically one or two orders of magnitude faster than the sparse estimation. 

To conclude, let as come back to the recoverability property mentioned in the previous section. We see from~\eqref{eqn:MAP:estimate:gaussian:PCA} that if an eigenvector of the covariance matrix is killed by the operator $\bU_i$, then its contribution to the recovery of  $\by_i$ is virtually null, while it pays a price proportional to the corresponding eigenvalue. Then, it will not be used in the optimization~\eqref{eqn:MAP:estimate:gaussian:PCA}, and thereby in the reconstruction of the signal following~\eqref{eqn:est:reconstruct:k}. This means that the wrong model might be selected and an improper reconstruction obtained. This further stresses the importance of a correct design of dictionary elements, which from the description just presented, it is equivalent to the correct design of the covariance matrix, including the initialization, which is described next. 

\subsection{Initialization of MAP-EM}
\label{subsec:init}

The PCA formulation just described not only reveals the connection between the MAP-EM estimator and structured sparse modeling, but it is crucial for understanding how to initialize the Gaussian models as well. 

\subsubsection{Sparsity}
\label{subsubsec:init:sparsity}
\begin{figure}[htbp]
\vspace{-2ex}
\begin{center}
\begin{tabular}{cccc}
\epsfxsize=2.9cm \epsffile{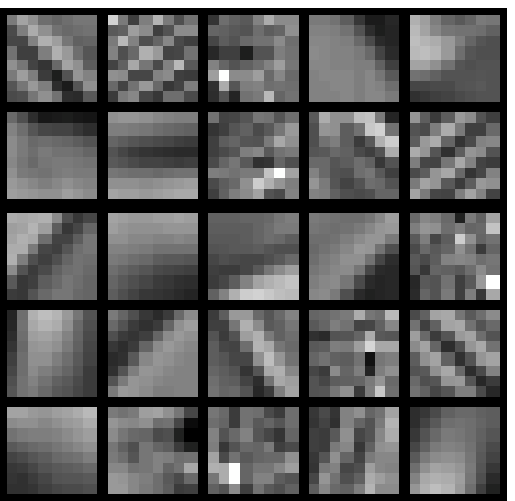}  &
\epsfxsize=3.25cm \epsffile{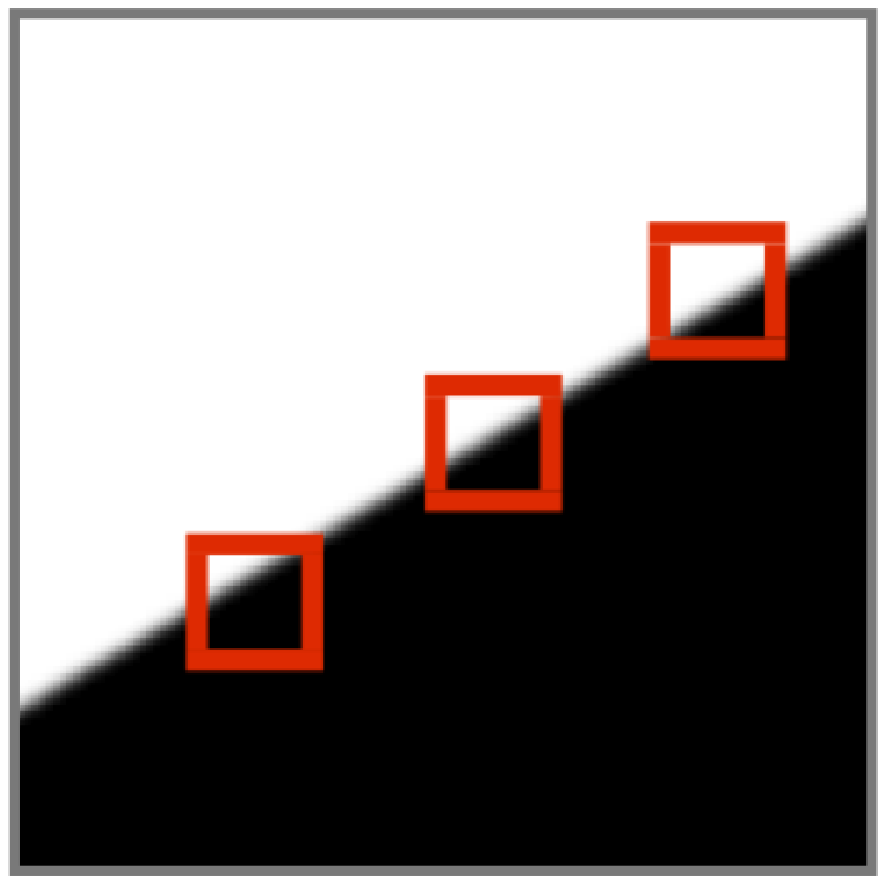} &
\epsfxsize=6cm \epsffile{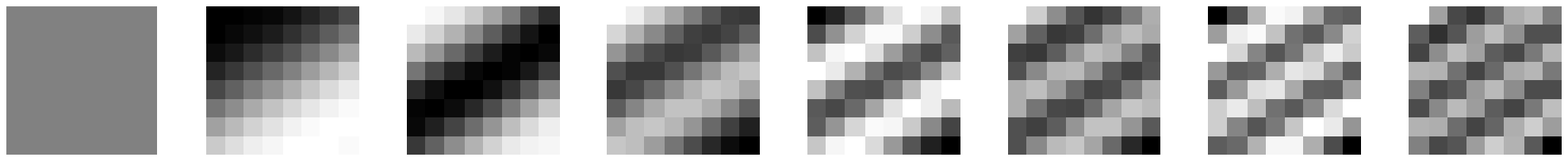}  &
\epsfxsize=2cm \epsffile{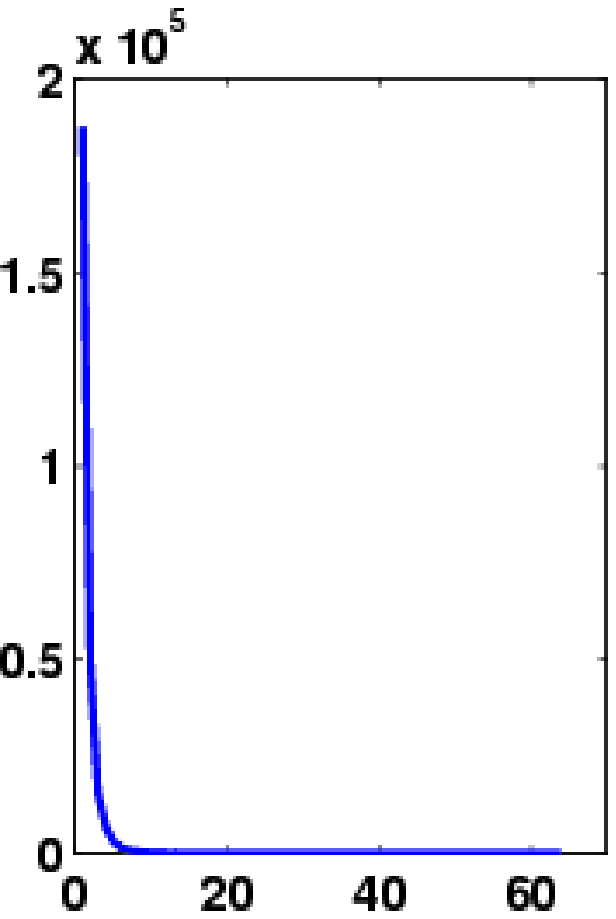}  \\
\textbf{(a)} & \textbf{(b)} & \textbf{(c)} & \textbf{(d)}
\end{tabular}
\end{center}
\vspace{-3ex}
\caption{\small (a) Some typical dictionary atoms learned from the image Lena (Figure~\ref{fig:Lena:clustering}-(a)) with K-SVD~\cite{aharon2006k}. (b)-(d) A numerical procedure to obtain the initial directional PCAs. (b) A synthetic edge image. Patches ($8 \times 8$) that touch the edge are used to calculate an initial PCA basis. (c) The first 8 atoms in the PCA basis with the largest eigenvalues. (d) Typical eigenvalues. } \label{fig:PCA:init} 
\vspace{-3ex}
\end{figure}

As explained in Section~\ref{sec:sparse:l1},  for the sparse inverse problem estimation model to have the super-resolution ability, the first requirement on the dictionary is to be able to provide sparse representations of the image. It has been shown that capturing image directional regularity is highly important for sparse representations~\cite{aharon2006k, candes2004new, mallat2008orthogonal}. In dictionary learning, for example, most prominent atoms look like local edges good at representing contours, as illustrated in Figure~\ref{fig:PCA:init}-(a). Therefore the initial PCAs in our framework, which following~\eqref{eqn:covariance:diag} will lead to the initial Gaussians, are designed to capture image directional regularity.  

The initial directional PCA bases are calculated following a simple numerical procedure. Directions from $0$ to $\pi$ are uniformly sampled to $K$ angles, and one PCA basis is calculated per angle. The calculation of the PCA at an angle $\theta$ uses a synthetic blank-and-white edge image following the same direction, as illustrated in Figure~\ref{fig:PCA:init}-(b). Local patches that touch the contour are collected and are used to calculate the PCA basis (following~\eqref{eqn:ML:covariacne} and~\eqref{eqn:covariance:diag}). The first atom, which is almost DC, is replaced by DC, and a Gram-Schmidt orthogonalization is calculated on the other atoms to ensure the orthogonality of the basis. The patches contain edges that are translation-invariant. As the covariance of a stationary process is diagonalized by the Fourier basis, unsurprisingly, the resulting PCA basis has first few important atoms similar to the cosines atoms oscillating in the direction $\theta$ from low-frequency to high-frequency, as shown in Figure~\ref{fig:PCA:init}-(c). Comparing with the Fourier vectors, these PCAs enjoy the advantage of being free of the periodic boundary issue, so that they can provide sparse representations for local image patches. The eigenvalues of all the bases are initiated with the same ones obtained from the synthetic contour image, that have fast decay, Figure~\ref{fig:PCA:init}-(d). These, following~\eqref{eqn:covariance:diag}, complete the covariance initialization. The Gaussian means are initialized with zeros. 

It is worth noting that this directional PCA basis not only provides sparse representations for contours and edges, but it captures well textures of the same directionality as well. Indeed, in a space of dimension $N$ corresponding to patches of size $\sqrt{N} \times \sqrt{N}$, the first about $\sqrt{N}$ atoms illustrated in Figure~\ref{fig:PCA:init}-(c) absorb most of the energy in local patterns following the same direction in real images, as indicated by the fast decay of the eigenvalues (very similar to Figure~\ref{fig:PCA:init}-(d)). 

A typical patch size is $\sqrt{N} \times \sqrt{N} = 8 \times 8$, as selected in previous works~\cite{aharon2006k, elad2006image}. The number of directions in a local patch is limited due to the pixelization. The DCT basis is also included in competition with the directional bases to capture isotropic image patterns. Our experiments have shown that in image inverse problems, there is a significant average gain in PSNR when $K$ grows from $0$ to $3$ (when $K=0$, the dictionary is initialized with only a DCT basis and all the patches are assigned to the same cluster), which shows that one Gaussian model, or equivalently a single linear estimator, is not enough to accurately describe the image. When $K$ increases, the gain reduces and gets stabilized at about $K=36$. Compromising between performance and complexity, $K=18$, which corresponds to a $10^\circ$ angle sampling step, is selected in all the future experiments. 

\begin{figure}[htbp]
\vspace{-2ex}
\begin{center}
\begin{tabular}{ccccc}
\hspace{-4ex}\epsfxsize=3.5cm \epsffile{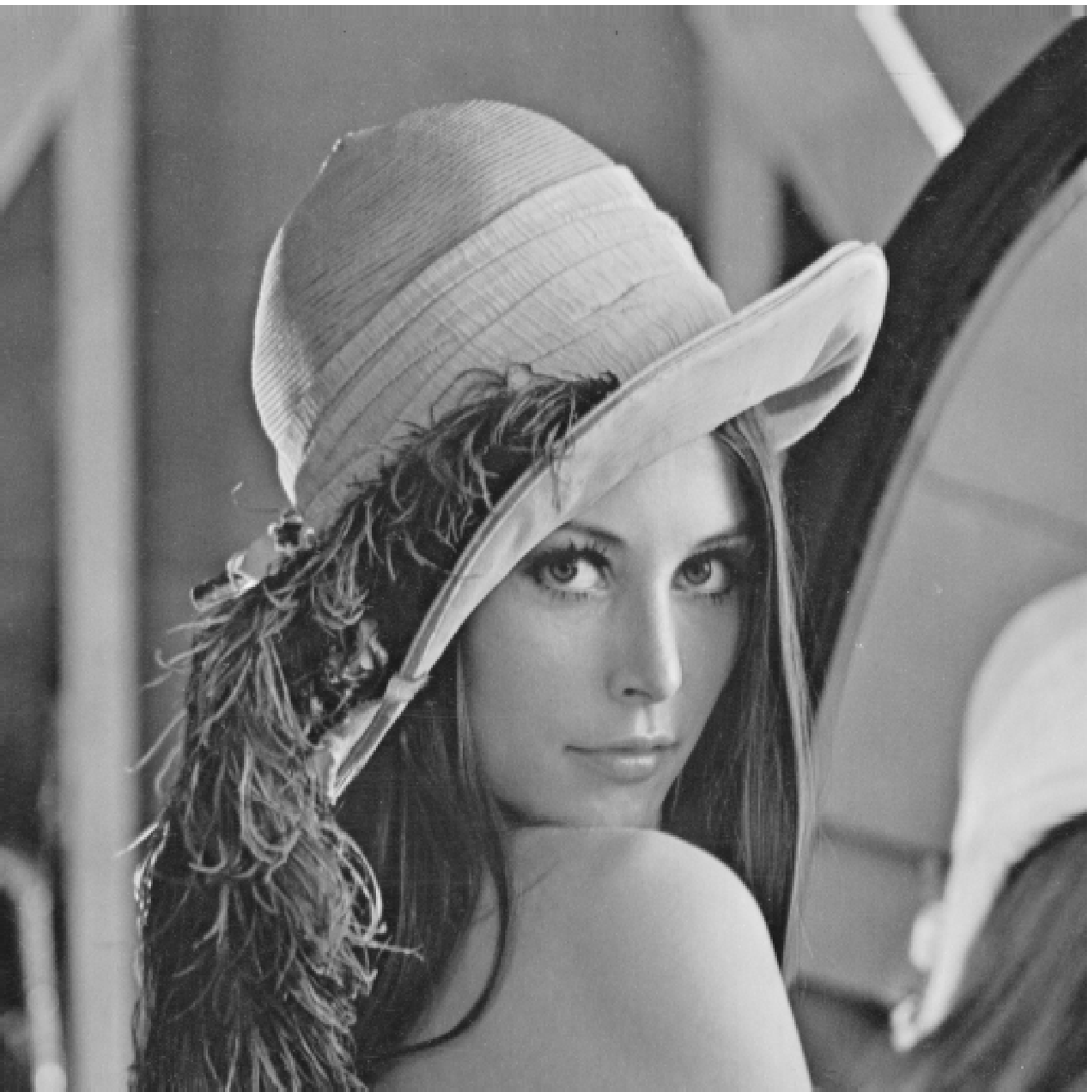} &
\hspace{-2ex}\epsfxsize=3.5cm \epsffile{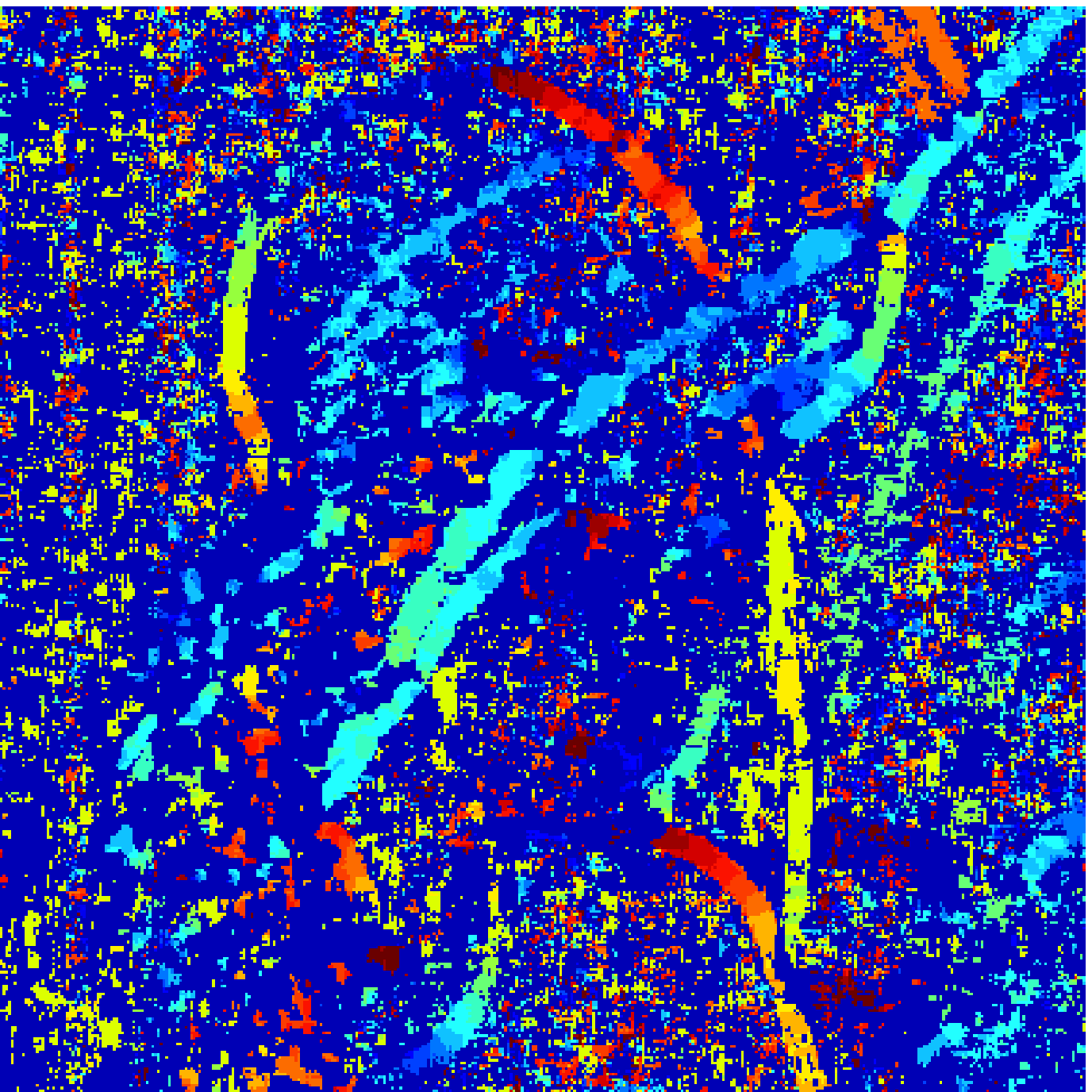}  &
\hspace{-2ex}\epsfxsize=3.5cm \epsffile{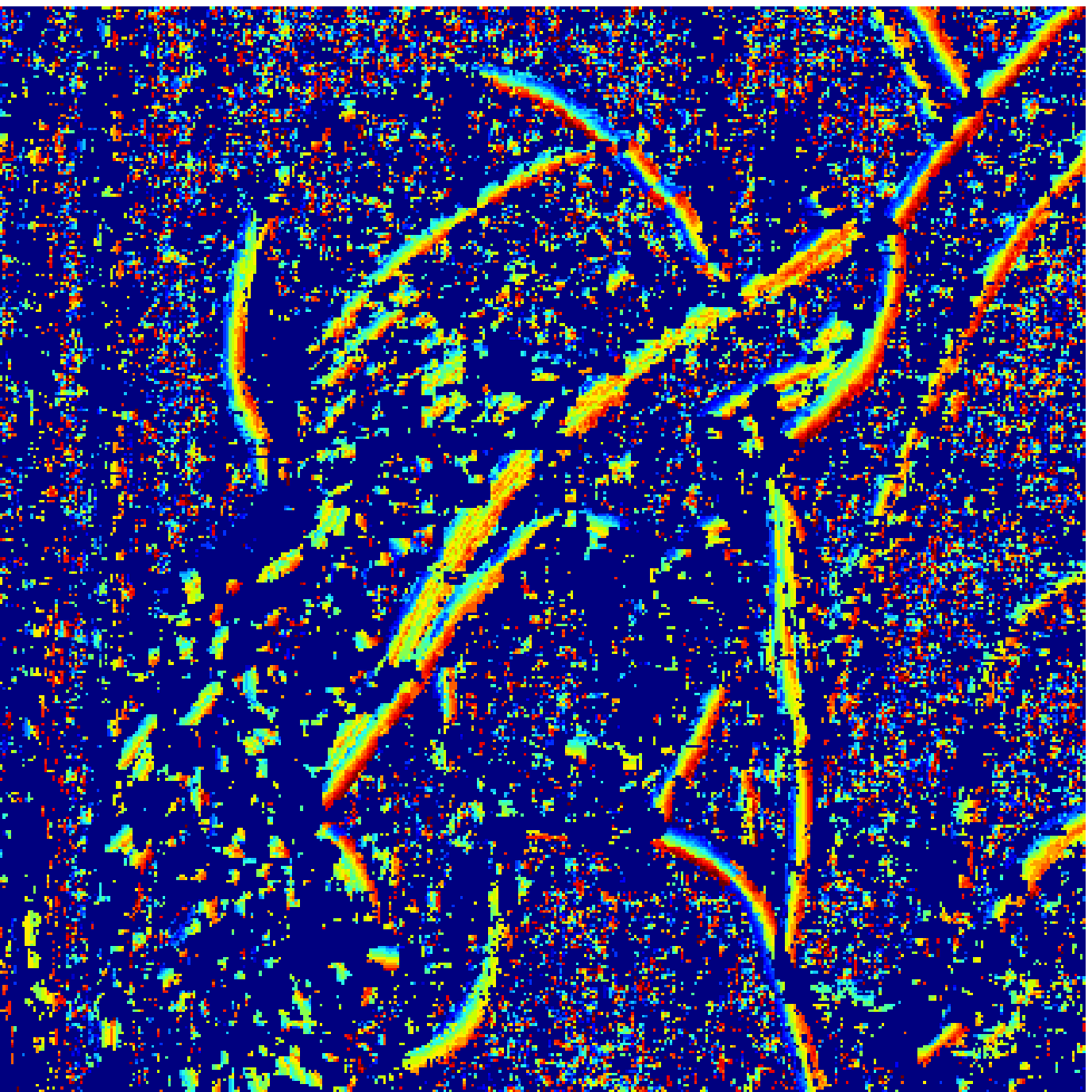} &
\hspace{-2ex}\epsfxsize=3.5cm \epsffile{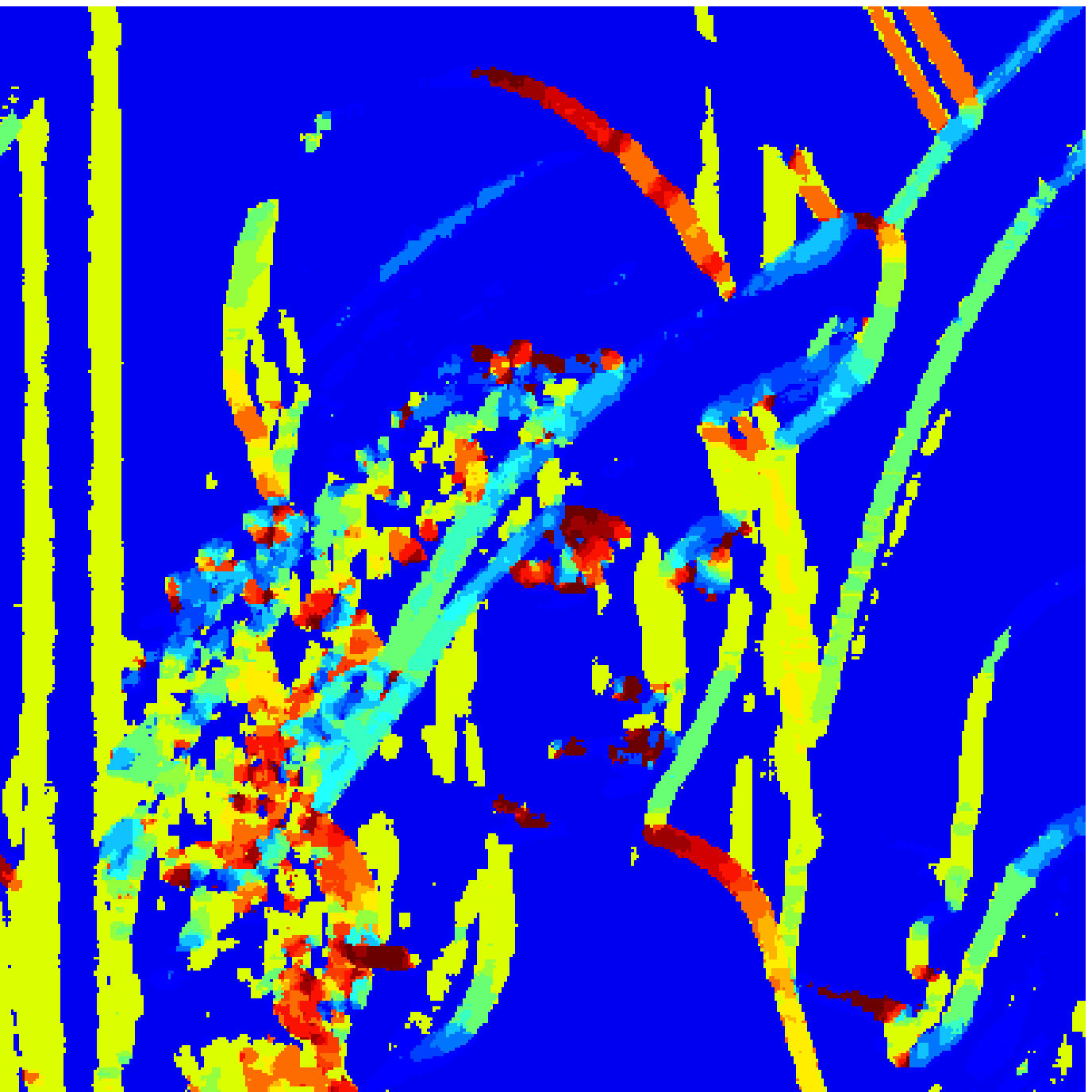}  &
\hspace{-2ex}\epsfxsize=3.5cm \epsffile{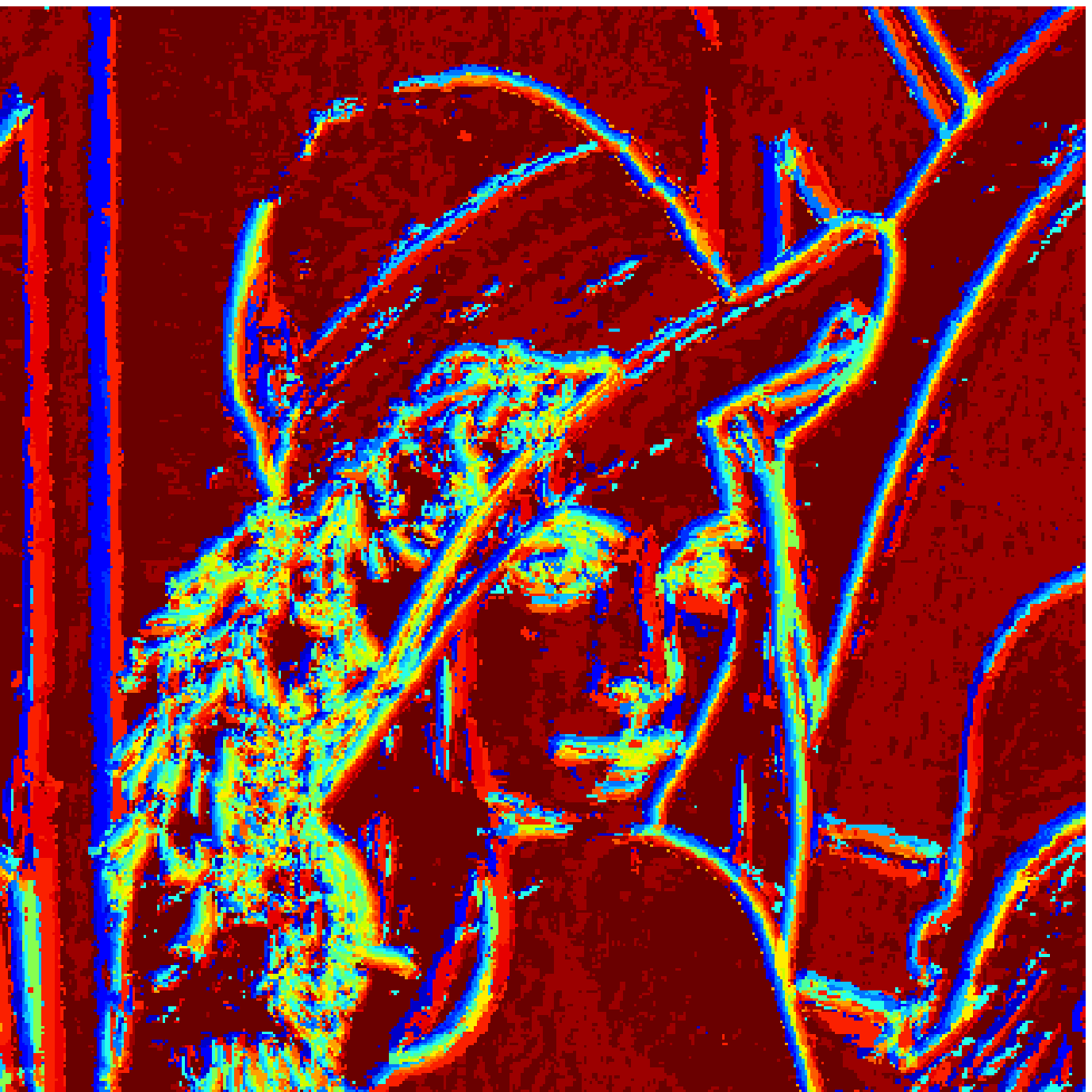} \\
\hspace{-4ex}\textbf{(a)} & \hspace{-2ex}\textbf{(b)} & \hspace{-2ex}\textbf{(c)} & \hspace{-2ex} \textbf{(d)} &\hspace{-2ex} \textbf{(e)} \\
\end{tabular}
\end{center}
\vspace{-3ex}
\caption{\small (a). Lena image. ((b) to (d) are color images.) (b). Patch clustering obtained with the initial directional PCAs (see Figure~\ref{fig:PCA:init}-(c)). The patches are densely overlapped and each pixel represents the model $k_i$ selected for the $8 \times 8$ patch around it, different colors encoding different direction values of $k_i$, from 1 to $K=19$. (c). Patch clustering obtained with the initial position PCAs (see Figure~\ref{fig:PCA:init:pos}). Different colors encoding different position values of $k_i$, from 1 to $P=12$. (d) and (e). Patch clustering with respectively directional and position PCAs after the 2nd iteration.} \label{fig:Lena:clustering} 
\vspace{-4ex}
\end{figure}

Figures~\ref{fig:Lena:clustering}-(a) and (b) illustrates the Lena image and the corresponding patch clustering, i.e., the model selection $k_i$, obtained for the above initialization, calculated with~\eqref{eqn:MAP:model:selection} in the E-step described in Section~\ref{sec:SSMS}. The patches are densely overlapped and each pixel in Figure~\ref{fig:Lena:clustering}-(b) represents the model $k_i$ selected for the $8 \times 8$ patch around it, different colors encoding different values of $k_i$, from 1 to 19 (18 directions plus a DCT). One can observe, for example on the edges of the hat, that patches where the image patterns follow similar directions are clustered together, as expected. Let us note that on the uniform regions such as the background, where there is no directional preference, all the bases provide equally sparse representations. As the $\log |\Sigma_k| = \Pi_{m=1}^N \lambda_{m}^k$ term in the model selection~\eqref{eqn:MAP:model:selection} is initialized as identical for all the Gaussian models, the clustering is random is these regions. The clustering will improve as the MAP-EM progresses. 

\subsubsection{Recoverability}
\label{subsubsec:reconverability}
\begin{figure}[htbp]
\vspace{-2ex}
\begin{center}
\begin{tabular}{c}
\epsfxsize=10cm \epsffile{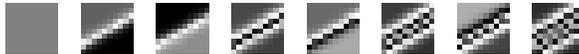} 
\end{tabular}
\end{center}
\vspace{-4ex}
\caption{\small The first 8 atoms in the position PCA basis with the largest eigenvalues.} \label{fig:PCA:init:pos} 
\vspace{-3ex}
\end{figure}

The oscillatory atoms illustrated in Figure~\ref{fig:PCA:init}-(c) are spread in space. Therefore, for diagonal operators in space such as masking and subsampling, they satisfy well the recoverability condition  $\|\bU \phi_m^k\|^2 \ge 0$ for super-resolution described in Section~\ref{sec:sparse:l1}. However, as these oscillatory atoms have Dirac supports in Fourier, for convolution operators, the recoverability condition is violated. For convolution operators $\bU$, $\|\bU \phi_m^k\|^2 \ge 0$ requires that the atoms have spread Fourier spectrum, and therefore be localized in space. Following a similar numerical scheme as described above, patches touching the edge at a \textit{fixed} position are extracted from synthetic edge images with different amounts of blur. The resulting PCA basis, named position PCA basis hereafter, contains localized atoms of different polarities and at different scales, following the same direction $\theta$, as illustrated in Figure~\ref{fig:PCA:init:pos} (which look like wavelets along the appropriate direction). For each direction $\theta$, a family of localized PCA bases $\{\bB_{k,p}\}_{1 \leq p \leq P}$ are calculated at all the positions translating within the patch. The eigenvalues are initialized with the same fast decay ones as illustrated in Figure~\ref{fig:PCA:init}-(d) for all the position PCA bases. Each pixel in Figure~\ref{fig:Lena:clustering}-(c) represents the model $p_i$ selected for the $8 \times 8$ patch around it, different colors encoding different position values of $p_i$, from 1 to 12. The rainbow-like color transitions on the edges show that the position bases are accurately fitted to the image structures. 


\subsubsection{Wiener Filtering Interpretation}
\begin{figure}[htbp]
\vspace{-3ex}
\begin{center}
\begin{tabular}{cccc}
\epsfxsize=1.6cm \epsffile{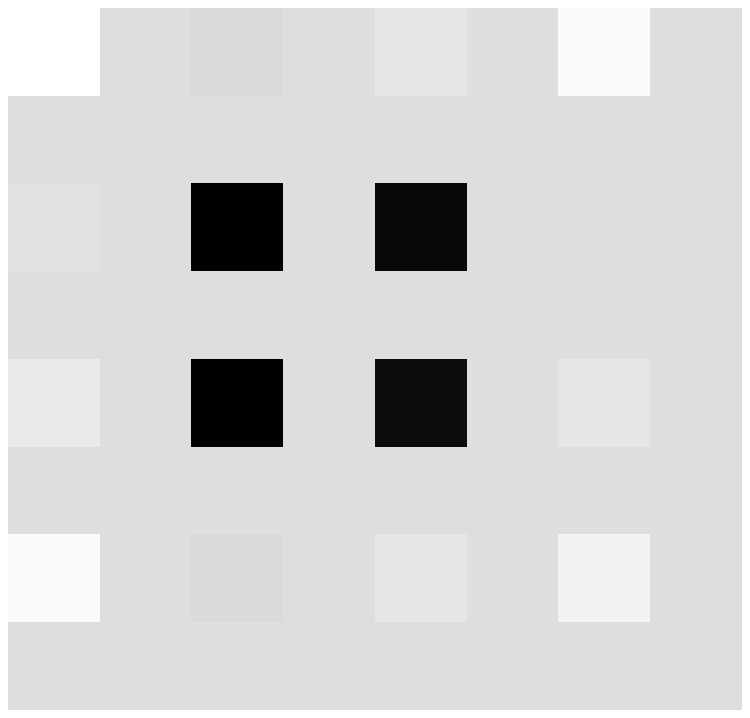}  &
\epsfxsize=1.6cm \epsffile{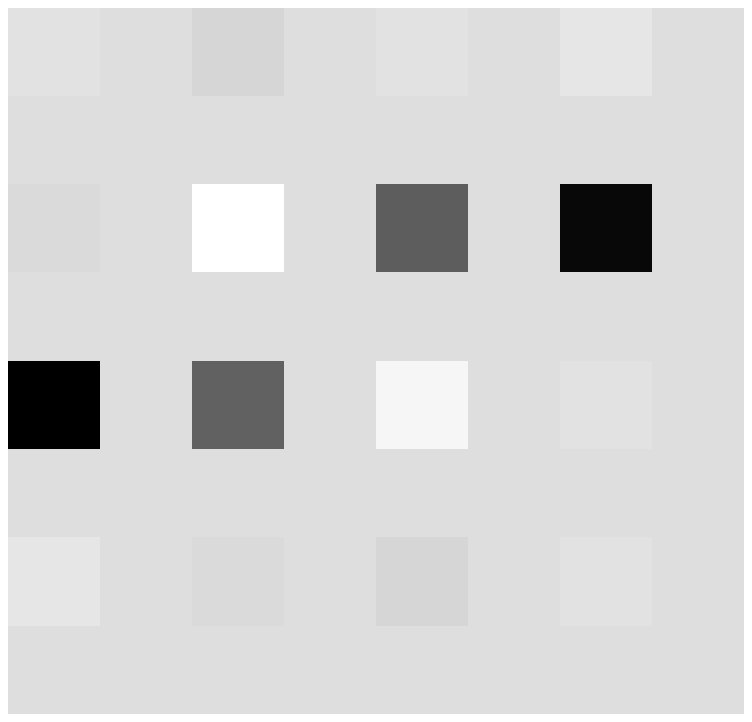}  &
\epsfxsize=1.6cm \epsffile{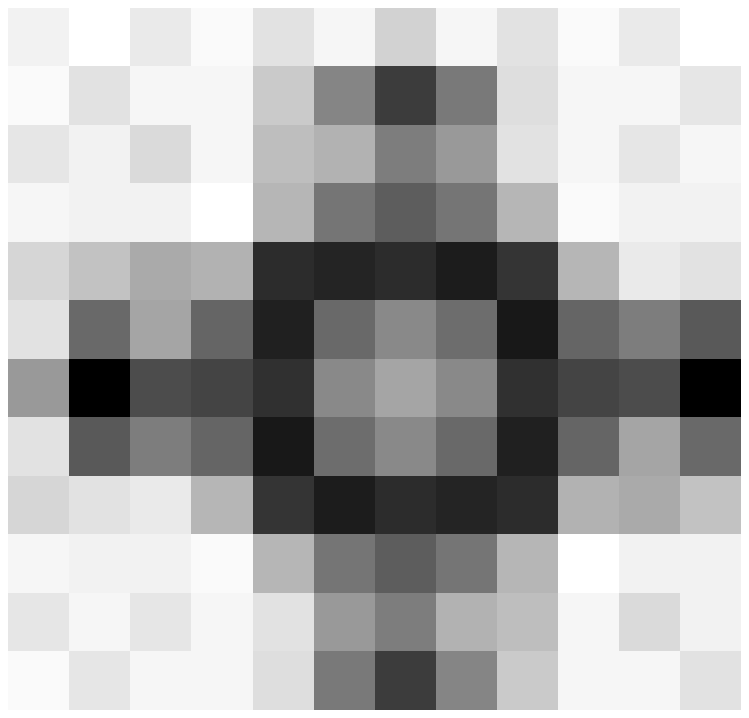}  &
\epsfxsize=1.6cm \epsffile{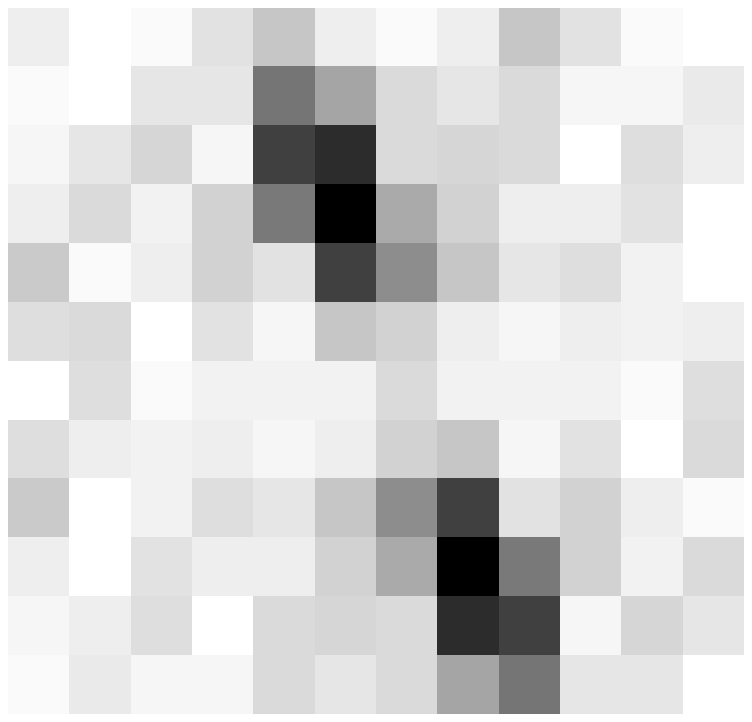}  \vspace{-1ex} \\
\textbf{(a)} & \textbf{(b)} & \textbf{(c)} &  \textbf{(d)}\\
\end{tabular}
\end{center}
\vspace{-3ex}
\caption{\small Some filters generated by the MAP estimator. (a) and (b) are for image zooming, where the degradation operator $\bU$ is a $2\times2$ subsampling operator. Gray-level from white to black: values from negative to positive. (a) is computed with a Gaussian distribution whose PCA basis is a DCT basis, and it implements an isotropic interpolator. (b) is computed with a Gaussian distribution whose PCA basis is a directional PCA basis (angle $\theta = 30^\circ$), and it implements a directional interpolator.  (c) and (d) are shown in Fourier and are for image deblurring, where the degradation operator $\bU$ is a Gaussian convolution operator. Gray-level from white to black: Fourier modules from zero to positive. (c) is computed with a Gaussian distribution whose PCA basis is a DCT basis, and it implements an isotropic deblurring filter.  (d) is computed with a Gaussian distribution whose PCA basis is a directional PCA basis (angle $\theta = 30^\circ$, at a fixed position), and it implements a directional deblurring filter.} \label{fig:MAP:filters} 
\vspace{-5ex}
\end{figure}

Figure~\ref{fig:MAP:filters} illustrates some typical Wiener filters, which are the rows of $\bW_{k}$ in~\eqref{eqn:MAP:wiener}, calculated with the initial PCA bases described above for zooming and deblurring. The filters have intuitive interpretations, for example directional interpolator for zooming and directional deconvolution for deblurring, confirming the effectiveness of the initialization. 

\subsection{Additional comments on related works}
Before proceeding with experimental results and applications, let us further comment on some related works, in addition to those already addressed in Section~\ref{sec:intro}. 

The MAP-EM algorithm using various probability distributions such as Gaussian, Laplacian, Gamma and Gibbs have been widely applied in medical image reconstruction and analysis (see for example~\cite{zhou2007bayesian, liang2009approach}). Following the Gaussian mixture models, MAP-EM alternates between image patch estimation and clustering, and Gaussian models estimation. Clustering-based estimation has been shown effective for image restoration. To achieve accurate clustering-based estimation, an appropriate clustering is at the heart. In a denoising setting where images are noisy but not degraded by the linear operator $\bU$, clustering with block matching, i.e., calculating Euclidian distance between image patch gray-levels~\cite{buades2006review, dabov2007image, mairal2009non}, and with image segmentation algorithms such as k-means on local image features~\cite{chatterjee2009clustering}, have been shown to improve the denoising results. For inverse problems where the observed images are degraded, for example images with holes in an inpainting setting, clustering becomes more difficult. The generalized PCA~\cite{vidal2005generalized} models and segments data using an algebraic subspace clustering technique based on polynomial fitting and differentiation, and while it has been shown effective in image segmentation, it does not reach state-of-the-art in image restoration. In the recent non-parametric Bayesian approach~\cite{zhou2010nonparametric},  an image patch clustering is implemented with probability models, which improves the denoising and inpainting results, although still under performing, in quality and computational cost, the framework here introduced. The clustering in the MAP-EM procedure enjoys the advantage of being completely consistent with the signal estimation, and in consequence leads to state-of-the-art results in a number of imaging inverse problem applications.

\section{Initial Supportive Experiments}
\label{sec:initial:exp}

Before proceeding with detailed experimental results for a number of applications of the proposed framework, this section shows through some basic experiments the effectiveness and importance of the initialization proposed above, the evolution of the representations as the MAP-EM algorithm iterates, as well as the improvement brought by the structure in PLE with respect to traditional sparse estimation.

Following some recent works, e.g.,~\cite{mairal2008learning}, an image is decomposed into $128 \times 128$ regions, each region treated with the MAP-EM algorithm separately. The idea is that image contents are often more coherent semi-locally than globally, and Gaussian model estimation or dictionary learning can be slightly improved in semi-local regions. This also saves memory and enables the processing to proceed as the image is being transmitted. Parallel processing on image regions is also possible when the whole image is available. Regions are half-overlapped to eliminate the boundary effect between the regions, and their estimates are averaged at the end to obtain the final estimate. 

\subsection{Initialization}
\begin{figure}[htbp]
\vspace{-4ex}
\begin{center}
\begin{tabular}{cccc}
\hspace{-6ex}\epsfxsize=5cm \epsffile{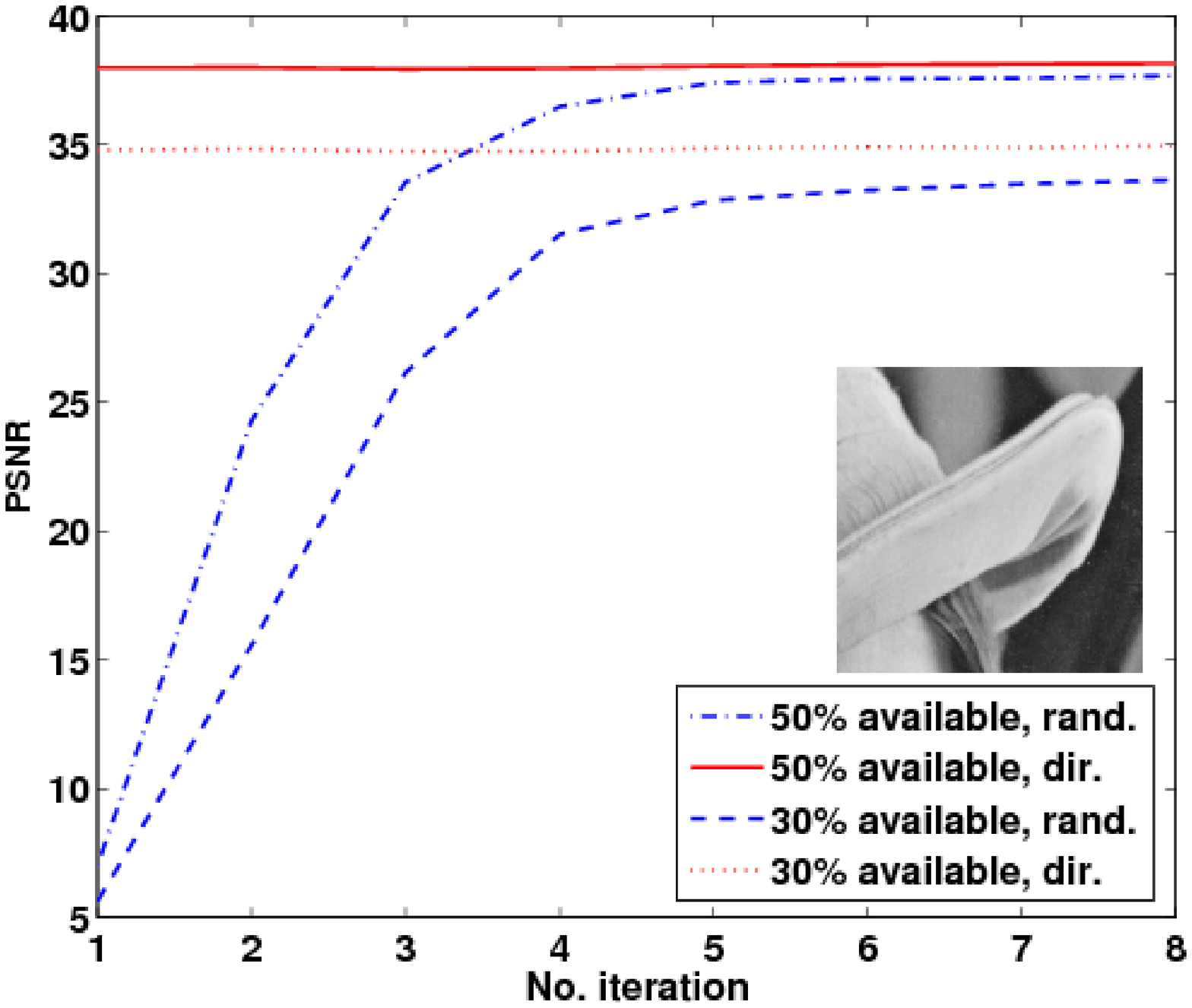}  &
\hspace{-5.5ex}\epsfxsize=5cm \epsffile{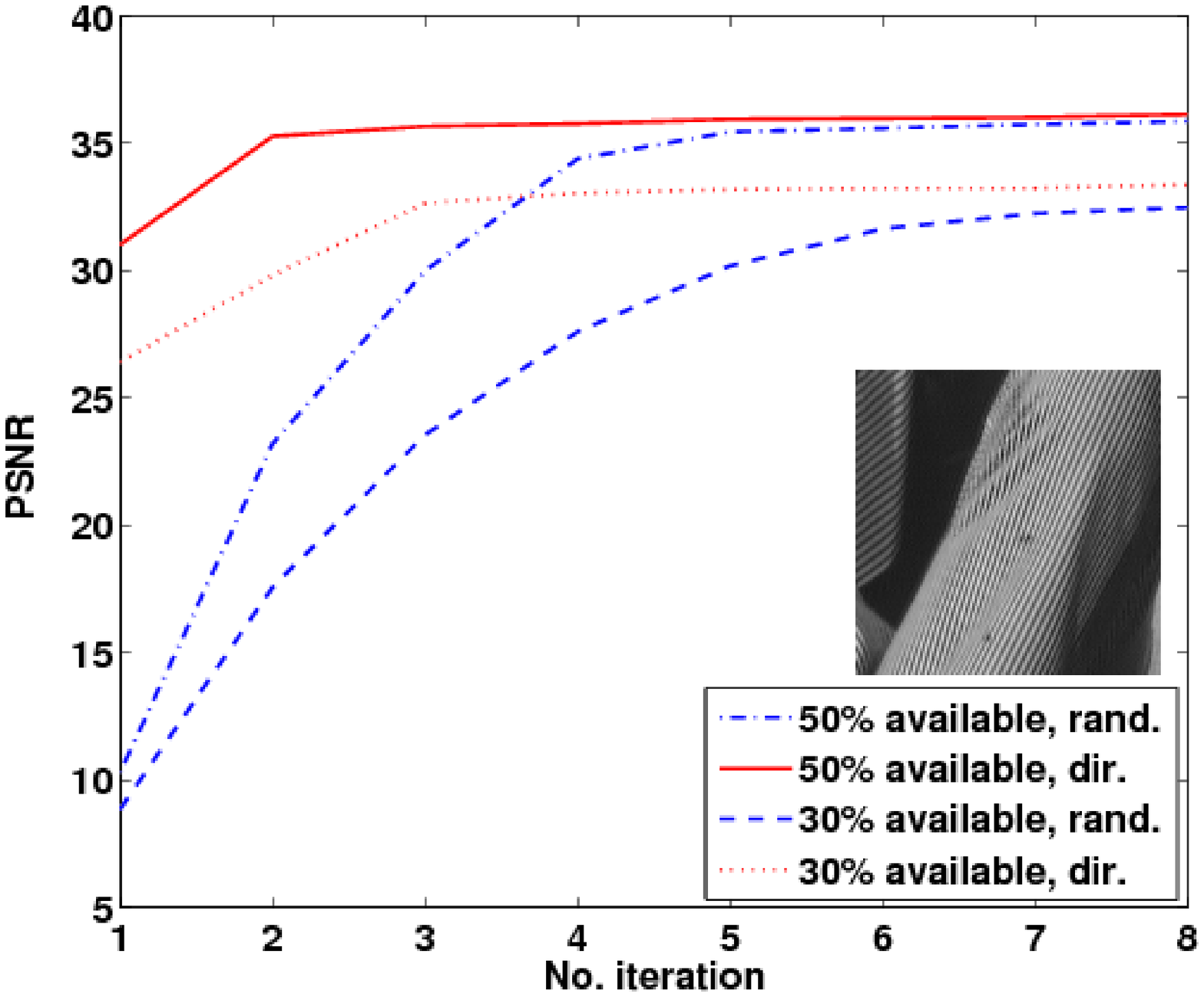}  &
\hspace{-5.5ex}\epsfxsize=5cm \epsffile{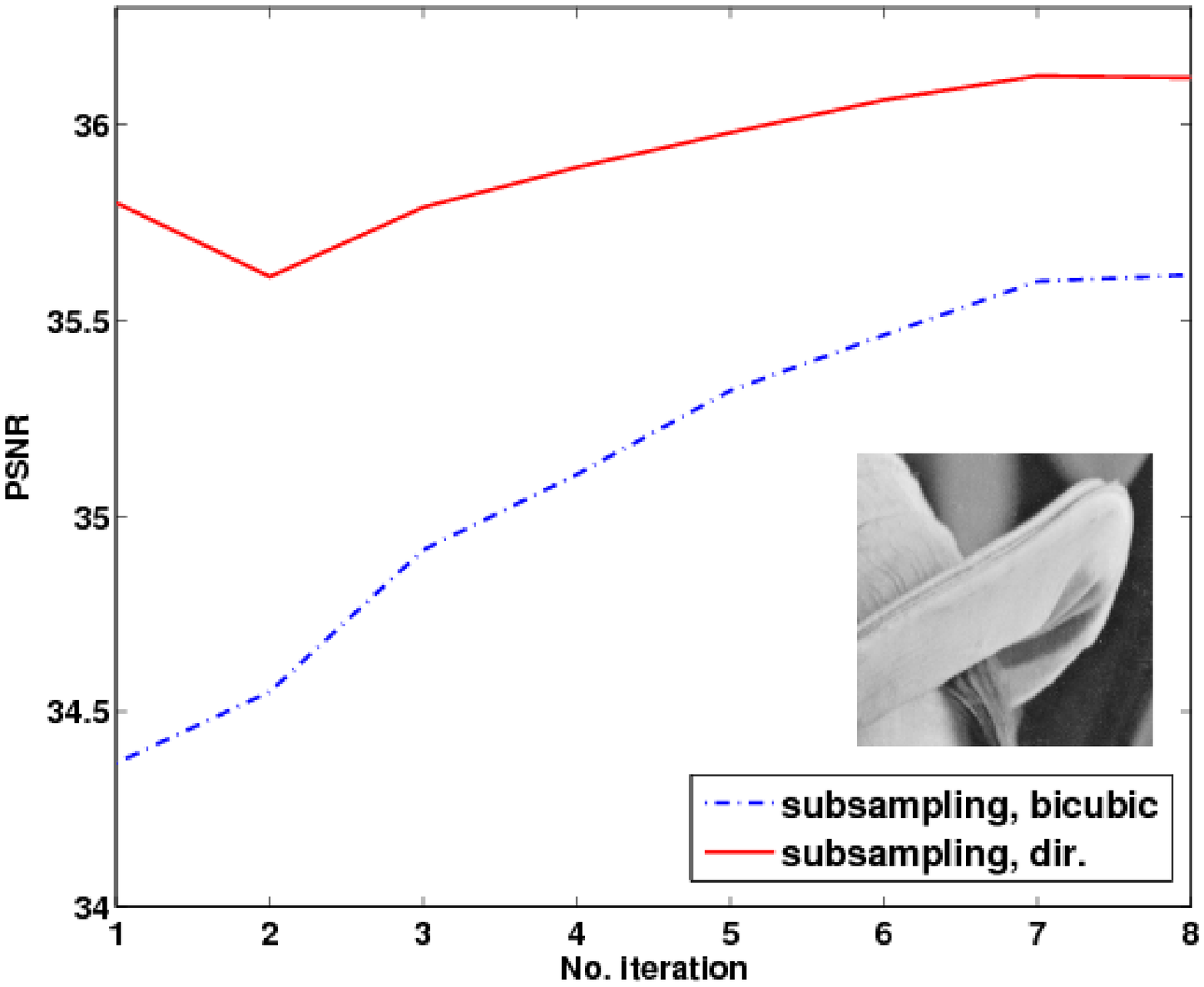} &
\hspace{-5.5ex}\epsfxsize=5cm \epsffile{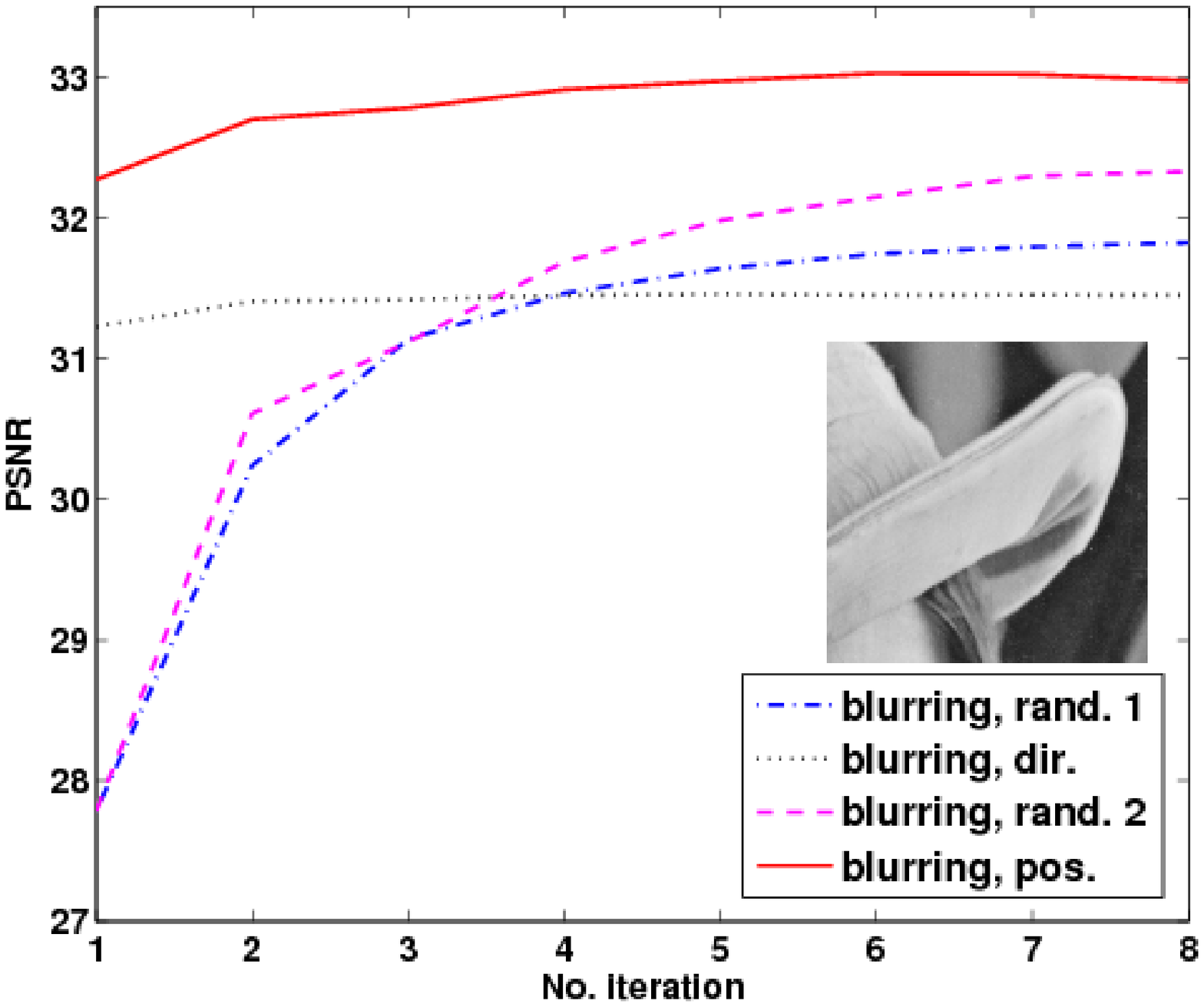} \vspace{-1ex}\\
\hspace{-6ex}\textbf{(a)} &\hspace{-5.5ex}\textbf{(b)}&\hspace{-5.5ex}\textbf{(c)}&\hspace{-5.5ex}\textbf{(d)}
\end{tabular}
\end{center}
\vspace{-3ex}
\caption{\small PSNR comparison of the MAP-EM algorithm with different initializations on different inverse problems. The horizontal axis corresponds to the number of iterations. (a) and (b). Inpainting with $50\%$ and $30\%$ available data, on Lena's hat and Barbara's cloth. The initializations under consideration are the random initialization and the initialization with directional PCA bases. (c) Zooming, on Lena's hat. The initializations under consideration are bicubic initialization and the initialization with directional PCA bases. (Random initialization completely fails to work.) (d) Deblurring, on Lena's hat.  The initializations under consideration are the initialization with directional PCAs ($K$ directions plus a DCT basis), the initialization with the \textit{position} PCA bases ($P$ positions per each of the $K$ directions), and two random initializations with the blurred image itself as the initial estimate and a random patch clustering with, respectively, $K+1$ (rand. 1) and $(K+1)P$ (rand. 2) clusters. See text for more details.} 
\label{fig:init}
\vspace{-4ex}
\end{figure}

Different initializations are compared in the context of different inverse problems, inpainting, zooming and deblurring. The reported experiments are performed on some typical image regions, Lena's hat with sharp contours and Barbara's cloth rich in texture, as illustrated in Figure~\ref{fig:init}. 

\noindent \textbf{Inpainting.}
In the addressed case of inpainting, the image is degraded by $\bU$, that is a random masking operator which randomly sets pixel values to zeros. The initialization described above is compared with a random initialization, which initializes in the E-step all the missing pixel value with zeros and starts with a random patch clustering. Figure~\ref{fig:init}-(a) and (b) compare the PSNRs obtained by the MAP-EM algorithm with those two initializations. The algorithm with the random initialization converges to a PSNR close to, about 0.4 dB lower than, that with the proposed initialization, and the convergence takes much longer time (about 6 iterations) than the latter (about 3 iterations). 

It is worth noting that on the contours of Lena's hat, with the proposed initialization the resulting PSNR is stable from the initialization, which already produces accurate estimation, since the initial directional PCA bases themselves are calculated over synthetic contour images, as described in Section~\ref{subsec:init}. 

\noindent \textbf{Zooming.} 
In the context of zooming, the degradation $\bU$ is a subsampling operator on a uniform grid, much structured than that for inpainting. The MAP-EM algorithm with the random initialization completely fails to work: It gets stuck in the initialization and does not lead to any changes on the degraded image. Instead of initializing the missing pixels with zeros, a bicubic initialization is tested, which initializes the missing pixels with bicubic interpolation. Figure~\ref{fig:init}-(c) shows that, as the MAP-EM algorithm iterates, it significantly improves the PSNR over the bicubic initialization, however, the PSNR after a slower convergence is still about 0.5 dB lower than that obtained with the proposed initialization.

\noindent \textbf{Deblurring.}
In the deblurring setting,  the degradation $\bU$ is a convolution operator, which is very structured, and the image is further contaminated with a white Gaussian noise. Four initializations are under consideration: the initialization with directional PCAs ($K$ directions plus a DCT basis), which is exactly the same as that for inpainting and zooming tasks, the proposed initialization with the \textit{position} PCA bases for deblurring as described in Section~\ref{subsubsec:reconverability} ($P$ positions per each of the $K$ directions, all with the same eigenvalues as for the directional PCAs initialization), and two random initializations with the blurred image itself as the initial estimate and a random patch clustering with, respectively, $K+1$ and $(K+1)P$ clusters. As illustrated in Figure~\ref{fig:init}-(d), the algorithm with the directional PCAs initialization gets stuck in a local minimum since the second iteration, and converges to a PSNR 1.5 dB lower than that with the initialization using the position PCAs. Indeed, since the recoverability condition for deblurring, as explained in Section~\ref{subsubsec:reconverability}, is violated with just directional PCA bases, the resulting images remain still quite blurred. The random initialization with $(K+1)P$ clusters results in better results than with $K+1$ clusters, which is 0.7 dB worse than the proposed initialization with position PCAs. 

These experiments confirm the importance of the initialization in the MAP-EM algorithm to solve inverse problems. The sparse coding dual interpretation of GMM/MAP-EM helps to deduce effective initializations for different inverse problems. While for inpainting with random masking operators, trivial initializations slowly converge to a solution moderately worse than that obtained with the proposed initialization, for more structured degradation operators such as uniform subsampling and convolution, simple initializations either completely fail to work or lead to significantly worse results than with the proposed initialization. 

\subsection{Evolution of Representations}

\begin{figure}[htbp]
\vspace{0ex}
\begin{center}
\begin{tabular}{ccccc}
\hspace{-2ex}\epsfxsize=3cm \epsffile{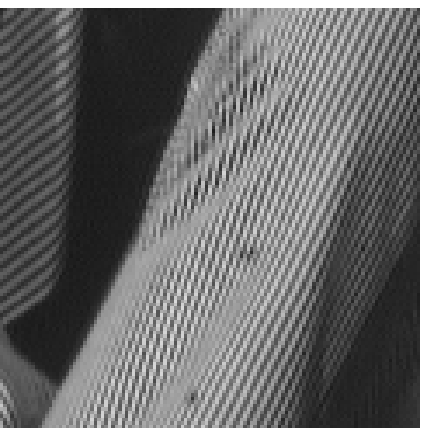} &
\hspace{0ex}\epsfxsize=3cm \epsffile{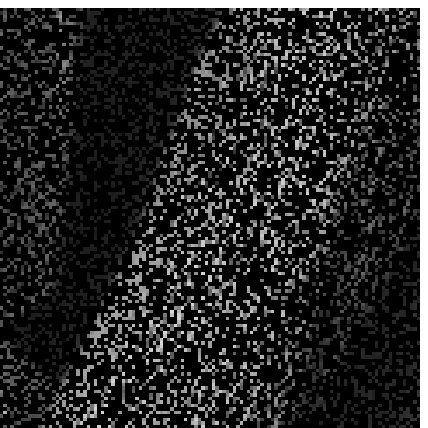}  &
\hspace{0ex}\epsfxsize=3cm\epsffile{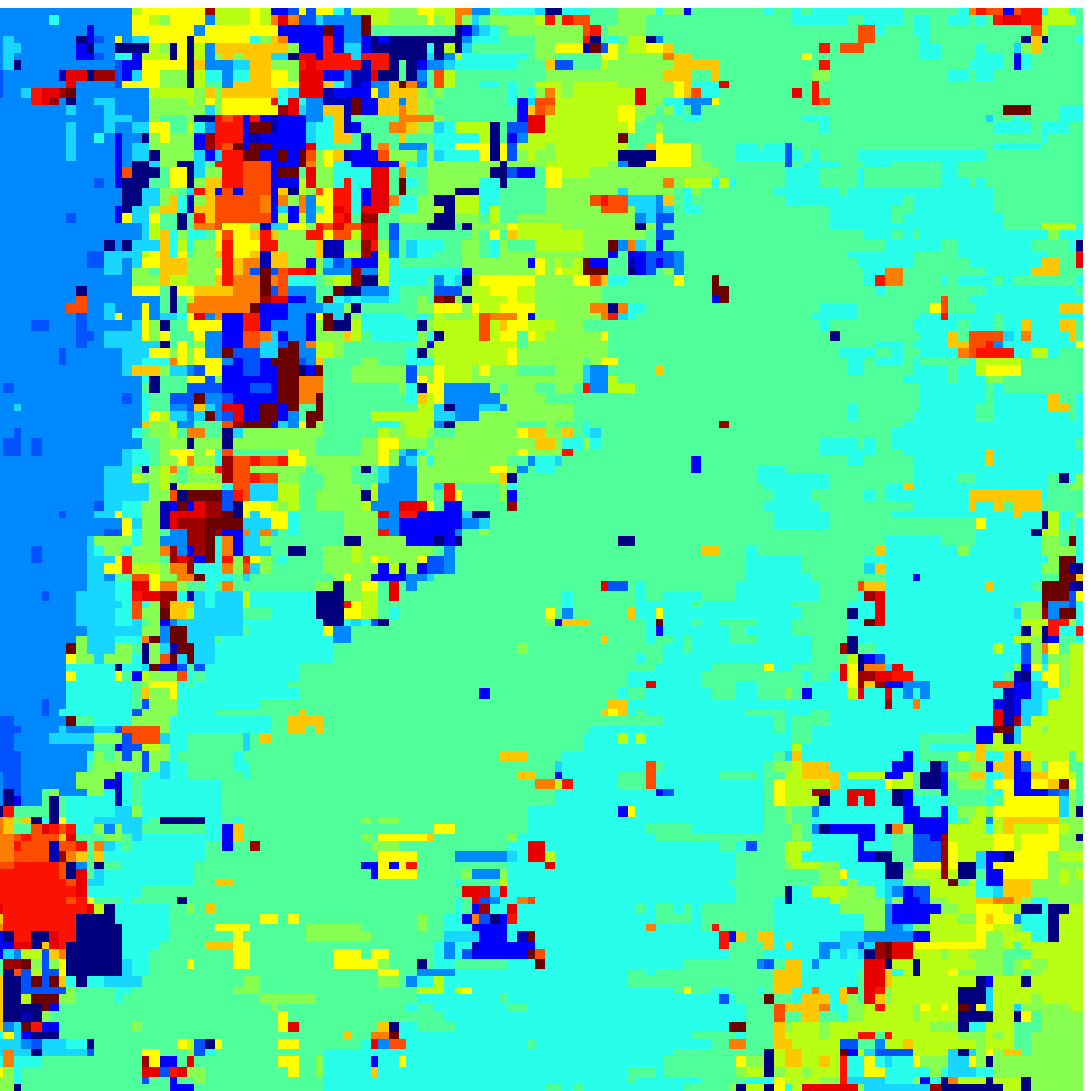} &
\hspace{0ex}\epsfxsize=3cm \epsffile{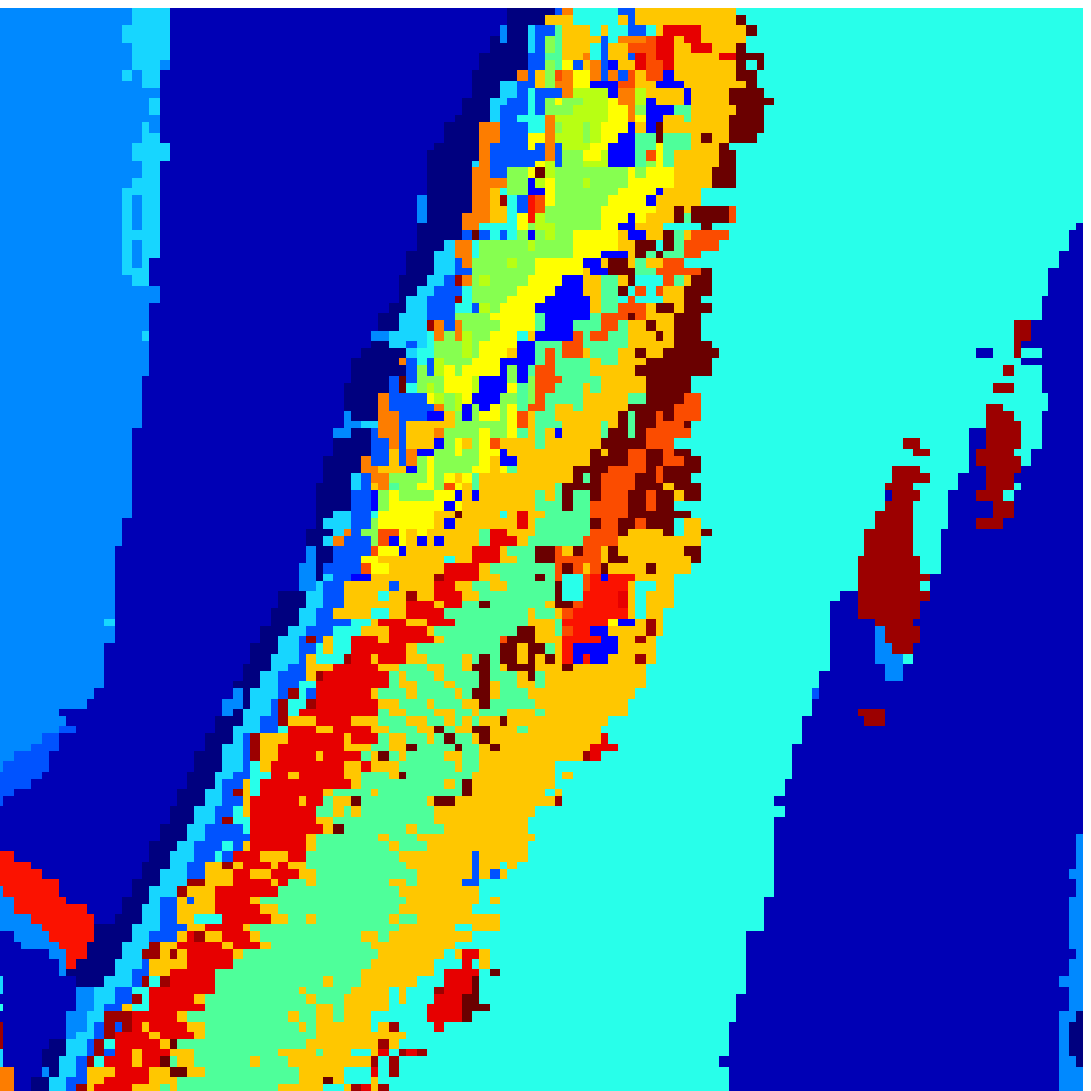} &
\hspace{0ex}\epsfxsize=3cm \epsffile{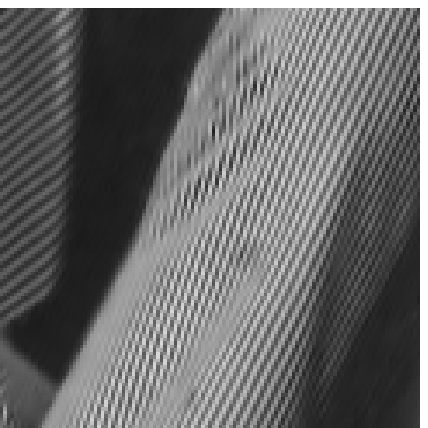}  \\
\hspace{-2ex} & & \hspace{-1ex} \epsfxsize=3cm \epsffile{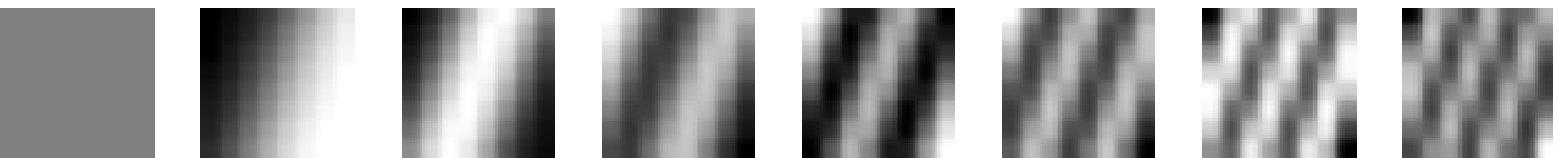} &  \hspace{-1ex} \epsfxsize=3cm \epsffile{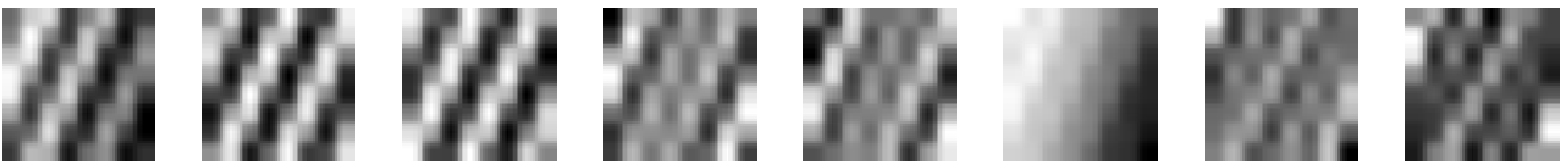}  & \vspace{1ex} \vspace{-1ex}\\
\hspace{-2ex} \textbf{(a)} &\textbf{(b)}&\textbf{(c)}&\textbf{(d)}&\textbf{(e)}
\end{tabular}
\end{center}
\vspace{-3ex}
\caption{\small Evolution of the representations. (a) The original image cropped from Barbara. (b) The image masked with $30\%$ available data. (c) and (d) are color images. (c) Bottom: The first few atoms of an initial PCA basis corresponding to the texture on the right of the image. Top: The resulting patch clustering after the 1st iteration. Different colors represent different clusters. (d) Bottom: The first few atoms of the PCA basis updated after the 1st iteration. Top: The resulting patch clustering after the 2nd iteration. (e) The inpainting estimate after the 2nd iteration (32.30 dB).} \label{fig:evolution:Barb}
\vspace{-2ex}
\end{figure}

Figure~\ref{fig:evolution:Barb} illustrates, in an inpainting context on Barbara's cloth, which is rich in texture, the evolution of the patch clustering as well as that of a typical PCA bases as the MAP-EM algorithm iterates. The clustering gets cleaned up as the algorithm iterates. (See figures~\ref{fig:Lena:clustering}-(d) and (e) for another example.) Some high-frequency atoms are promoted to better capture the oscillatory patterns, resulting in a significant PSNR improvement of more than 3 dB. On contour images such as Lena's hat illustrated in Figure~\ref{fig:init}, on the contrary, although the patch clustering is cleaned up as the algorithm iterates, the resulting local PSNR evolves little after the initialization, which already produces accurate estimation, since the directional PCA bases themselves are calculated over synthetic contour images, as described in Section~\ref{subsec:init}. The eigenvalues have always fast decay as the iteration goes on, visually similar to the plot in Figure~\ref{fig:PCA:init}-(d). The resulting PSNRs typically converge in 3 to 5 iterations.  

\subsection{Estimation Methods}

\begin{figure}[htbp]
\vspace{-0ex}
\begin{center}
\begin{tabular}{cccc}
\hspace{-2ex}
\epsfxsize=3cm \epsffile{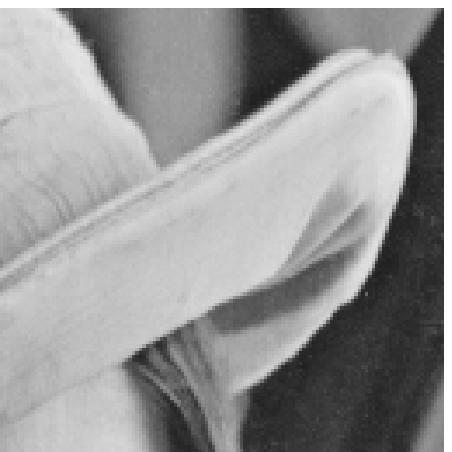}  &
\epsfxsize=3cm \epsffile{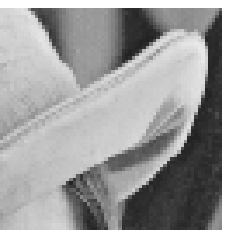}  &
\epsfxsize=3cm \epsffile{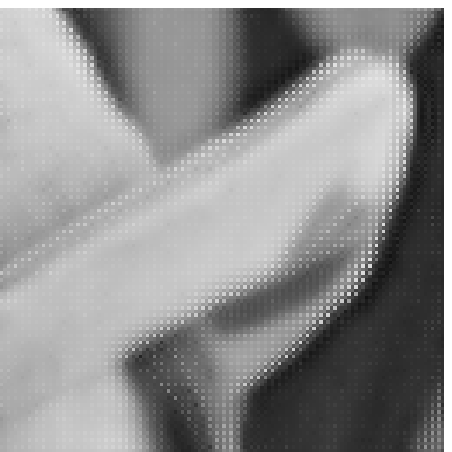}  &
\epsfxsize=3cm \epsffile{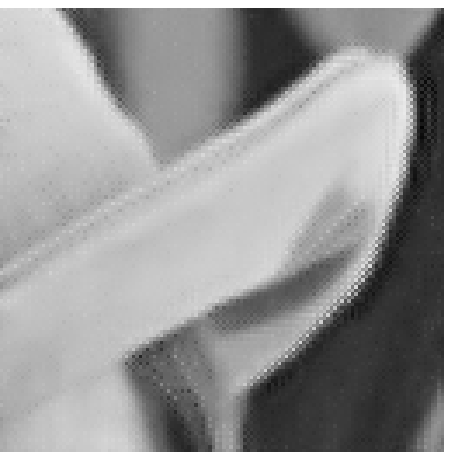}   \\
\hspace{-2ex}
\scriptsize{\textbf{(a) Original image.}} & \scriptsize{\textbf{(b) Low-resolution image.} }& \scriptsize{\textbf{(c) Global $l_1$: 22.70 dB}} & \scriptsize{\textbf{(d) Global OMP: 28.24 dB}}\vspace{2ex}\\
\hspace{-2ex}
\epsfxsize=3cm \epsffile{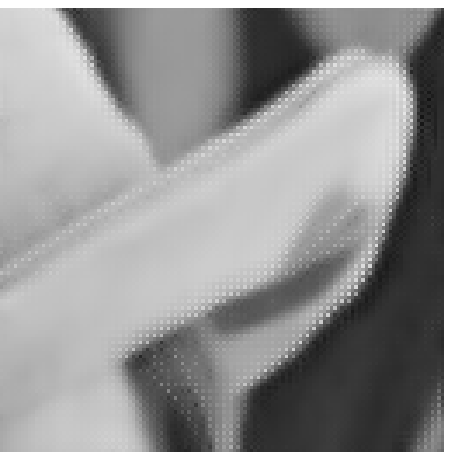}  &
\epsfxsize=3cm \epsffile{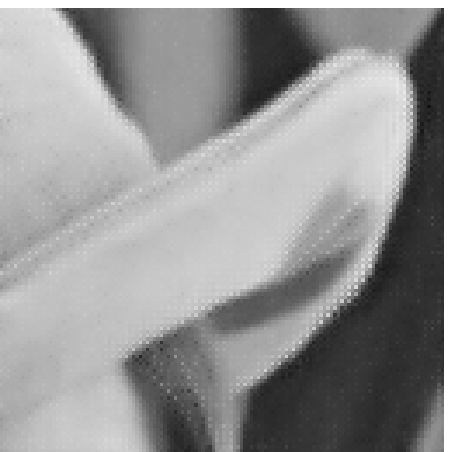}  &
\epsfxsize=3cm \epsffile{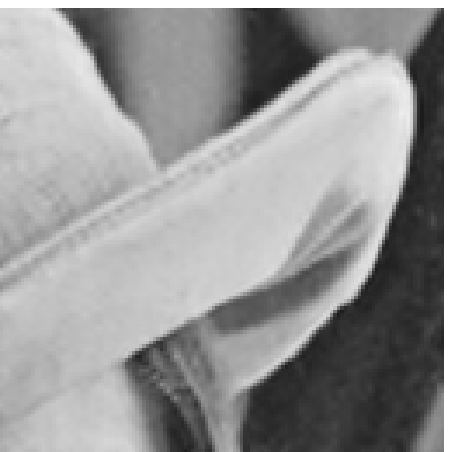}  &
\epsfxsize=3cm \epsffile{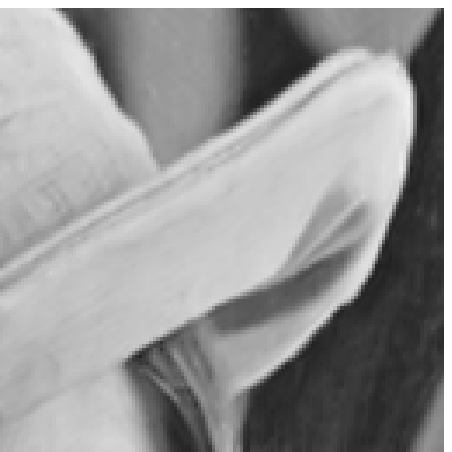}   \\
\hspace{-2ex}
\scriptsize{\textbf{(e) Block $l_1$: 26.35 dB}} & \scriptsize{\textbf{(f) Block OMP: 29.27 dB} }& \tiny{\textbf{(g) Block weighted $l_1$: 35.94 dB}} & \tiny{\textbf{(h) Block weighted $l_2$: 36.45 dB}}
\end{tabular}
\end{center}
\vspace{-3ex}
\caption{\small Comparison of different estimation methods on super-resolution zooming. (a) The original image cropped from Lena. (b)  The low-resolution image, shown at the same scale by pixel duplication. From (c) to (h) are the super-resolution results obtained with different estimation methods. See text for more details.} 
\label{fig:comparison:l2:vs:l1}
\vspace{-4ex}
\end{figure}

From the sparse coding point of view, the gain of introducing structure in sparse inverse problem estimation as described in Section~\ref{sec:MAP:EM:sparsity} is now shown through some experiments. An overcomplete dictionary $\bD$ composed of a family of PCA bases $\{\bB_k\}_{1 \leq k \leq K}$, illustrated in Figure~\ref{fig:dict}-(b), is learned as described in Section~\ref{sec:SSMS}, and is then fed to the following estimation schemes. (i) \textbf{Global $l_1$ and OMP}: the ensemble of $\bD$ is used as an overcomplete dictionary, and the zooming estimation is calculated with the sparse estimate~\eqref{eq:sparse:le} through, respectively, an $l_1$ minimization or an orthogonal matching pursuit (OMP). 
(ii) \textbf{Block $l_1$ and OMP}: the sparse estimate is calculated in each PCA basis $\bB_k$ through, respectively an $l_1$ minimization and an OMP, and the best estimate is selected with a model selection procedure similar to~\eqref{eqn:MAP:model:selection}, thereby reducing the degree of freedom in the estimation with respect to the global $l_1$ and OMP.~\cite{yu2010SSMS}. (iii) \textbf{Block weighted $l_1$}: on top of the block $l_1$, weights are included for each coefficient amplitude in the regularizer,
\begin{equation}
\label{eqn:weighted $l_1$}
\tilde{\ba}_i^{k} = \arg \min_{{\ba}_i}  \left( \|\bU_i \bB_k {\ba}_i - \by_i\|^2 + \sigma^2 \sum_{m=1}^N \frac{|{\ba}_i[m]|}{\tau_m^k} \right),
\end{equation}
with the weights $\tau_m^k = {(\lambda_m^k)^{1/2}}$, where $\lambda_m^k$ are the eigenvalues of the $k$-th PCA basis. The weighting scheme penalizes the atoms that are less likely to be important, following the spirit of the weighted $l_2$ deduced from the MAP estimate. 
(iv) \textbf{Block weighted $l_2$}: the proposed PLE. Comparing with~\eqref{eqn:weighted $l_1$}, the difference is that the weighted $l_2$~\eqref{eqn:MAP:estimate:gaussian:PCA} takes the place of the weighted $l_1$, thereby transforming the problem into a stable and computationally efficient piecewise linear estimation. 

The comparison on a typical region of Lena in the $2 \times 2$ image zooming context is shown in Figure~\ref{fig:comparison:l2:vs:l1}. The global $l_1$ and OMP produce some clear artifacts along the contours, which degrade the PSNRs. The block $l_1$ or OMP considerably improves the results (especially for $l_1$). Comparing with the block $l_1$ or OMP, a very significant improvement is achieved by adding the collaborative weights on top of the block $l_1$. The proposed PLE with the block weighted $l_2$, computed with linear filtering, further improves the estimation accuracy over the block weighted $l_1$, with a much lower computational cost. 

\begin{figure}[htbp]
\vspace{-2ex}
\begin{center}
\begin{tabular}{cccccccccc}
\hspace{-5ex}\epsfxsize=1.7cm\epsffile{./figures/Lena.eps} &
\hspace{-1.7ex}\epsfxsize=1.7cm \epsffile{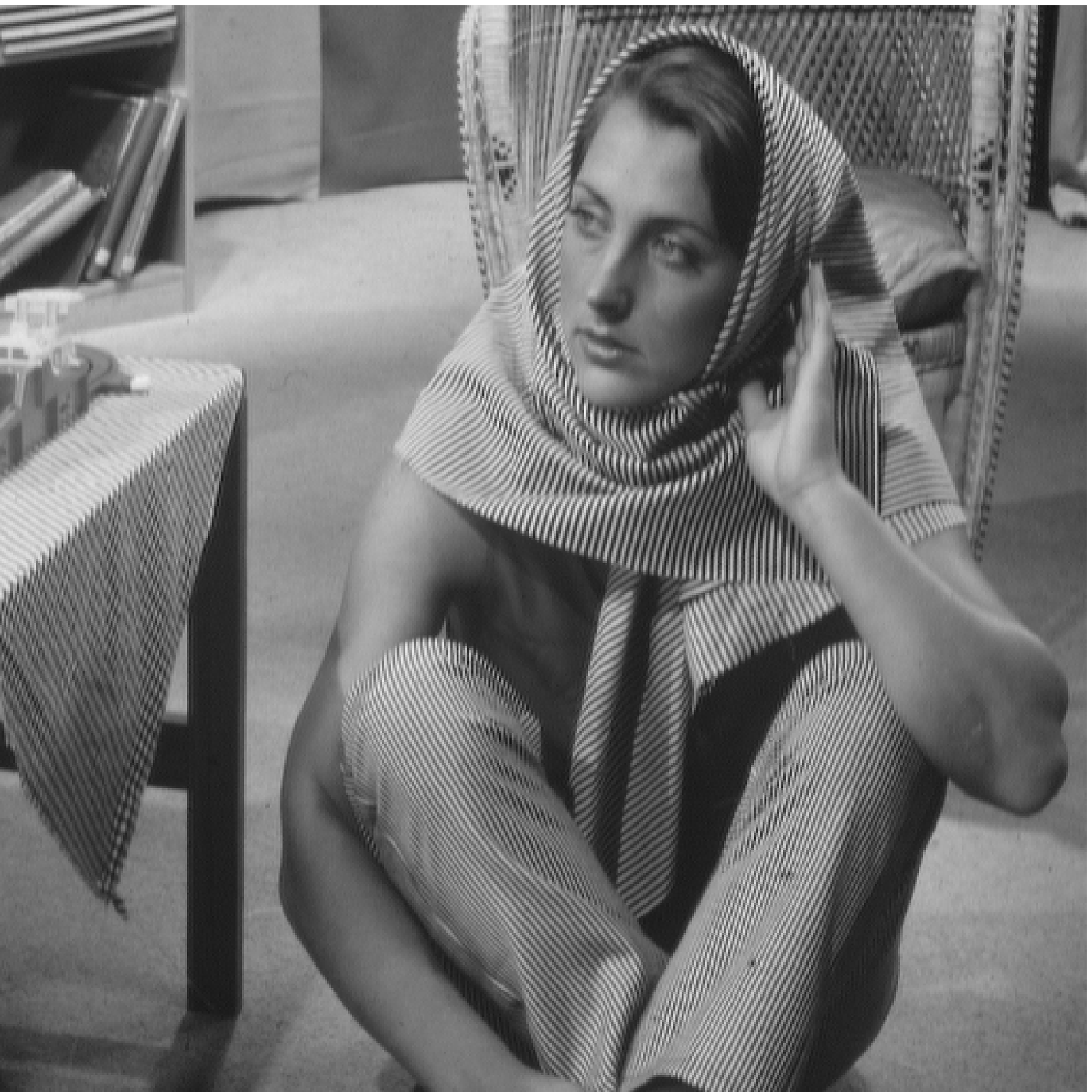} &
\hspace{-1.7ex}\epsfxsize=1.7cm \epsffile{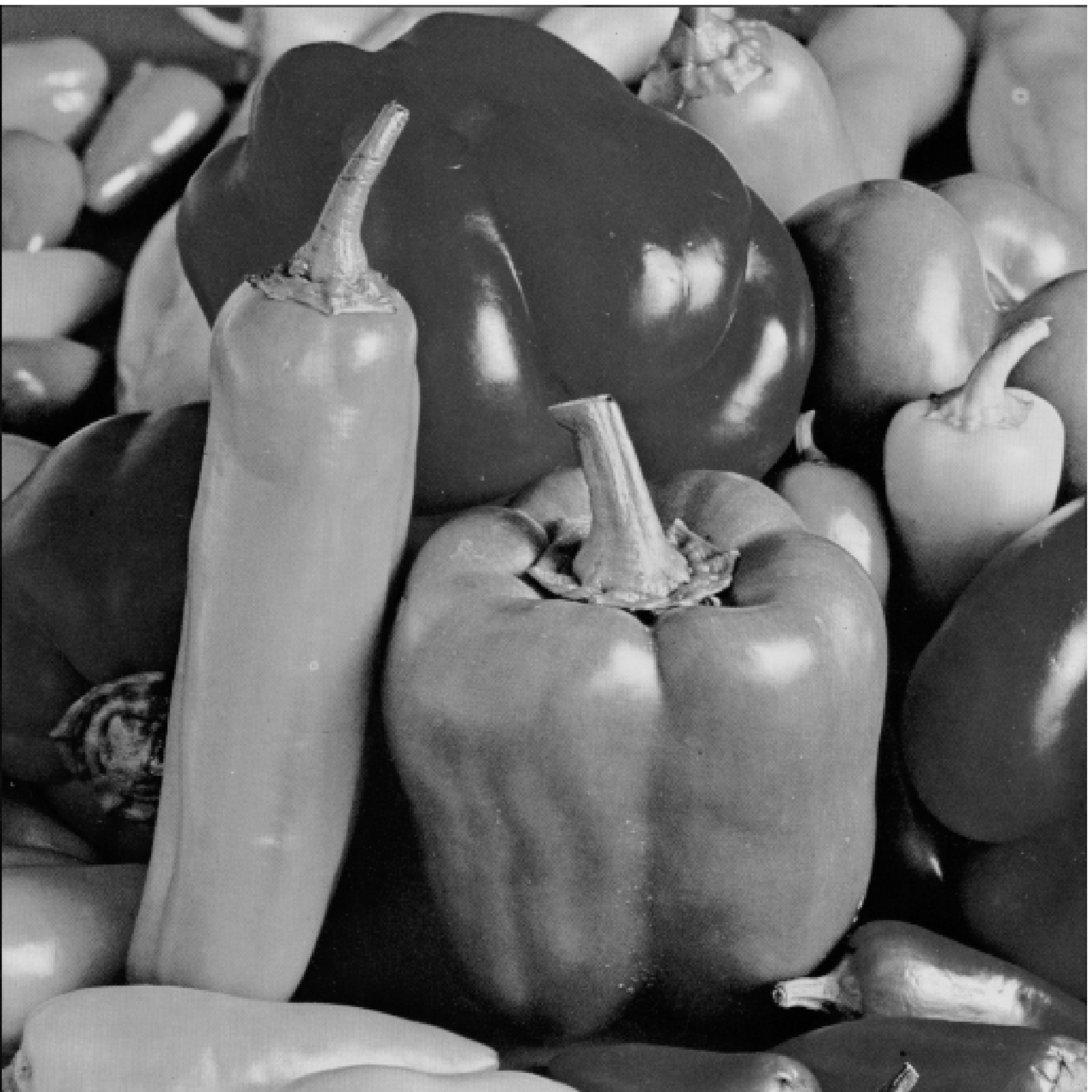} &
\hspace{-1.7ex}\epsfxsize=1.7cm \epsffile{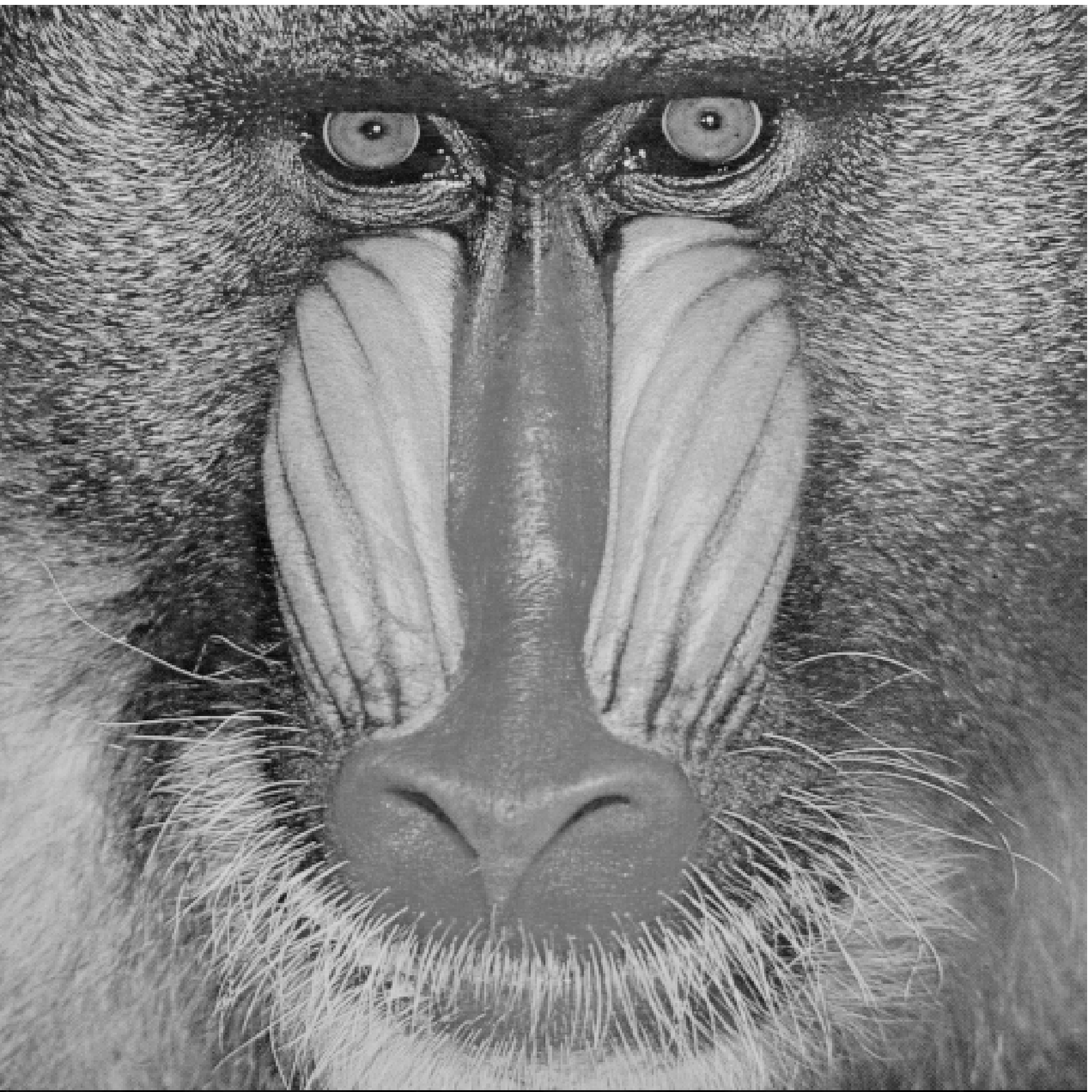} &
\hspace{-1.7ex}\epsfxsize=1.7cm\epsffile{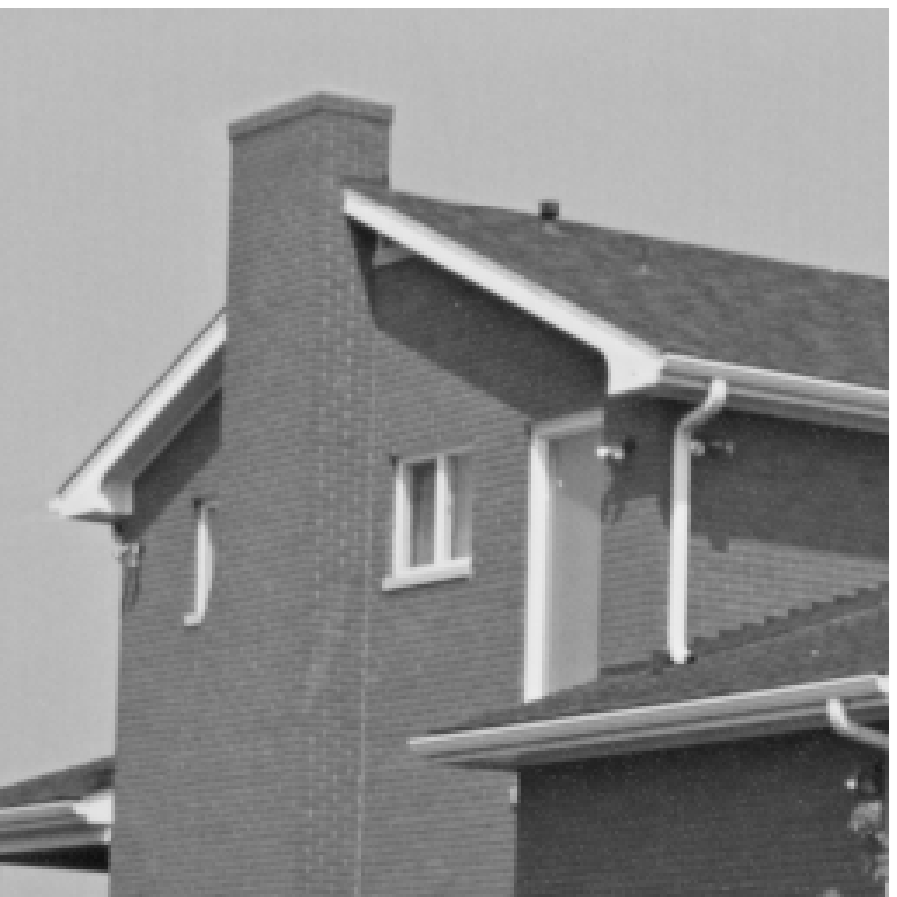} &
\hspace{-1.7ex}\epsfxsize=1.7cm\epsffile{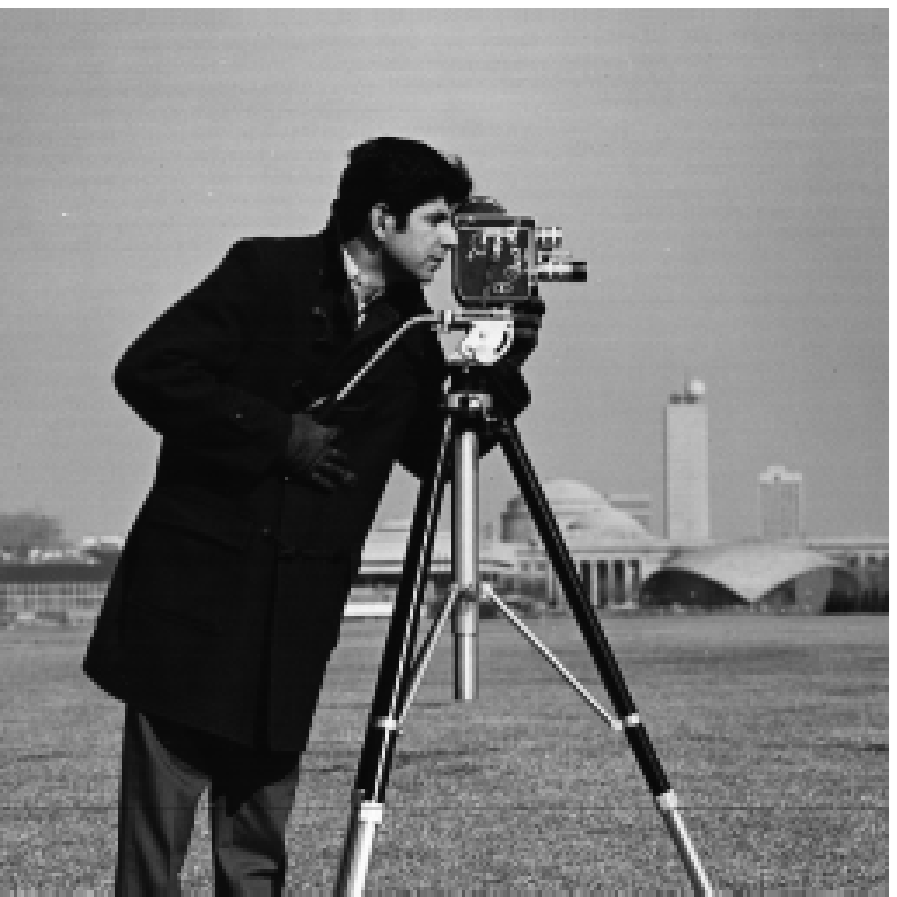} &
\hspace{-1.7ex}\epsfxsize=1.7cm \epsffile{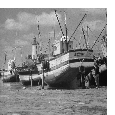}  &
\hspace{-1.7ex}\epsfxsize=1.7cm \epsffile{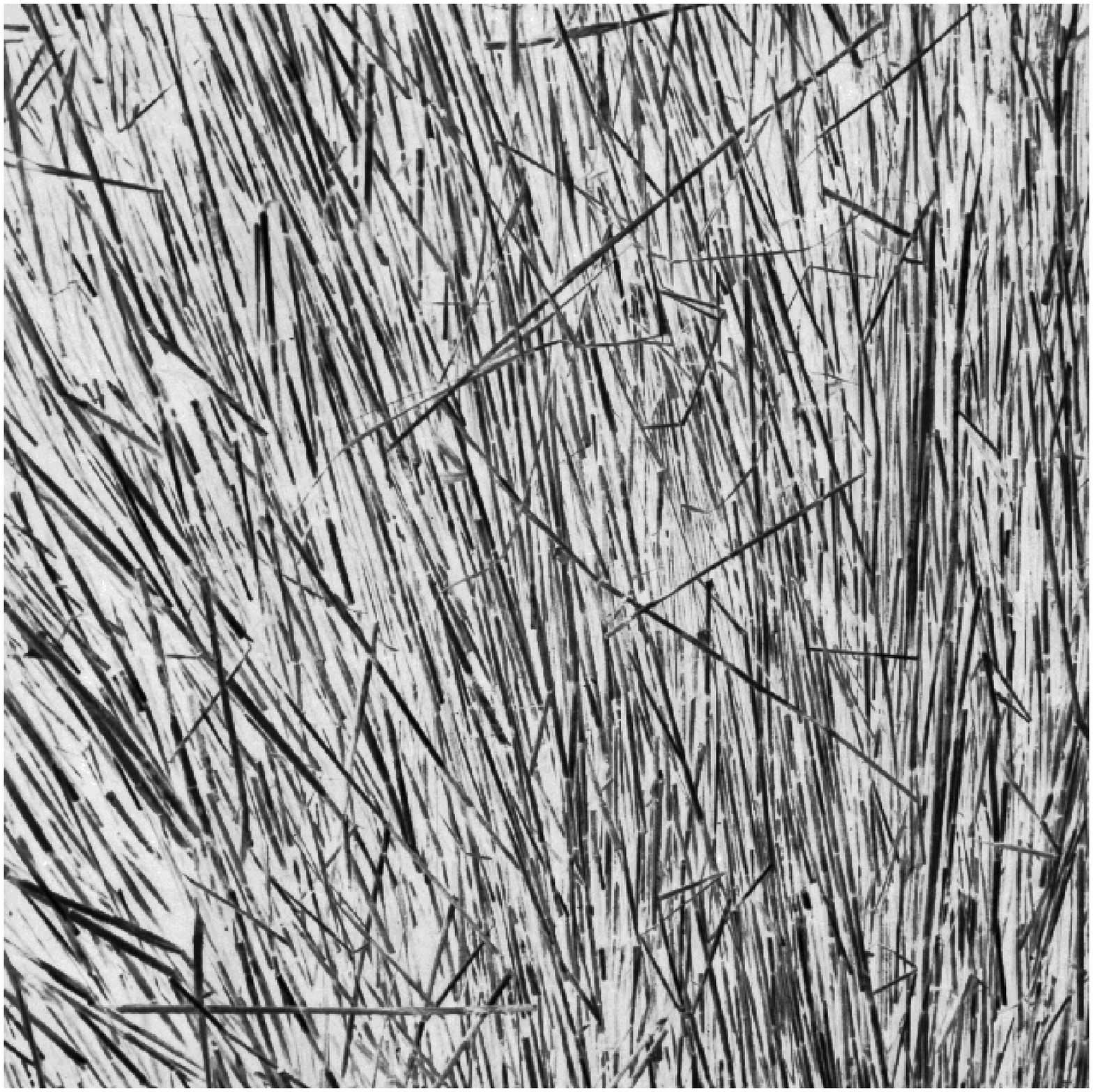} &
\hspace{-1.7ex}\epsfxsize=1.7cm\epsffile{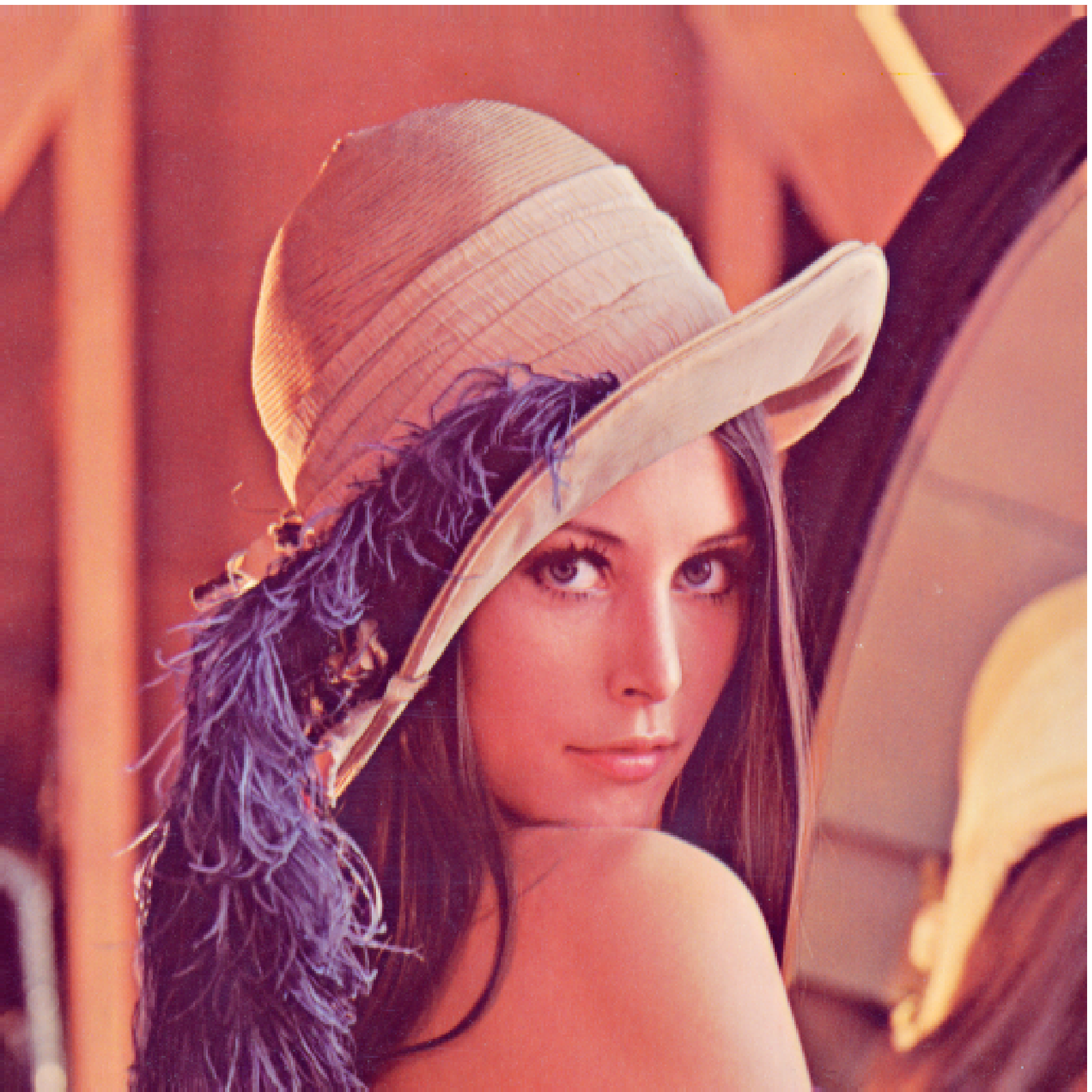} &
\hspace{-1.7ex}\epsfxsize=1.7cm\epsffile{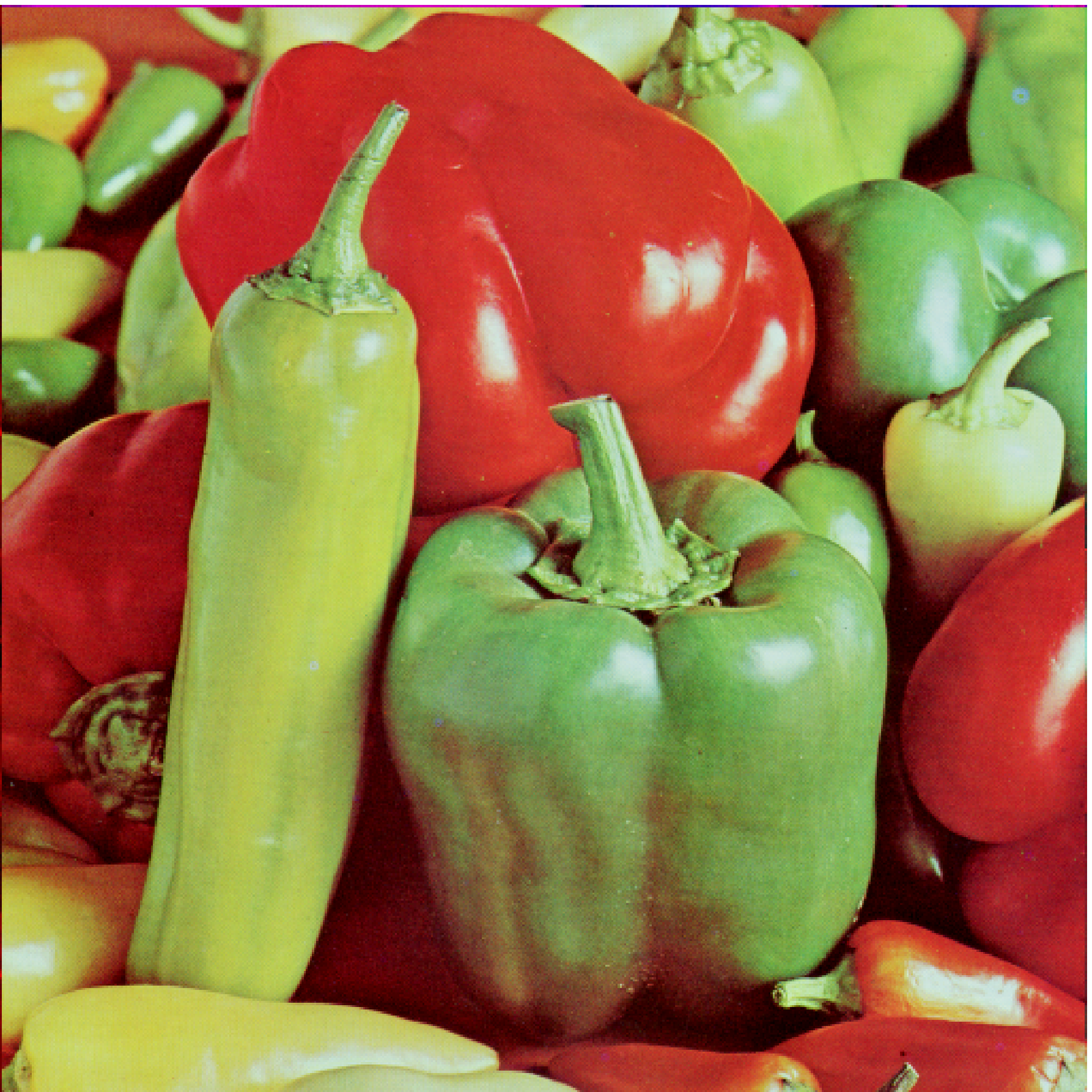} \\
\hspace{-5ex}\epsfxsize=1.7cm \epsffile{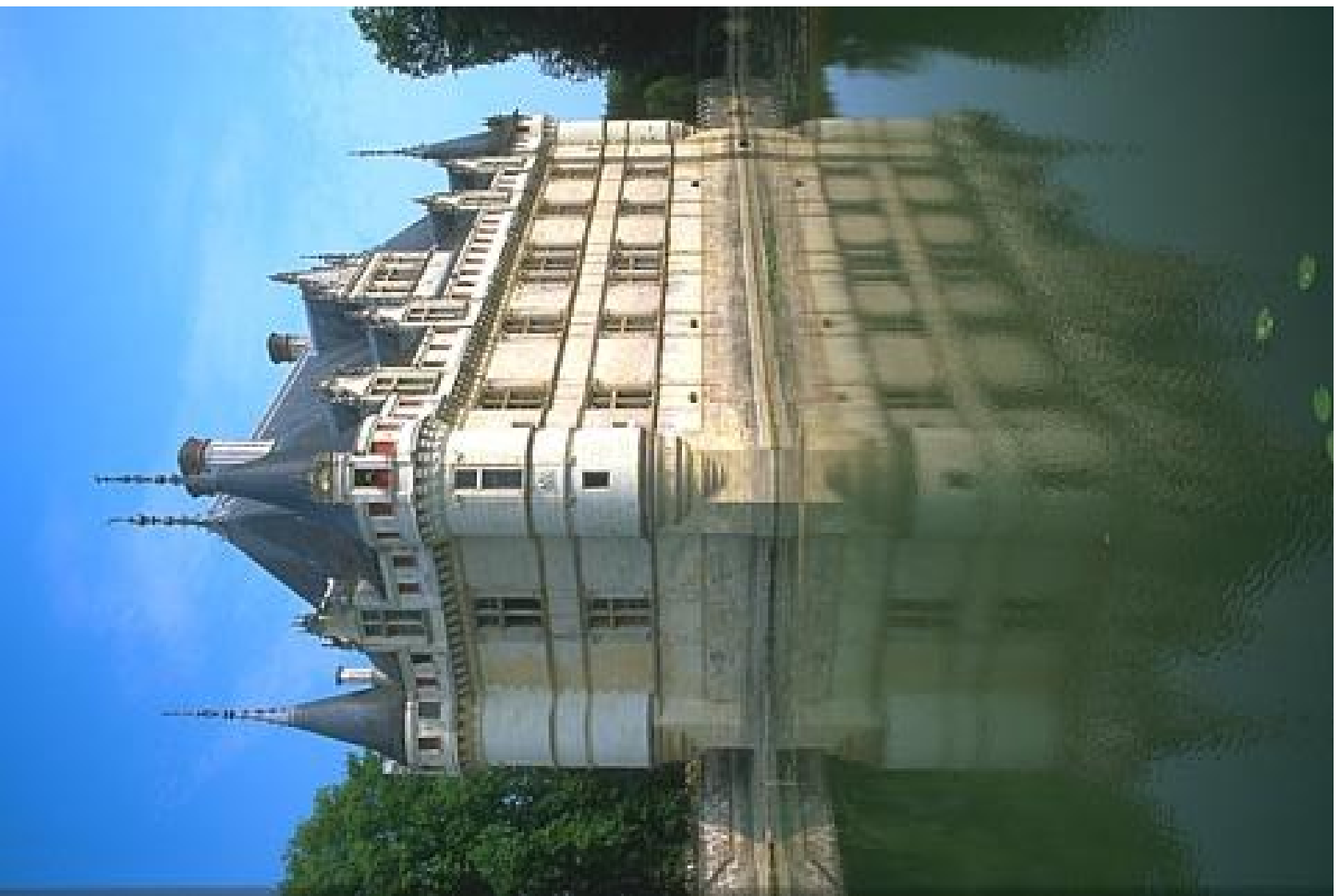} &
\hspace{-1.7ex}\epsfxsize=1.7cm \epsffile{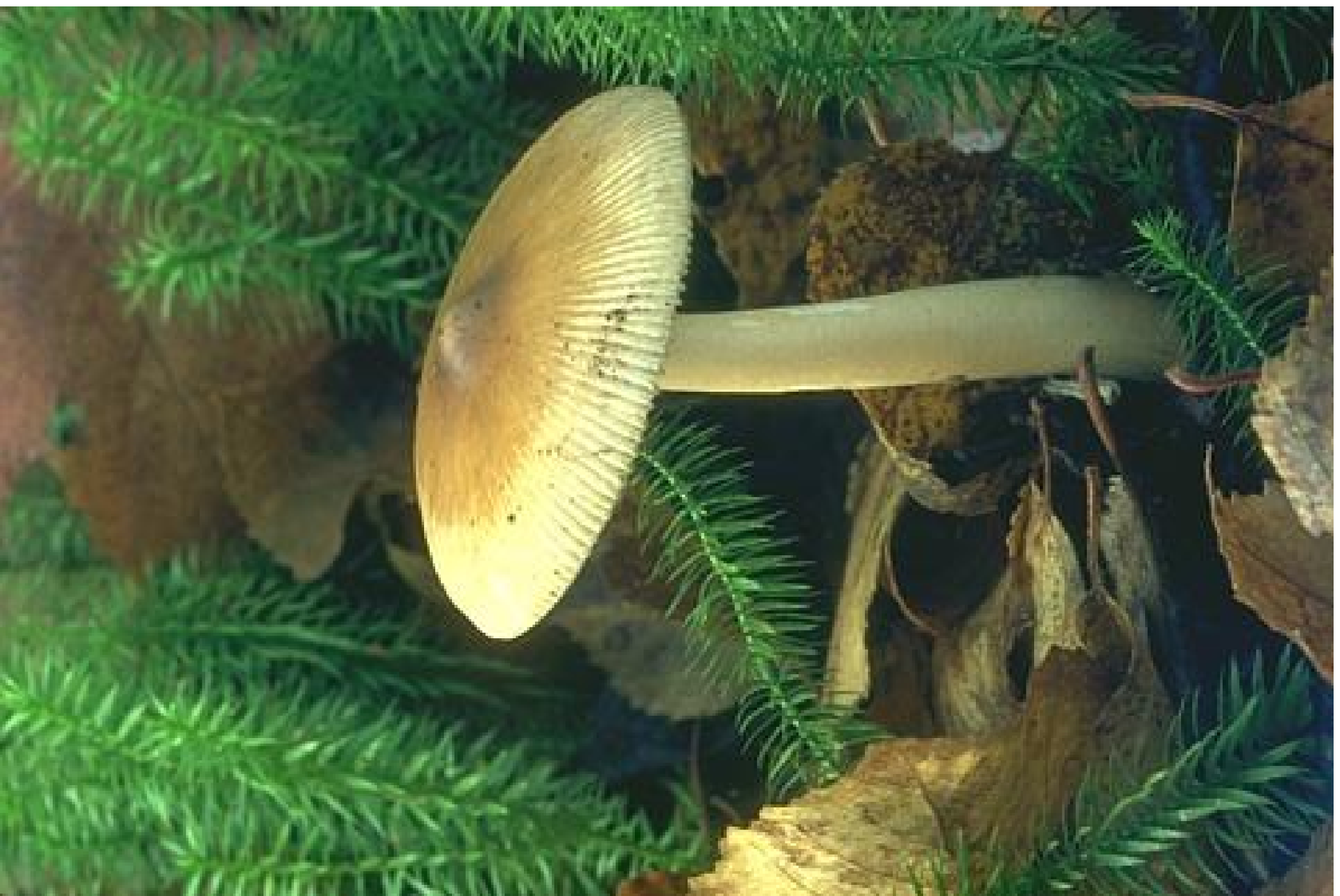} &
\hspace{-1.7ex}\epsfxsize=1.7cm\epsffile{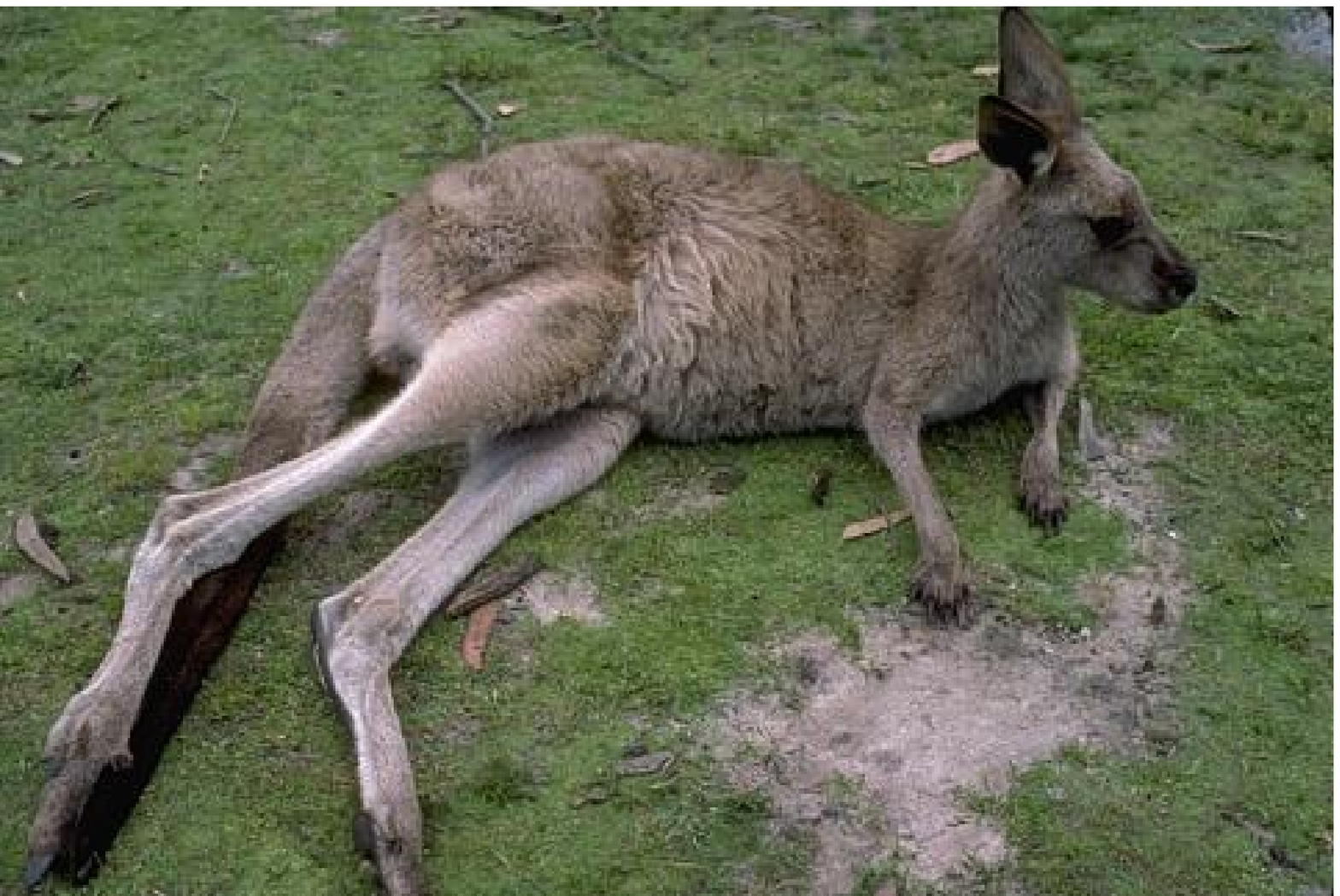} &
\hspace{-1.7ex}\epsfxsize=1.7cm\epsffile{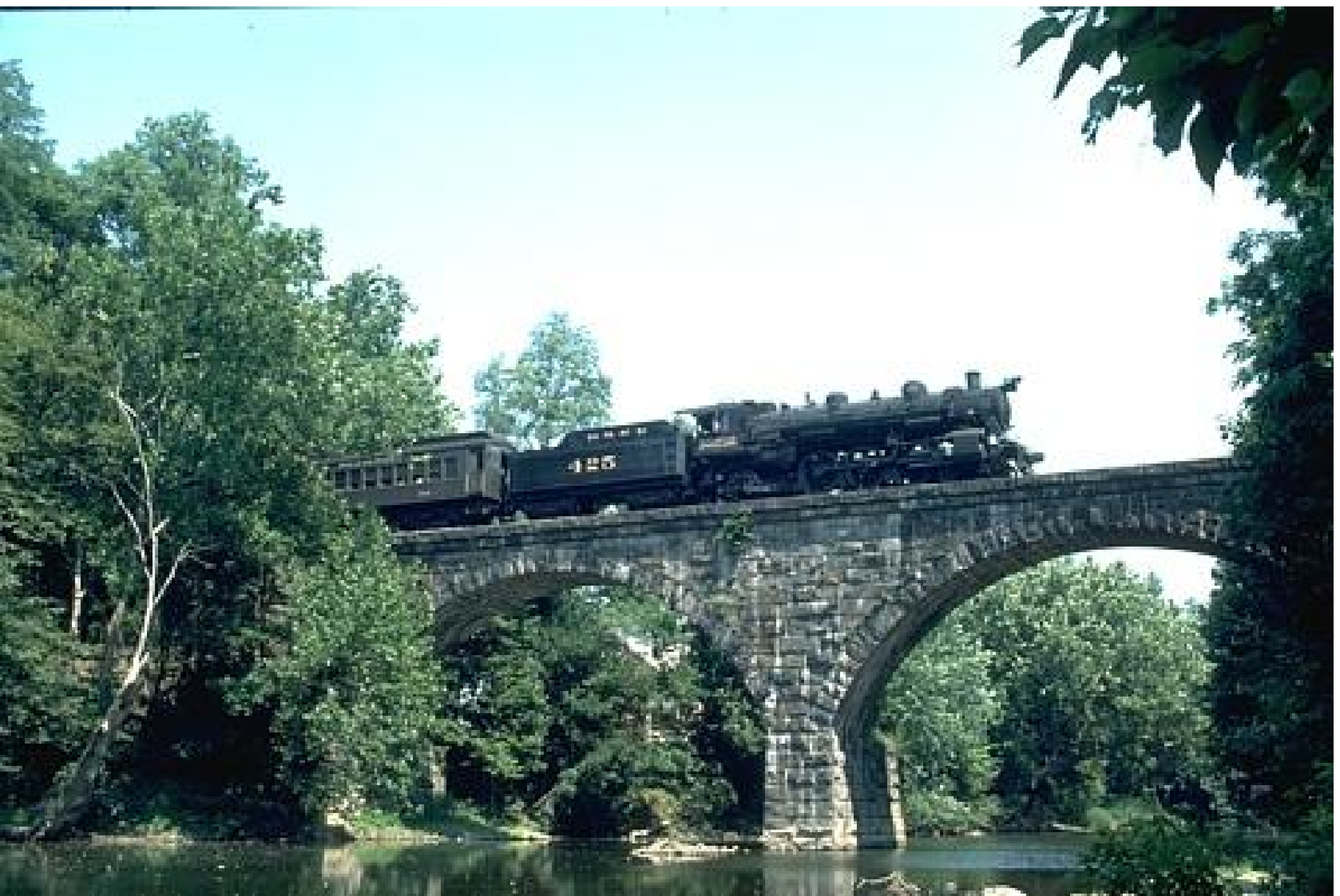} &
\hspace{-1.7ex}\epsfxsize=1.7cm \epsffile{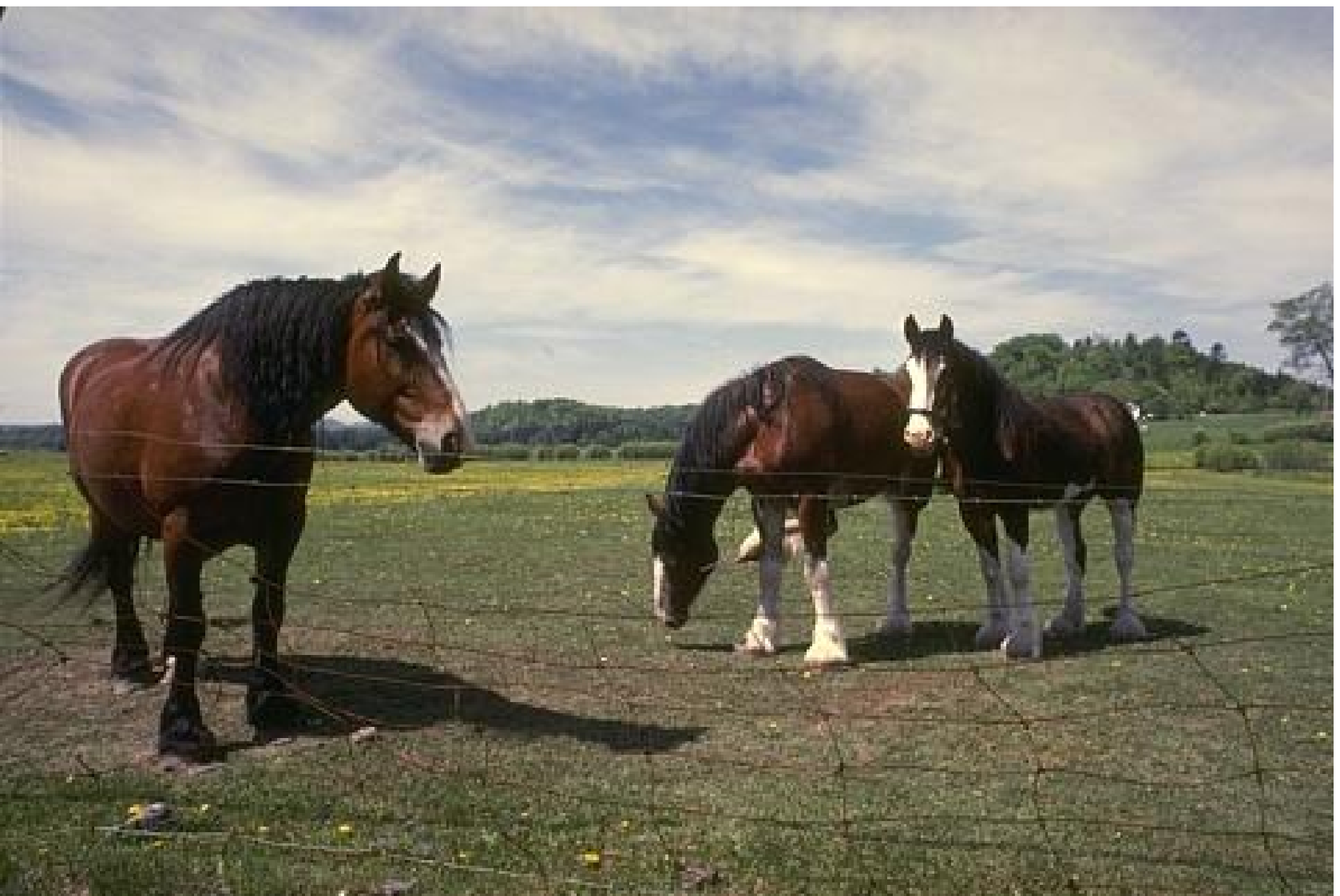} &
\hspace{-1.7ex}\epsfxsize=1.7cm \epsffile{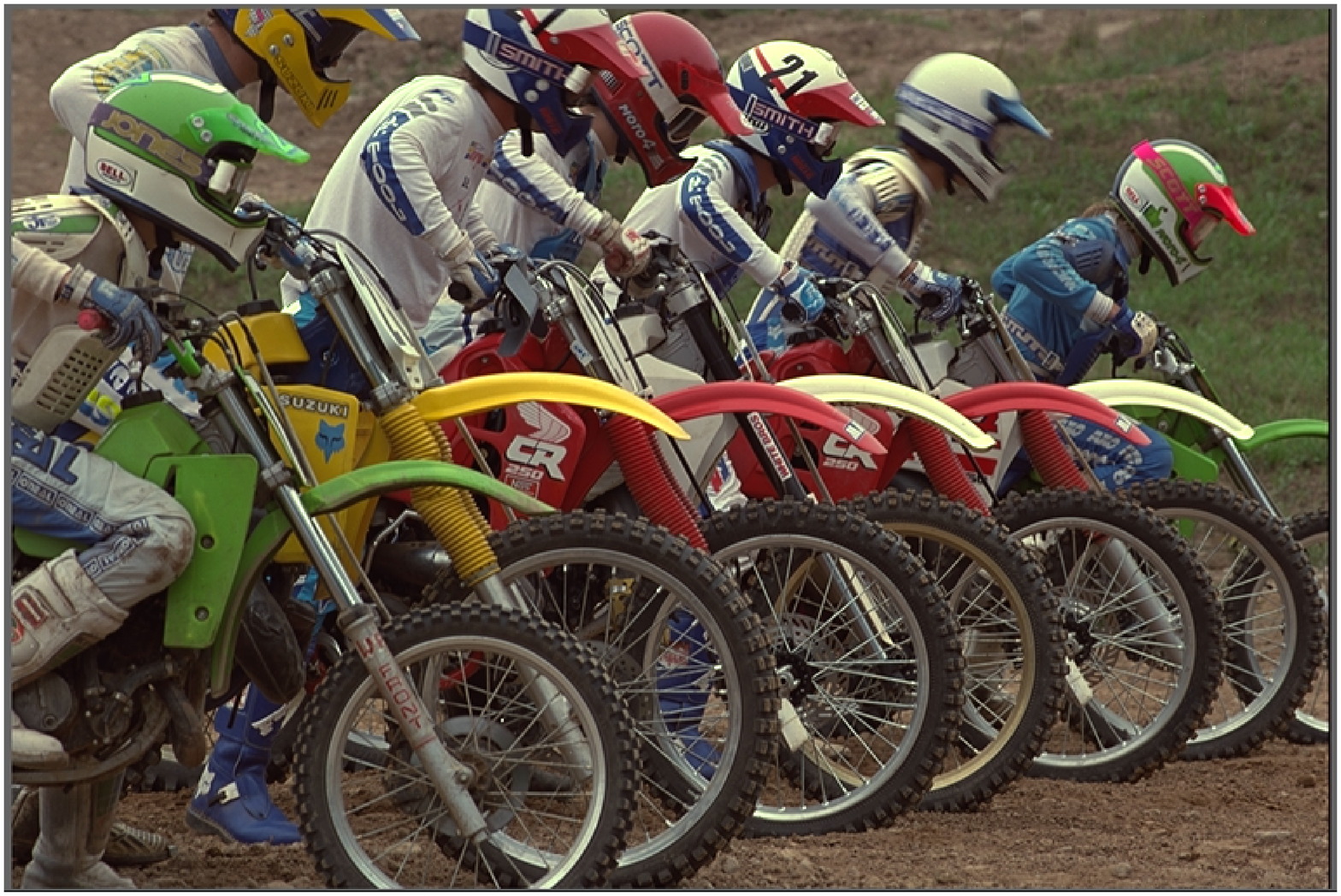} &
\hspace{-1.7ex}\epsfxsize=1.7cm\epsffile{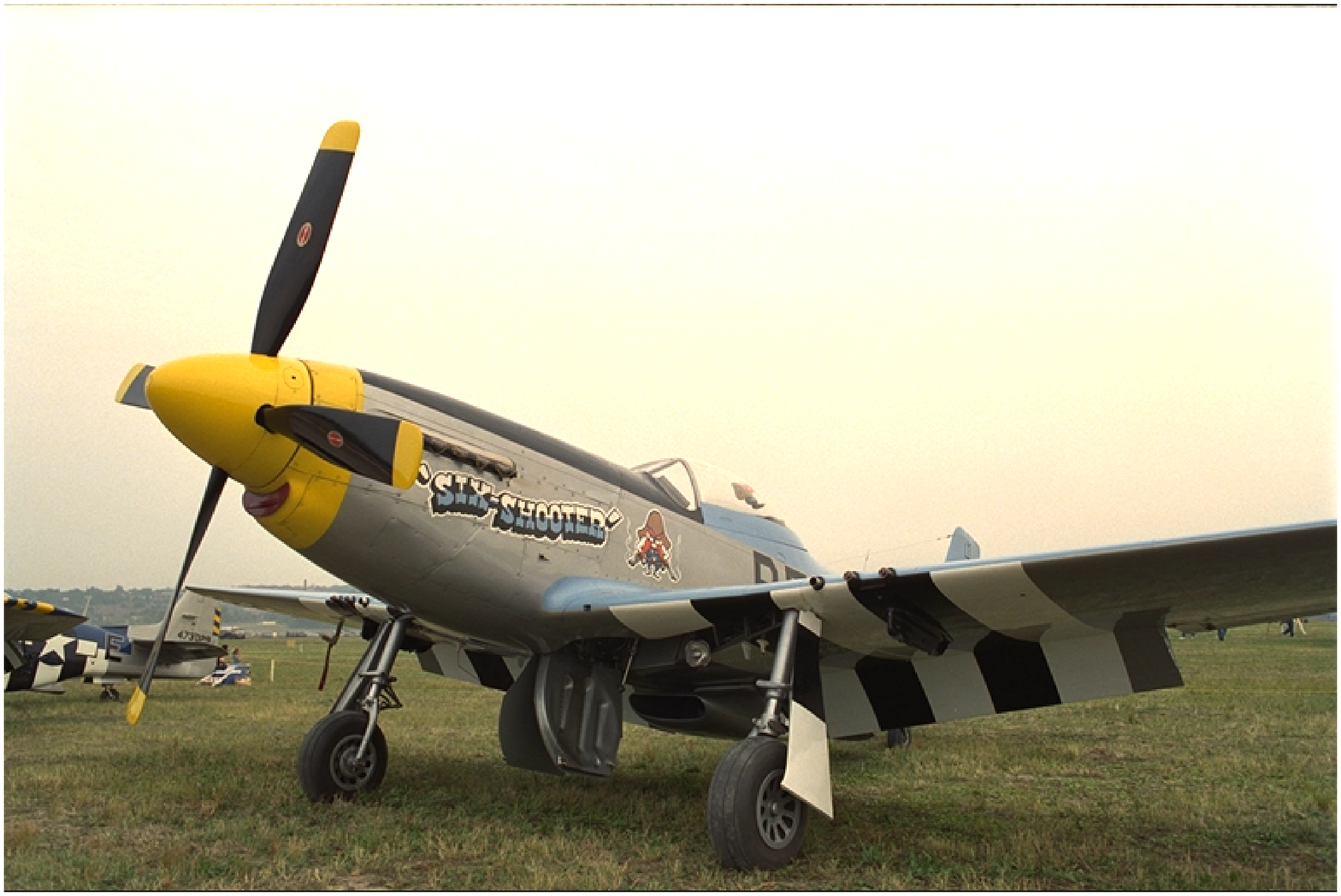} &
\hspace{-1.7ex}\epsfxsize=1.1cm \epsffile{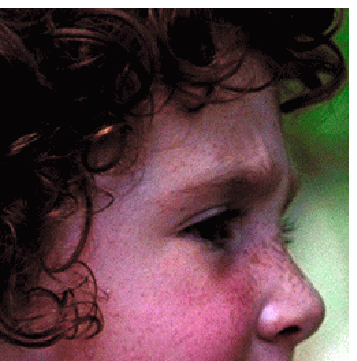} &
\hspace{-1.7ex}\epsfxsize=1.7cm \epsffile{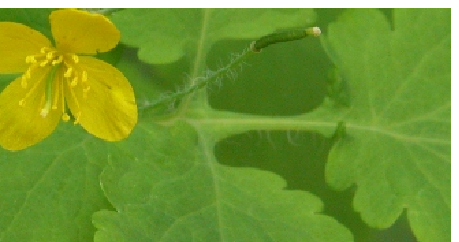}  
\end{tabular}
\end{center}
\vspace{-3ex}
\caption{\small Images used in the numerical experiments. From top to
bottom, left to right. The first eight are gray-level images:  Lena ($512 \times 512$), Barbara ($512 \times 512$), Peppers ($512 \times 512$), Mandril ($512 \times 512$), House ($256 \times 256$),  Cameraman ($256 \times 256$), Boats($512 \times 512$), and Straws ($640 \times 640$). The rest are color images: Castle ($481 \times 321$), Mushroom ($481 \times 321$), Kangaroo ($321 \times 481$), Train ($321 \times 481$), Horses ($321 \times 481$), Kodak05 ($512 \times 768$), Kodak20 ($512 \times 768$), Girl ($258 \times 255$), and Flower ($171 \times 330$).} \label{fig:numeric:images}
\vspace{-4ex}
\end{figure}

In the following sections, PLE will be applied to a number of inverse problems, including image inpainting, zooming and deblurring. The experiments are performed on some standard gray-level and color images, illustrated in Figure~\ref{fig:numeric:images}.

\section{Inpainting}
\label{sec:inpainting}

In the addressed case of inpainting, the original image $\bbf$ is masked with a random mask, $\by = \bU \bbf$,
where $\bU$ is a diagonal matrix whose diagonal entries are randomly either $1$ or $0$, keeping or killing the corresponding pixels. Note that this can be considered as a particular case of compressed sensing, or when collectively considering all the image patches, as matrix completion (and as here demonstrated, in contrast with the recent literature on the subject, a single subspace is not sufficient, see also~\cite{zhou2010nonparametric}). 

The experiments are performed on the gray-level images Lena, Barbara, House, and Boat, and the color images Castle, Mushroom, Train and Horses. Uniform random masks that retain $80\%$, $50\%$, $30\%$ and $20\%$ of the pixels are used. The masked images are then 
inpainted with the algorithms under consideration. 

For gray-level images, the image patch size is $\sqrt{N} \times \sqrt{N} =  8 \times 8$  when the available data is $80\%$, $50\%$, and $30\%$. Larger patches of size $12 \times 12$ are used when images are heavily masked with only $20\%$ pixels available. For color images, patches of size $\sqrt{N} \times \sqrt{N} \times 3$ throughout the RGB color channels are used to exploit the redundancy among the channels~\cite{mairal2008sparse}. To simplify the initialization in color image processing,  the E-step in the first iteration is calculated with ``gray-level'' patches of size $\sqrt{N} \times \sqrt{N}$ on each channel, but with a unified model selection across the channels: The same model selection is performed throughout the channels by minimizing the sum of the model selection energy~\eqref{eqn:MAP:model:selection} over all the channels; the signal estimation is calculated in each channel separately. The M-step then estimates the Gaussian models with the ``color'' patches of size $\sqrt{N} \times \sqrt{N} \times 3$ based on the model selection and the signal estimate previously obtained in the E-step. Starting from the second iteration, both the E- and M-steps are calculated with ``color'' patches, treating the $\sqrt{N} \times \sqrt{N} \times 3$ patches as vectors of size $3N$. $\sqrt{N}$ is set to 6 for color images, as in the previous works~\cite{mairal2008sparse,zhou2010nonparametric}. The MAP-EM algorithm runs for 5 iterations. The noise standard deviation $\sigma$ is set to 3, which corresponds to the typical noise level in these images. The small constant $\epsilon$ in the covariance regularization is set to 30 in all the experiments.

The PLE inpainting is compared with a number of recent methods, including ``MCA'' (morphological component analysis)~\cite{elad2005simultaneous}, ``ASR'' (adaptive sparse reconstructions)~\cite{guleryuz2006nonlinear} ,  ``ECM'' (expectation conditional maximization)~\cite{fadili2009inpainting} , ``KR'' (kernel regression)~\cite{takeda2007kernel}, ``FOE'' (fields of experts)~\cite{roth2009fields},  ``BP'' (beta process)~\cite{zhou2010nonparametric}, and ``K-SVD''~\cite{mairal2008sparse}. MCA and ECM compute the sparse inverse problem estimate in a dictionary that combines a curvelet frame~\cite{candes2004new}, a wavelet frame~\cite{mallat2008wts} and a local DCT basis. ASR calculates the sparse estimate with a local DCT. BP infers a nonparametric Bayesian model from the image under test (noise level is automatically estimated). Using a natural image training set, FOE and K-SVD learn respectively a Markov random field model and an overcomplete dictionary that gives sparse representation for the images. The results of MCA, ECM, KR, FOE are generated by the original authors' softwares, with the parameters manually optimized, and those of ASR are calculated with our own implementation. The PSNRs of BP and K-SVD are cited from the corresponding papers. K-SVD and BP currently generate the best inpainting results in the literature. 

Table~\ref{tab:inpainting:gray}-left gives the inpainting results on gray-level images. PLE considerably outperforms the other methods in all the cases,  with a PSNR improvement of about 2 dB on average over the second best algorithms (BP, FOE and MCA). With $20\%$ available data on Barbara, which is rich in textures, it gains as much as about 3 dB over MCA, 4 dB over ECM and 6 dB over all the other methods. Let us remark that when the missing data ratio is high, MCA generates quite good results, as it benefits from the curvelet atoms that have large support relatively to the local patches used by the other methods. 

\begin{figure}[htbp]
\vspace{-2ex}
\begin{center}
\begin{tabular}{cccc}
\hspace{0ex}\epsfxsize=3.5cm\epsffile{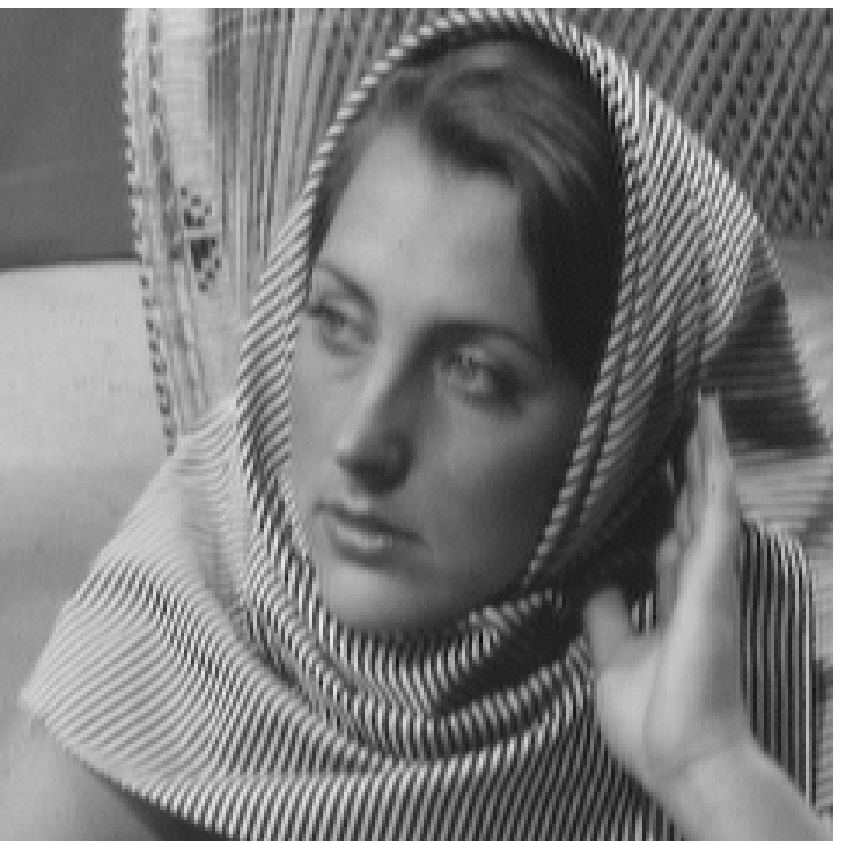} &
\hspace{0ex}\epsfxsize=3.5cm \epsffile{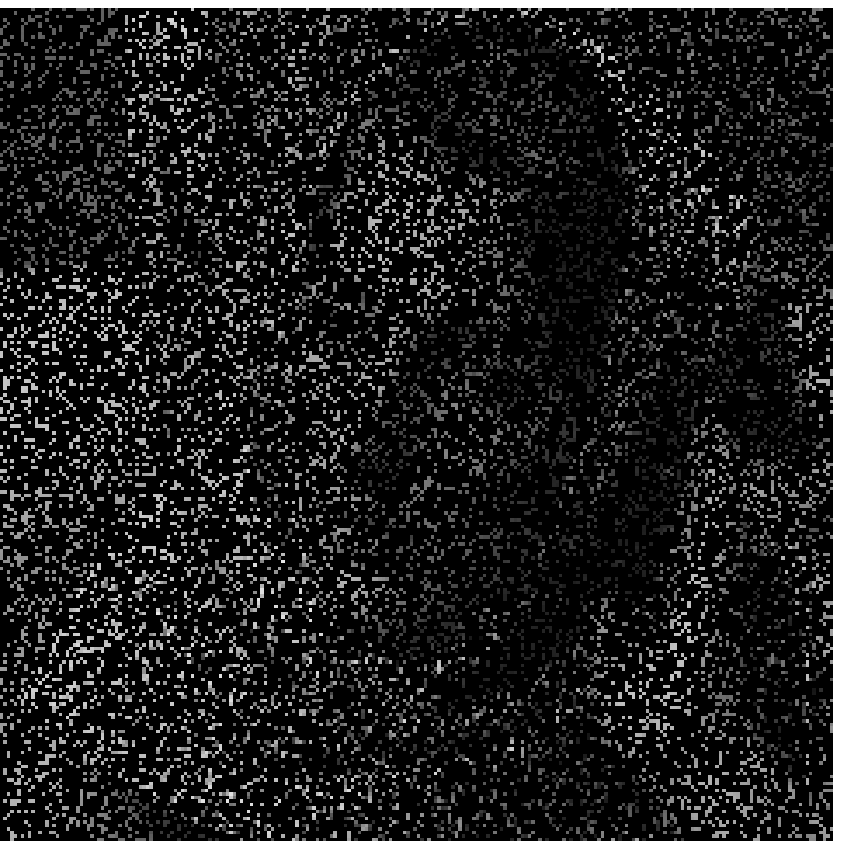} &
\hspace{0ex}\epsfxsize=3.5cm\epsffile{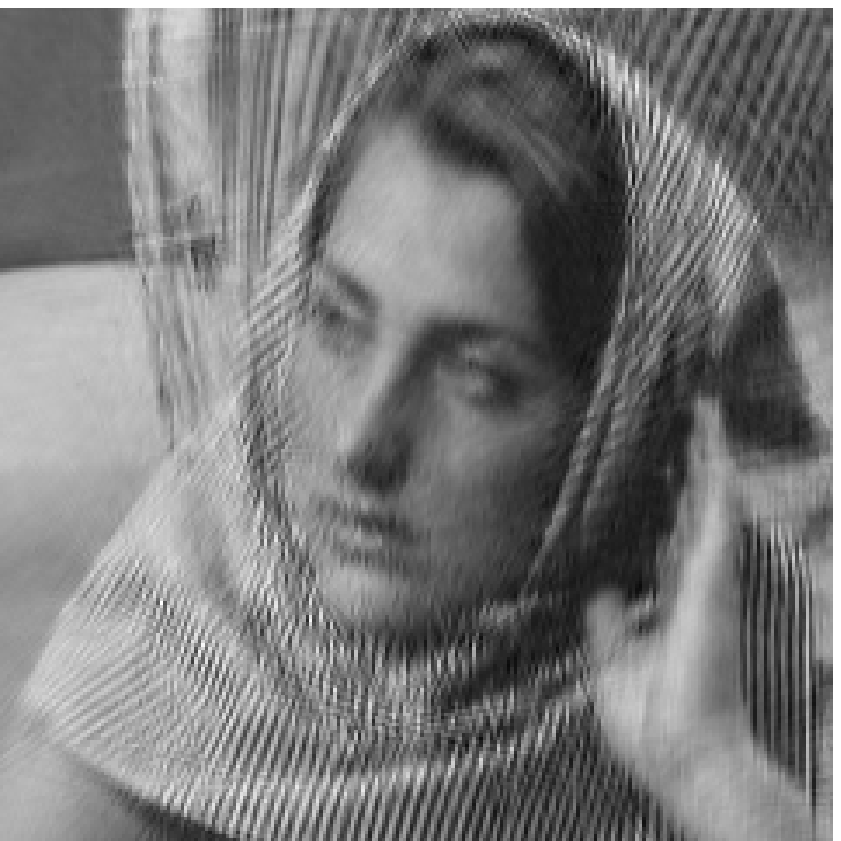}  & 
\hspace{0ex}\epsfxsize=3.5cm \epsffile{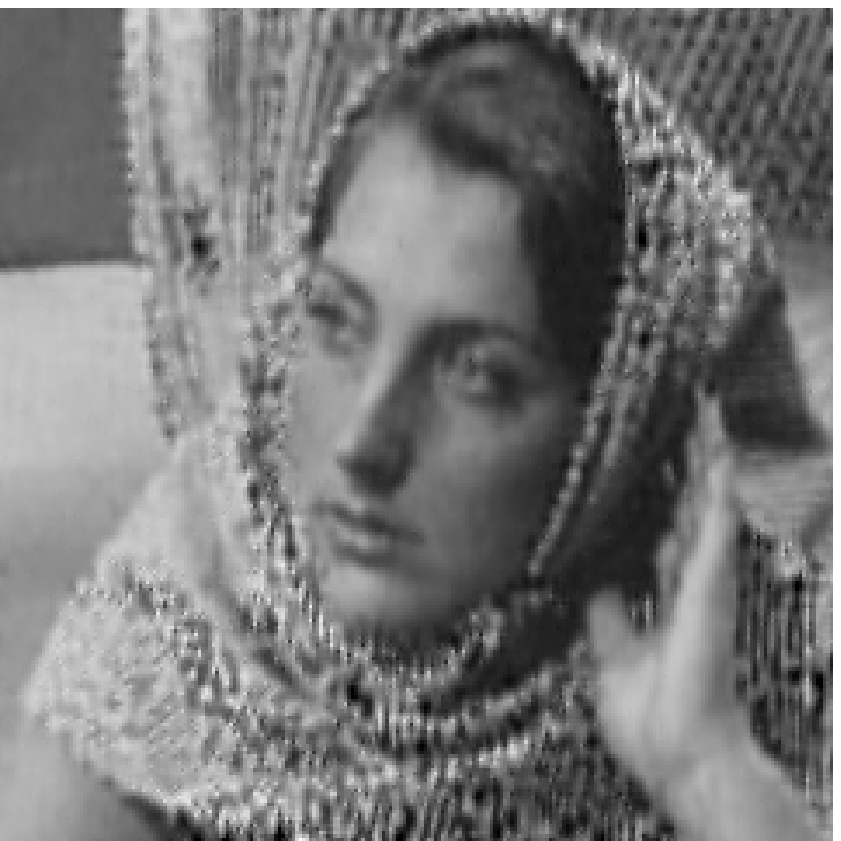} \\
\hspace{0ex} \small{\textbf{(a) Original}} & \small{\textbf{(b) Masked}} & \small{\textbf{(c) MCA (24.18 dB)}} & \small{\textbf{(d) ASR (21.84 dB)}} \vspace{1ex}\\
\hspace{0ex}\epsfxsize=3.5cm\epsffile{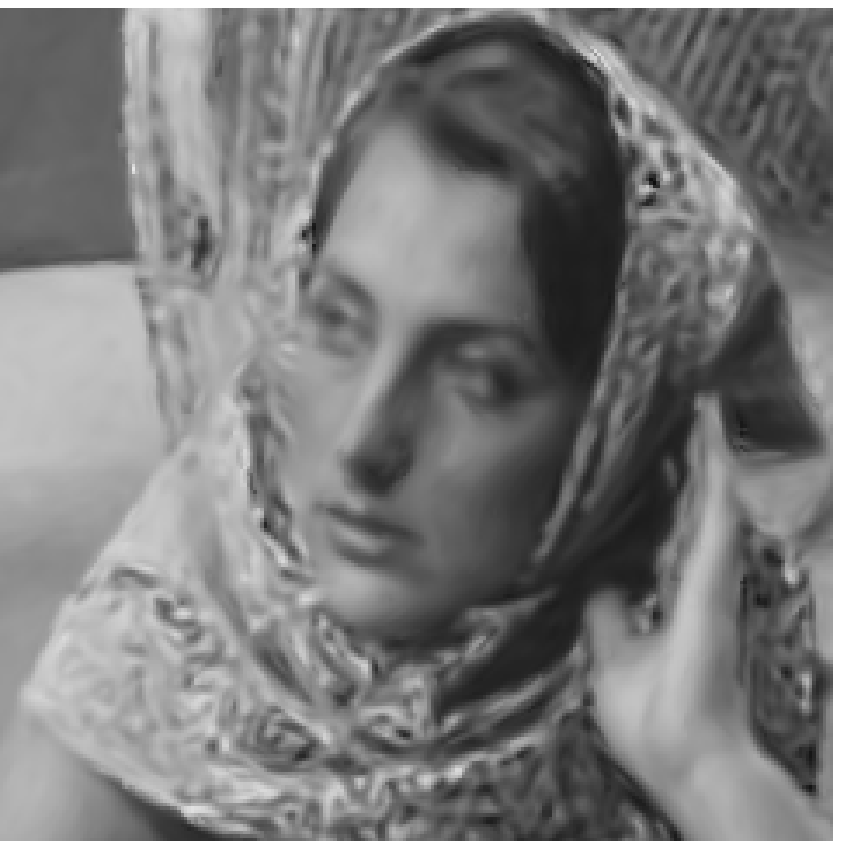} &
\hspace{0ex}\epsfxsize=3.5cm \epsffile{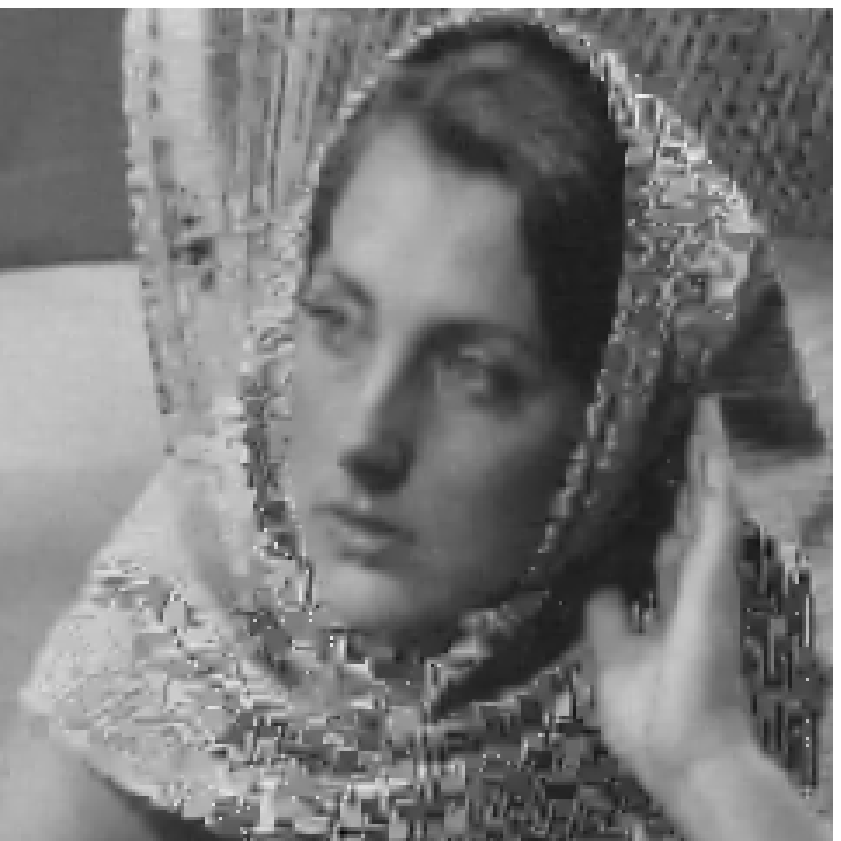} &
\hspace{0ex}\epsfxsize=3.5cm\epsffile{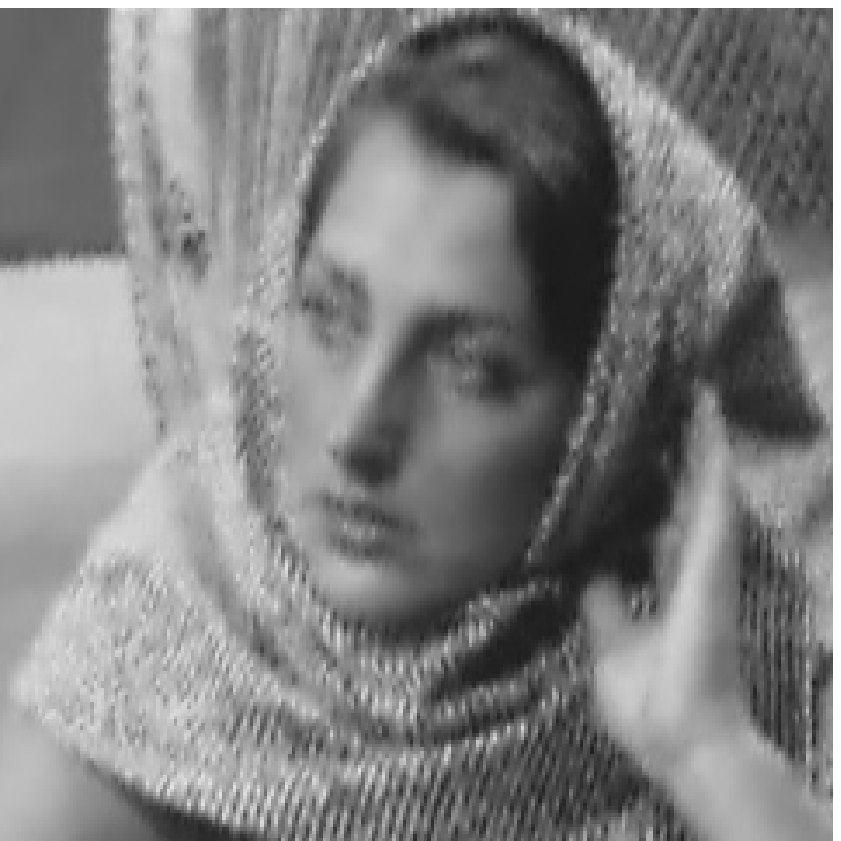} &
\hspace{0ex}\epsfxsize=3.5cm \epsffile{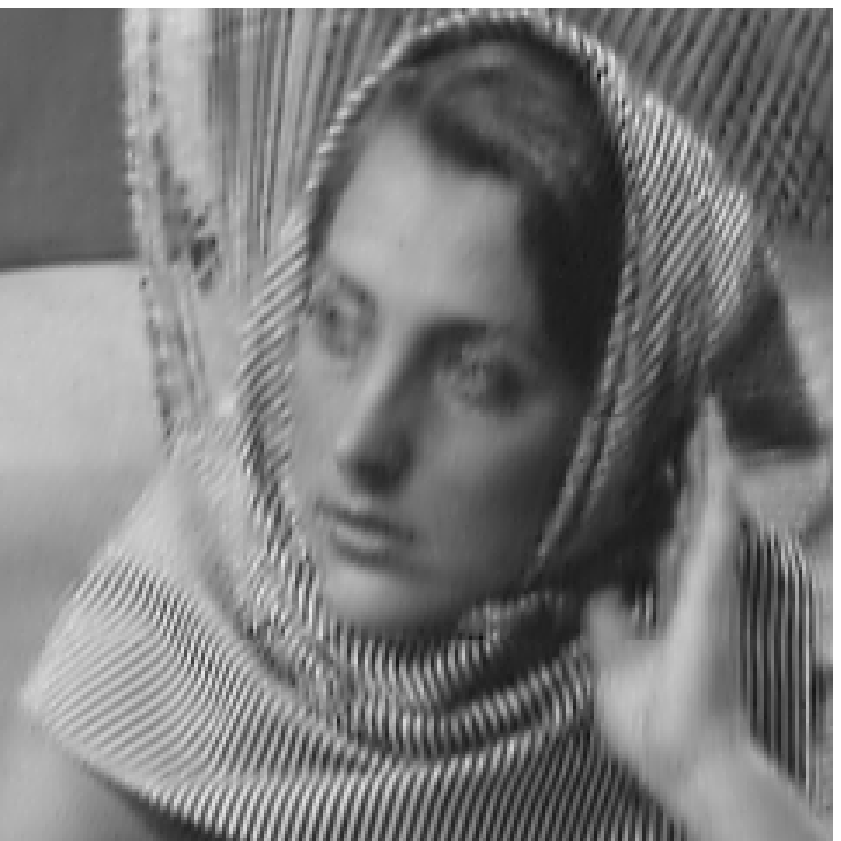}  \\
\hspace{0ex} \small{\textbf{(e) KR (21.55 dB)}} & \small{\textbf{(f) FOE (21.92 dB)}} & \small{\textbf{(g) BP (25.54 dB)}} & \small{\textbf{(h) PLE (27.65 dB)}} \\
\end{tabular}
\end{center}
\vspace{-3ex}
\caption{\small Gray-level image inpainting. (a) Original image cropped from Barbara. (b) Masked image with $20\%$ available data (6.81 dB). From (c) to (g): Image inpainted by different algorithms. Note the overall superior visual quality obtained with the proposed approach. The PSNRs are calculated on the cropped images.} \label{fig:inpainting:gray:Barbara}
\vspace{-4ex}
\end{figure}

Figure~\ref{fig:inpainting:gray:Barbara} compares the results of different algorithms. All the methods lead to good inpainting results on the smooth regions. MCA and KR are good at capturing contour structures. However, when the curvelet atoms are not correctly selected, MCA produces noticeable elongated curvelet-like artifacts that degrade the visual quality and offset its gain in PSRN (see for example the face of Barbara). MCA restores better textures than BP, ASR, FOE and KR. PLE leads to accurate restoration on both the directional structures and the textures, producing the best visual quality with the highest PSNRs.  An additional PLE inpainting examples is shown in Figure~\ref{fig:evolution:Barb}.

\begin{table}[htbp]
\vspace{-2ex}
\begin{center}
{\small
\begin{tabular}{cc}
\hspace{-5ex}
\begin{tabular}{|c|c||c|c|c|c|c|c|c|}
\hline
\multicolumn{2}{|c||}{\textit{Data ratio}}  & MCA & ASR & ECM & KR & FOE* & BP & PLE \\
\hline
\multirow{4}{*} {Lena}  
& $80\%$ & 40.60 & 42.18 & 39.51 &  41.68 & 42.17 & 41.27 & \textbf{43.38} \\ \cline{2-9}
& $50\%$ & 35.63 & 36.16 & 34.43 &  36.77 & 36.66 & 36.94 & \textbf{37.78} \\ \cline{2-9}
& $30\%$ & 32.33 & 32.48 & 31.11 & 33.55 & 33.22 & 33.31 & \textbf{34.37} \\ \cline{2-9}
& $20\%$ & 30.30 & 30.37 & 28.93 & 31.21 & 31.06 & 31.00 & \textbf{32.22} \\ \hline \hline
\multirow{4}{*} {Barbara}  
& $80\%$ & 41.50 & 39.63 & 39.10 & 37.81 & 38.27 & 40.76 & \textbf{43.85} \\ \cline{2-9}
& $50\%$ & 34.29 & 30.42 & 32.54 & 27.98 & 29.47 &  33.17 & \textbf{37.03} \\ \cline{2-9}
& $30\%$ & 29.98 & 25.72 & 28.46 & 24.00 & 25.36 & 27.52 & \textbf{32.73} \\ \cline{2-9}
& $20\%$ & 27.47 & 24.66 & 26.45 & 23.34 & 23.93 & 24.80 & \textbf{30.94} \\ \hline \hline
\multirow{4}{*} {House}  
& $80\%$ & 42.91 & 43.79 & 40.61 & 42.57 & 44.70 & 43.03 & \textbf{44.77} \\ \cline{2-9}
& $50\%$ & 37.02 & 36.06 & 35.16 & 36.82 & 37.99 & 38.02 & \textbf{38.97} \\ \cline{2-9}
& $30\%$ & 33.41 & 31.86 & 31.46 & 33.62 & 33.86 & 33.14 & \textbf{34.88} \\ \cline{2-9}
& $20\%$ & 30.67 & 29.91 & 28.97 & 31.19 & 31.28 & 30.12 & \textbf{33.05} \\ \hline \hline
\multirow{4}{*} {Boat}  
& $80\%$ & 38.61 & 39.52 & 37.45 & 37.91 & 38.33 & 39.50 & \textbf{40.49} \\ \cline{2-9}
& $50\%$ & 32.77 & 32.84 & 31.84 & 32.70 & 33.22 & 33.78 & \textbf{34.36} \\ \cline{2-9}
& $30\%$ & 29.57 & 29.55 & 28.46 & 29.28 & 29.80 & 30.00 & \textbf{30.77} \\ \cline{2-9}
& $20\%$ & 27.73 & 27.34 & 26.39 & 27.05 & 27.86 & 27.81 & \textbf{28.66} \\  \hline \hline
\multirow{4}{*} {\textit{Average}}  
& $80\%$ & 40.90 & 41.28 & 39.16 & 39.99 & 40.86 & 41.14 & \textbf{43.12} \\ \cline{2-9}
& $50\%$ & 34.93 & 33.87 & 33.49 & 33.56 & 34.33 & 35.47 & \textbf{37.03} \\ \cline{2-9}
& $30\%$ & 31.32 & 29.90 & 29.87 & 30.11 & 30.56 & 30.99 & \textbf{33.18} \\ \cline{2-9}
& $20\%$ & 29.04 & 28.07 & 27.68 & 28.19 & 28.53 & 28.43 & \textbf{31.21} \\ \hline 
\end{tabular}
\hspace{-2ex}
&
\begin{tabular}{|c|c||c|c|}
\hline
\multicolumn{2}{|c||}{\textit{Data ratio}}  & BP & PLE \\
\hline
\multirow{4}{*} {Castle}  
& $80\%$ & 41.51 & \textbf{48.26} \\ \cline{2-4}
& $50\%$ & 36.45 &  \textbf{38.34} \\ \cline{2-4}
& $30\%$ & 32.02 & \textbf{33.01} \\ \cline{2-4}
& $20\%$ & 29.12 & \textbf{30.07} \\ \hline
\hline
\multirow{4}{*} {Mushroom}  
& $80\%$ & 42.56 & \textbf{49.25} \\ \cline{2-4}
& $50\%$ & 38.88 &  \textbf{40.72} \\ \cline{2-4}
& $30\%$ & 34.63 & \textbf{35.36} \\ \cline{2-4}
& $20\%$ & 31.56 & \textbf{32.06} \\ \hline
\hline
\multirow{4}{*} {Train}  
& $80\%$ & 40.73 &   \textbf{44.01} \\ \cline{2-4}
& $50\%$ & 32.00 &  \textbf{32.75} \\ \cline{2-4}
& $30\%$ & 27.00 & \textbf{27.46} \\ \cline{2-4}
& $20\%$ & 24.59 & \textbf{24.73} \\ \hline
\hline
\multirow{4}{*} {Horses}  
& $80\%$ & 41.97 & \textbf{48.83} \\ \cline{2-4}
& $50\%$ & 37.27 & \textbf{38.52} \\ \cline{2-4}
& $30\%$ & 32.52 & \textbf{32.99} \\ \cline{2-4}
& $20\%$ & 29.99 & \textbf{30.26} \\ \hline
\hline
\multirow{4}{*} {\textit{Average}}  
& $80\%$ & 41.69 & \textbf{47.59} \\ \cline{2-4}
& $50\%$ & 36.15 & \textbf{37.58} \\ \cline{2-4}
& $30\%$ & 31.54 & \textbf{32.18} \\ \cline{2-4}
& $20\%$ & 28.81 & \textbf{29.28} \\ \hline
\end{tabular}
\end{tabular}
}
\end{center}
\vspace{-2ex}
\caption{\small PSNR comparison on gray-level 
(left) and color (right) image inpainting. For each image, uniform random masks with four available data ratios are tested. The algorithms under consideration are MCA~\cite{elad2005simultaneous}, ASR~\cite{guleryuz2006nonlinear} ,  ECM~\cite{fadili2009inpainting} , KR~\cite{takeda2007kernel},  FOE~\cite{roth2009fields}, BP~\cite{zhou2010nonparametric}, and the proposed PLE framework. The bottom box shows the average PSNRs given by each method over all the images at each available data ratio. The highest PSNR in each row is in boldface. The algorithms with * use  a training dataset.} 
\label{tab:inpainting:gray}
\vspace{-6ex}
\end{table}

Table~\ref{tab:inpainting:gray}-right compares the PSNRs of the PLE color image inpainting results with those of BP (the only one in the literature that reports the comprehensive comparison in our knowledge). Again, PLE generates higher PSNRs in all the cases. While the gain is especially large, at about 6 dB, when the available data ratio is high (at $80\%$), for the other masking rates, it is mostly between 0.5 and 1 dB. Both methods use only the image under test to learn the dictionaries. 

\begin{figure}[htbp]
\vspace{-2ex}
\begin{center}
\begin{tabular}{ccc}
\hspace{-3ex}\epsfxsize=2.5cm\epsffile{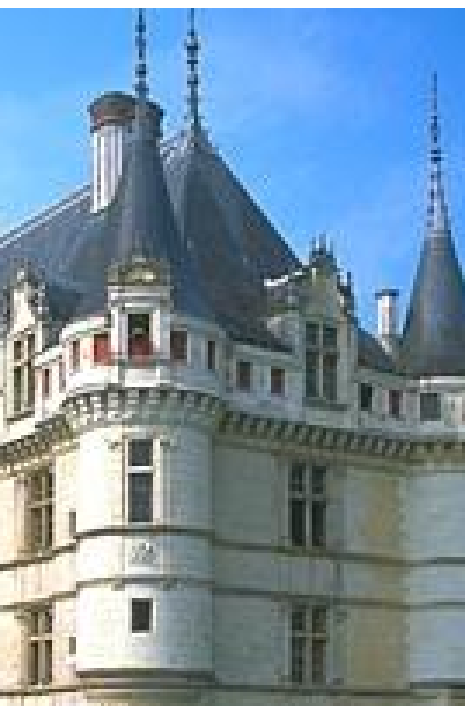} &
\hspace{0ex}\epsfxsize=2.5cm \epsffile{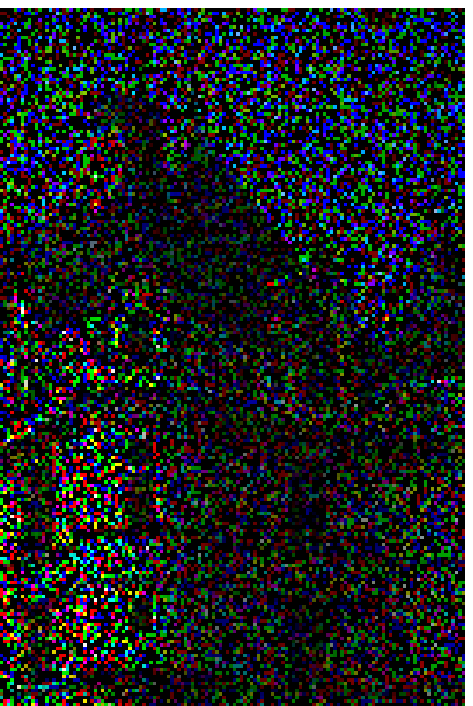} &
\hspace{0ex}\epsfxsize=2.5cm \epsffile{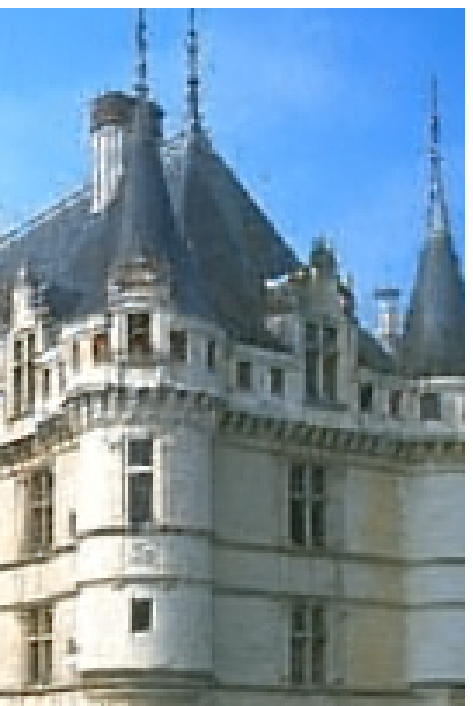}  \\
\hspace{-4ex} \textbf{(a) Original} & \hspace{-2ex} \textbf{(b) Masked} &\hspace{-1ex}  \textbf{(c) PLE}  \\
\end{tabular}
\end{center}
\vspace{-3ex}
\caption{\small Color image inpainting. (a) Original image cropped from Castle. (b) Masked image with $20\%$ available data (5.44 dB). (c) Image inpainted by PLE (27.30 dB). The PSNR on the overall image obtained with PLE is 30.07 dB, higher than the best result reported so far in the literature 29.65 dB~\cite{mairal2008sparse}.} \label{fig:inpainting:color}
\vspace{-5ex}
\end{figure}

Figure~\ref{fig:inpainting:color} illustrates the PLE inpainting result on Castle with $20\%$ available data. Calculated with a much reduced computational complexity, the resulting 30.07 dB PSNR surpasses the highest PSNR, 29.65 dB, reported in the literature, produced by K-SVD~\cite{mairal2008sparse}, that uses a dictionary learned from a natural image training set, followed by 29.12 dB given by BP (BP has been recently improved adding spatial coherence in the code, unpublished results). As shown in the zoomed region, PLE accurately restores  the details of the castle from the heavily masked image. Let us remark that inpainting with random masks on color images is in general more favorable than on gray-level images, thanks to the information redundancy among the color channels. 

\section{Interpolation zooming} 
\label{sec:zooming}

Interpolation zooming is a special case of inpainting with regular subsampling on uniform grids. As explained in Section~\ref{sec:sparse:l1}, the regular subsampling operator $\bU$ may result in a highly coherent transformed dictionary $\bU \bD$. Calculating an accurate sparse estimation for interpolation zooming is therefore more difficult than that for inpainting with random masks. 

The experiments are performed on the gray-level images Lena, Peppers, Mandril, Cameraman, Boat, and Straws, and the color images Lena, Peppers, Kodak05 and Kokad20. The color images are treated in the same way as for inpainting. These high-resolution images are down-sampled by a factor $2 \times2$ without anti-aliasing filtering. The resulting low-resolution images are aliased, which corresponds to the reality of television images that are usually aliased, since this improves their visual perception. The low-resolution images are then zoomed by the algorithms under consideration. When the anti-aliasing blurring operator is included before subsampling, zooming can be casted as a deconvolution problem and will be addressed in Section~\ref{sec:deblurring}. 

The PLE interpolation zooming is compared with linear interpolators~\cite{blu2001moms, keys1981cubic, unser1999splines, munoz2001least} as well
as recent super-resolution algorithms ``NEDI'' (new edge directed
interpolation)~\cite{li2001ned}, ``DFDF'' (directional filtering and
data fusion)~\cite{zhang2006edge}, ``KR'' (kernel regression)~\cite{takeda2007kernel}, ``ECM'' (expectation conditional maximization)~\cite{fadili2009inpainting},
``Contourlet''~\cite{mueller2007iiu}, ``ASR'' (adaptive sparse reconstructions)~\cite{guleryuz2006nonlinear}, ``FOE'' (fields of experts)~\cite{roth2009fields}, ``SR'' (sparse representation)~\cite{yang2010SR}, ``SAI'' (soft-decision
adaptive Interpolation)~\cite{ZhangW08iia} and ``SME'' (sparse mixing estimators)~\cite{mallat10SME}.
KR, ECM, ASR and FOE are generic inpainting algorithms that have been described in Section~\ref{sec:inpainting}. NEDI, DFDF and SAI are adaptive directional
interpolation methods that take advantage of the image directional regularity. Contourlet is a sparse inverse
problem estimator as described in Section~\ref{sec:sparse:l1}, computed
in a contourlet frame. SR is also a sparse inverse estimator that learns the dictionaries from a training image set. 
SME is a recent zooming algorithm that exploits directional structured sparsity in wavelet representations. 
Among the previously published algorithms,
SAI and SME currently provide the best PSNR for spatial image
interpolation zooming~\cite{mallat10SME, ZhangW08iia}. The results of ASR are generated with our own implementation, and those of 
all the other algorithms are produced by the original authors' softwares, with the parameters manually optimized. As the anti-aliasing operator
is not included in the interpolation zooming model, to obtain correct results with SR, the anti-aliasing filter used in the original authors' SR software is deactivated in both dictionary training (with the authors' original training dataset of 92 images) and super-resolution estimation. PLE is configured in the same way as for inpainting as described in Section~\ref{sec:inpainting}, with patch size $8 \times 8$ for gray-level images, and $6 \times 6 \times 3$ for color images. 

Table~\ref{tab:zooming} gives the PSNRs generated by all algorithms on the gray-level
and the color images.
Bicubic interpolation provides
nearly the best results among all tested linear interpolators,
including cubic splines~\cite{unser1999splines}, MOMS~\cite{blu2001moms} and others
\cite{munoz2001least}, due to the aliasing produced by the down-sampling. PLE gives
moderately higher PSNRs than SME and SAI for all the images, with one exception 
where the SAI produces slightly higher PSNR. Their gain in PSNR is significantly larger than
with all the other algorithms. 

\begin{figure}[htbp]
\vspace{-2ex}
\begin{center}
\begin{tabular}{ccccc}
\hspace{-1ex}\epsfxsize=3.4cm\epsffile{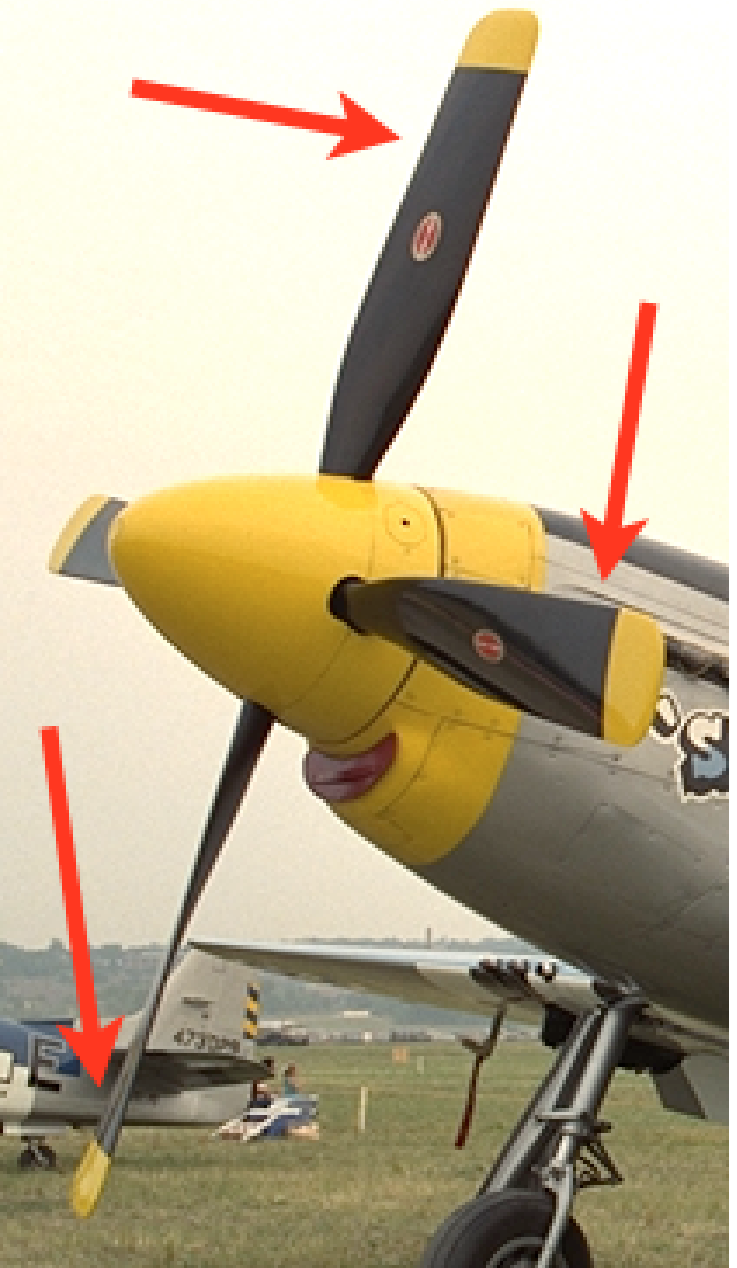} &
\hspace{-1ex}\epsfxsize=1.7cm \epsffile{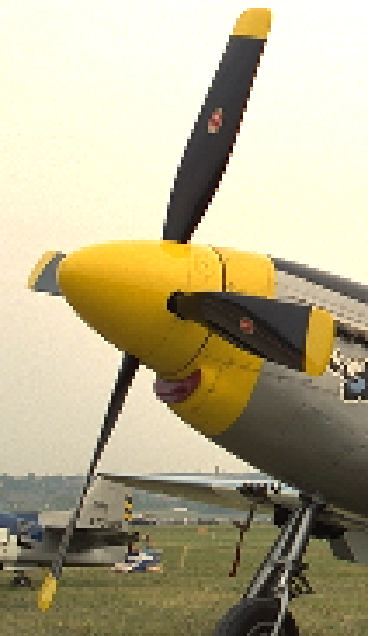} &
\hspace{-1ex}\epsfxsize=3.4cm \epsffile{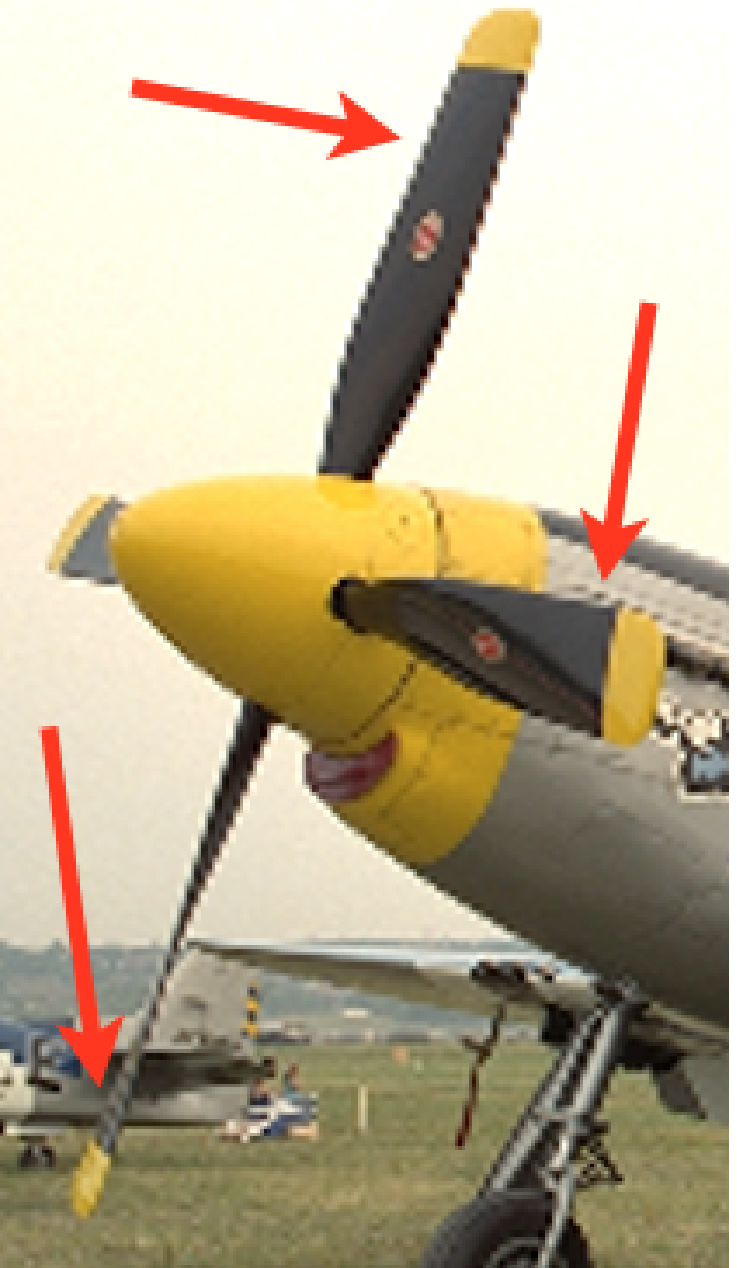} &
\hspace{-1ex}\epsfxsize=3.4cm \epsffile{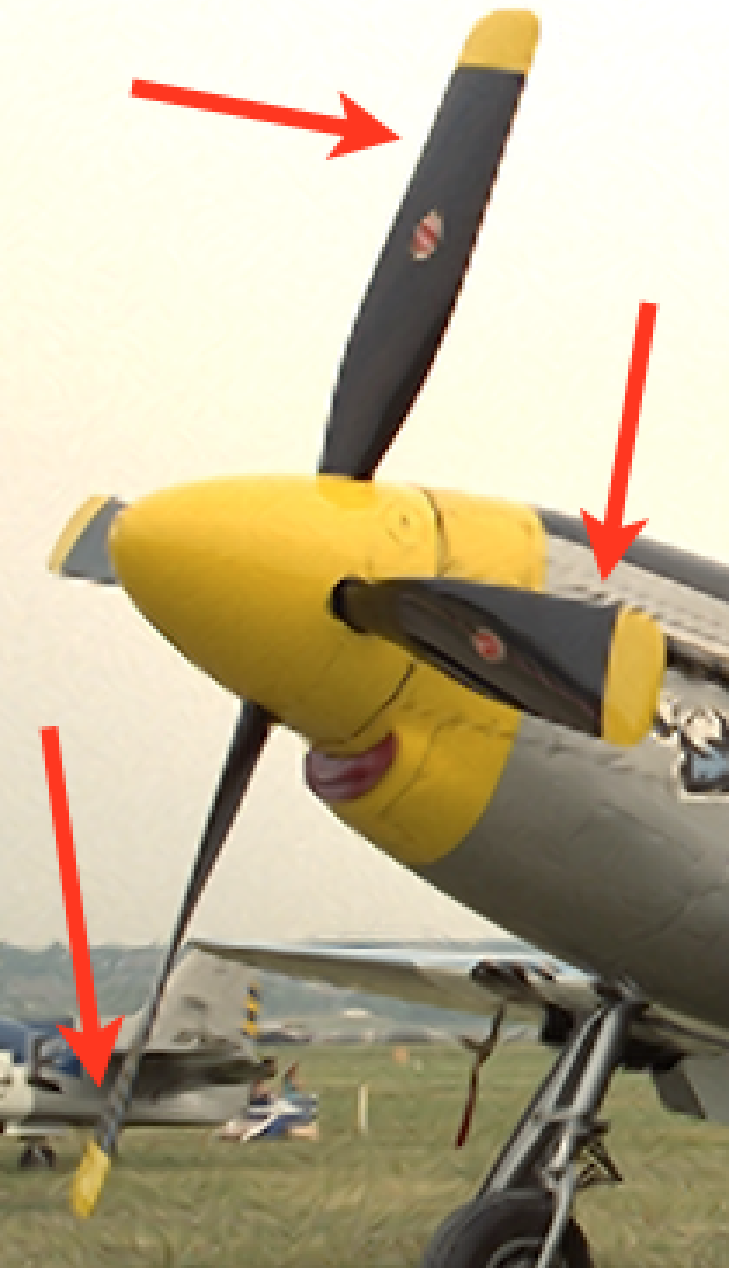} &
\hspace{-1ex}\epsfxsize=3.4cm \epsffile{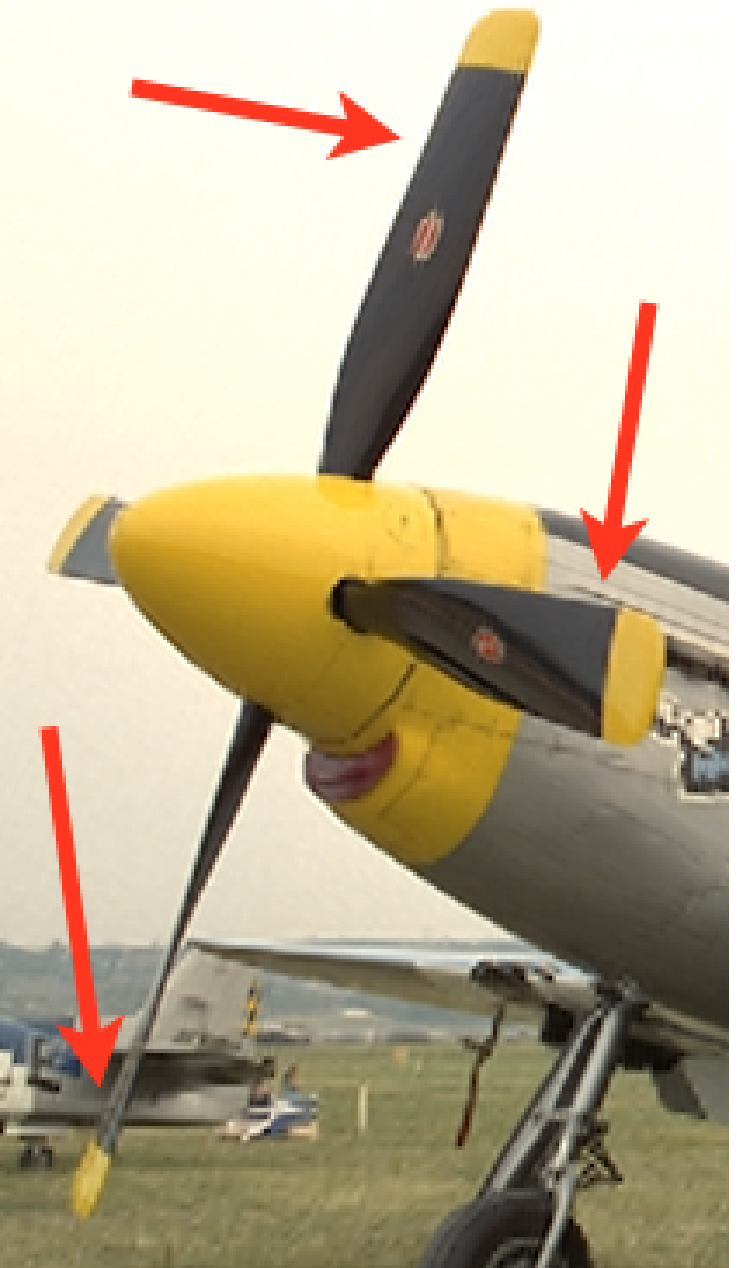} \\
\hspace{-2ex} \small{\textbf{(a) HR}} & 
\hspace{-2ex} \small{\textbf{(b) LR}} & 
\hspace{-2ex} \small{\textbf{(c) Bibubic}} & 
\hspace{-2ex} \small{\textbf{(d) SAI}} & 
\hspace{-2ex} \small{\textbf{(e) PLE}} 
\hspace{0ex} 
\end{tabular}
\end{center}
\vspace{-3ex}
\caption{\small Color image zooming. (a) Crop from the high-resolution image Kodak20. (b) Low-resolution image. From (c) to (e), images zoomed by bicubic interpolation (28.48 dB),  SAI (30.32 dB)~\cite{ZhangW08iia}, and proposed PLE framework (30.64 dB). PSNRs obtained by the other methods under consideration: NEDI (29.68 dB)~\cite{li2001ned}, DFDF (29.41 dB)~\cite{zhang2006edge}, KR (29.49 dB)~\cite{takeda2007kernel}, FOE (28.73 dB)~\cite{roth2009fields}, SR (23.85 dB)~\cite{yang2010SR}, and SME (29.90 dB)~\cite{mallat10SME}. Attention should be focused on the places indicated by the arrows.} 
\label{fig:zooming:kodak20}
\vspace{-6ex}
\end{figure}

Figure~\ref{fig:zooming:kodak20} compares an interpolated image
obtained by the baseline bicubic interpolation and the algorithms that generate the highest
PSNRs, SAI and PLE. The local PSNRs on the
cropped images produced by all the methods under consideration are reported as well. 
Bicubic interpolation produces
some blur and jaggy artifacts in the zoomed images. These
artifacts are reduced to some extent by the NEDI, 
DFDF, KR and FOE algorithms, but the image quality is still lower than
with PLE, SAI and SME algorithms, as also reflected in the PSNRs.
SR yields an image that looks sharp. 
However, due to the coherence of the transformed dictionary, as explained in Section~\ref{sec:sparse:l1}, when the approximating atoms are not correctly selected, it produces artifact patterns along the contours, which
degrade its PSNR. The PLE algorithm restores slightly
better  than SAI and SME on regular geometrical structures, as
can be observed on the upper and lower propellers, as well as on the fine lines on the side of the plane indicated by the arrows. 

\begin{table}[htbp]
\vspace{-2ex}
\begin{center}
{\small
\begin{tabular}{c}
\hspace{-5ex}
\begin{tabular}{|c||c|c|c|c|c|c|c|c|c|c|c|c|c|}
\hline  & Bicubic & NEDI & DFDF & KR & ECM & Contourlet & ASR & FOE* & SAI & SME & PLE
\\ \hline
Lena &  33.93  & 33.77 & 33.91 & 33.94 & 24.31 & 33.92 & 33.19 & 34.04 & 34.68 & 34.58 & \textbf{34.76}
\\ \hline
Peppers  & 32.83  & 33.00 & 33.18 & 33.15 & 23.60 & 33.10 & 32.33 & 31.90 & 33.52 & 33.52 & \textbf{33.62} \\
\hline
Mandril  & 22.92  & 23.16 & 22.83 & 22.93 & 20.34 & 22.53 & 22.66 & 22.99 & 23.19 & 23.16 & \textbf{23.27} \\
\hline
Cameraman  & 25.37 & 25.42 & 25.67  & 25.51 & 19.50 & 25.35 & 25.33 & 25.58 & 25.88 & 26.26 & \textbf{26.47} \\
\hline
Boat  & 29.24  & 29.19 & 29.32 & 29.18 & 22.20 & 29.25 & 28.96 & 29.36 & 29.68 & 29.76 & \textbf{29.93} \\
\hline
Straws  & 20.53  & 20.54 & 20.70 & 20.76 & 17.09 & 20.52 & 20.54& 20.47 & 21.48 & 21.61 & \textbf{21.82} \\
\hline \hline
\textit{Ave. gain}  & 0  & 0.04 & 0.13 & 0.11 & -6.30 & -0.02 & -0.30 & -0.08 & 0.60 & 0.68 & \textbf{0.84}\\
\hline
\end{tabular}
\vspace{1ex}\\
 \hspace{-5ex}
\begin{tabular}{|c||c|c|c|c|c|c|c|c|c|c|}
\hline  & Bicubic & NEDI & DFDF & KR &  FOE* & SR* & SAI & SME & PLE \\
\hline
Lena &  32.41 & 32.47 & 32.46 & 32.55 & 32.55 & 26.42 & 32.98 & 32.88 & \textbf{33.53} \\
\hline
Peppers & 30.95 & 31.06 & 31.24 & 31.26 & 31.05 & 26.43 & 31.37 & 31.35 & \textbf{31.88} \\
\hline
Kodak05 & 25.82  & 25.93 & 26.03 & 26.09 & 26.01 & 20.76 & \textbf{26.91} & 26.72 & 26.77 \\
\hline
Kodak20 &  30.65 & 31.06 & 31.08 & 30.97 & 30.84 & 25.92 & 31.51 & 31.38 & \textbf{31.72} \\
\hline
\hline
\textit{Ave. gain} &  0 & 0.17 & 0.25 & 0.27 & 0.16 & -5.07 & 0.74 & 0.63 & \textbf{1.02} \\
\hline
\end{tabular}
\end{tabular}
}
\end{center}
\vspace{-2ex}
\caption{\small PSNR comparison on gray-level (top) and color (bottom) image interpolation zooming. The algorithms under consideration are bicubic interpolation, NEDI~\cite{li2001ned}, DFDF~\cite{zhang2006edge}, KR~\cite{takeda2007kernel}, 
ECM~\cite{fadili2009inpainting}, Contourlet~\cite{mueller2007iiu}, ASR~\cite{guleryuz2006nonlinear}, FOE~\cite{roth2009fields}, SR~\cite{yang2010SR},
SAI~\cite{ZhangW08iia} , SME~\cite{mallat10SME} and the proposed PLE framework.
The bottom row shows the average gain of
each method relative to the bicubic interpolation. The highest PSNR in
each row is in boldface. The algorithms with * use a training dataset.} \label{tab:zooming}
\vspace{-5ex}
\end{table}

\section{Deblurring}
\label{sec:deblurring}

An image $\bbf$ is blurred and contaminated by additive noise, $\by = \bU \bbf + \bw$, where $\bU$ is a convolution operator and $\bw$ is the noise. Image deblurring aims at estimating $\bbf$ from the blurred and noisy observation $\by$. 

\subsection{Hierarchical PLE}
\begin{figure}[htbp]
\vspace{-2ex}
\begin{center}
\begin{tabular}{cc}
\epsfxsize=8cm \epsffile{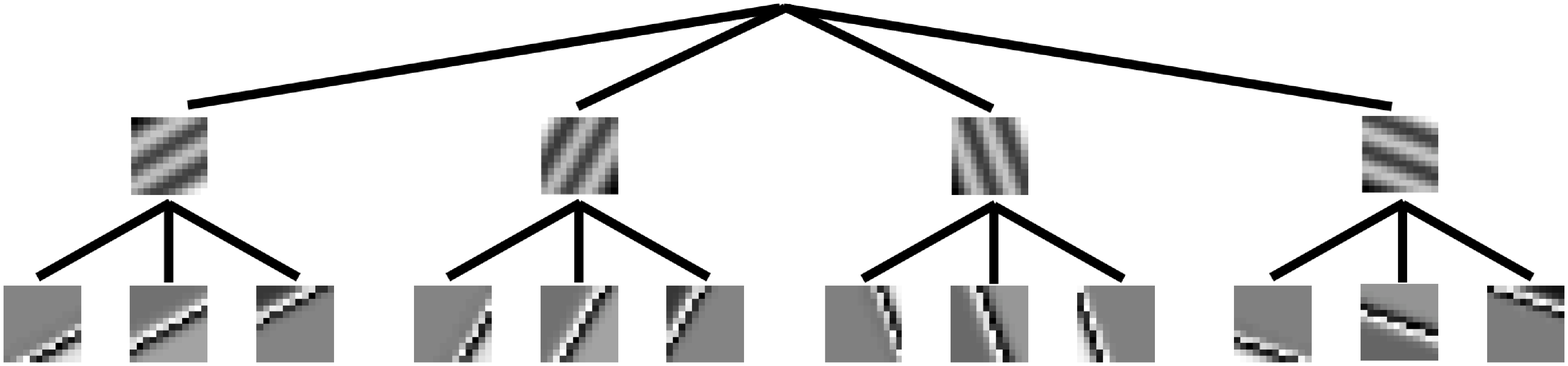}  &
\hspace{6ex} \epsfxsize=2cm \epsffile{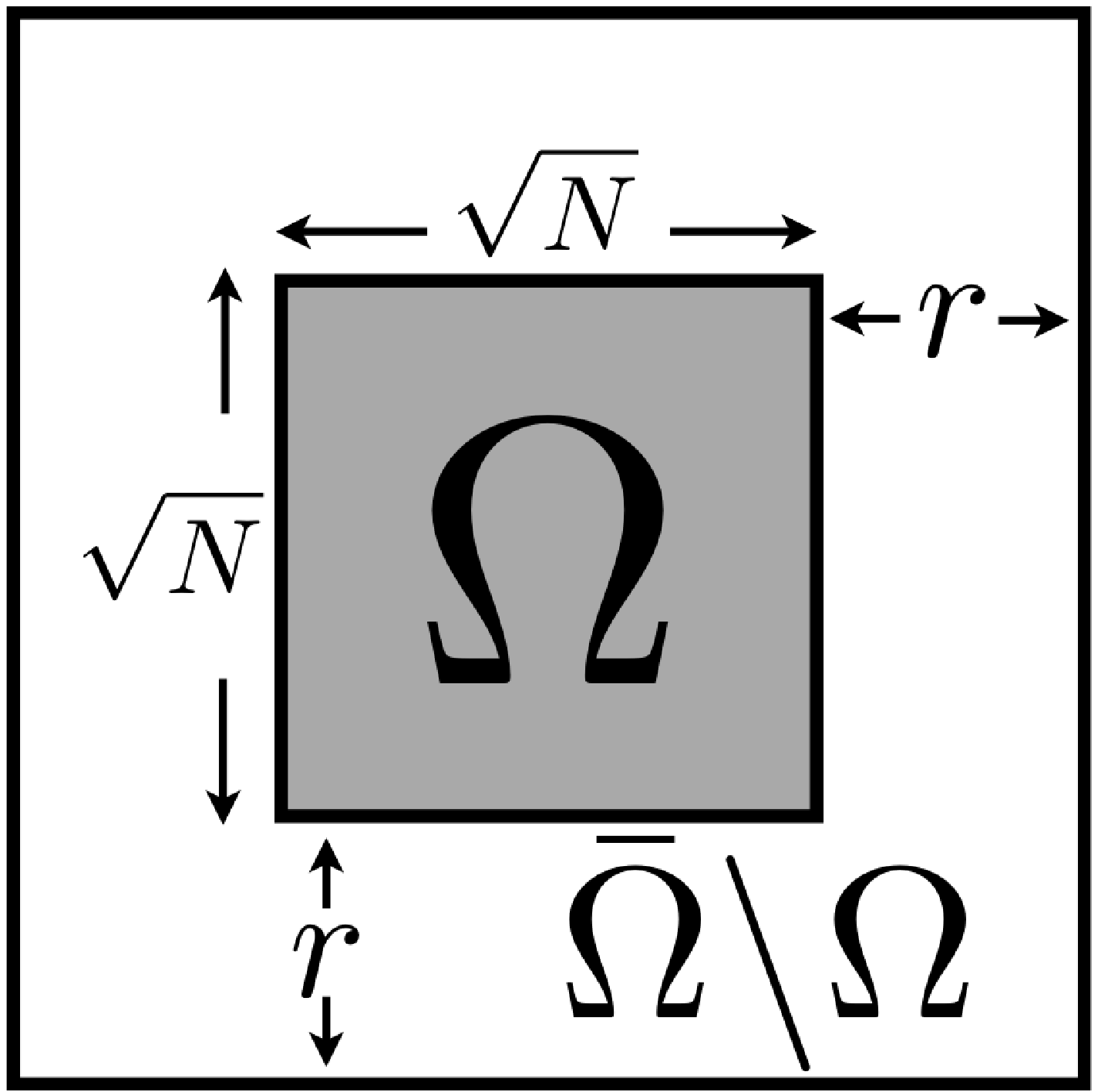}  \\
\small{\textbf{(a)}} & \hspace{6ex}\small{\textbf{(b)}}
\end{tabular}
\end{center}
\vspace{-3ex}
\caption{\small (a). Hierarchical PLE for deblurring. Each patch in the first layer symbolizes a directional PCA basis.  Each patch in the second layer symbolizes a position PCA basis.  (b) To circumvent boundary issues, deblurring a patch whose support is $\Omega$ can be casted as inverting an operator compounded by a masking and a convolution defined on a larger support $\bar \Omega$. See text for details.} 
\label{fig:hierarchy:deblur}
\vspace{-4ex}
\end{figure}

As explained in Section~\ref{subsubsec:reconverability}, the recoverability condition of sparse super-resolution estimates for deblurring requires a dictionary comprising atoms with spread Fourier spectrum and thus localized in space, such as the position PCA basis illustrated in Figure~\ref{fig:PCA:init:pos}. To reduce the computational complexity, model selection with a hierarchy of directional PCA bases and position PCA bases is proposed, in the same spirit of~\cite{yu2009sparse}.  Figure~\ref{fig:hierarchy:deblur}-(a) illustrates the hierarchical PLE with a cascade of the two layers of model selections. The first layer selects the direction, and given the direction, the second layer further specifies the position. 

In the first layer, the model selection procedure is identical to that in image inpainting and zooming, i.e., it is calculated with the Gaussian models corresponding to the directional PCA bases $\{\bB_k\}_{1 \leq k \leq K}$, Figure~\ref{fig:PCA:init}-(c). In this layer, a directional PCA $\bB_k$ of orientation $\theta$ is selected for each patch. Given the directional basis $\bB_k$ selected in the first layer, the second layer recalculates the model selection~\eqref{eqn:MAP:model:selection}, this time with a family of position PCA bases $\{\bB_{k,p}\}_{1 \leq p \leq P}$ corresponding to the same direction $\theta$ as the directional basis $\bB_k$ selected in the first layer, with atoms in each basis $\bB_{k,p}$ localized at one position, and the $P$ bases translating in space and covering the whole patch. The image patch estimation~\eqref{eqn:MAP:estimate:gaussian1:best} is obtained in the second layer. This hierarchical calculation reduces the computational complexity from $\mathcal{O}(KP)$ to $\mathcal{O}(K+P)$. 

For deblurring, boundary issues  on the patches need to be addressed. Since the convolution operator is non-diagonal, the deconvolution of each pixel $\by(x)$ in the blurred image $\by$ involves the pixels in a neighborhood around $x$ whose size depends on the blurring kernel. As the patch based methods deal with the local patches, for a given patch, the information outside of it is missing. Therefore, it is impossible to obtain accurate deconvolution estimation on the boundaries of the patches. To circumvent this boundary problem, a larger patch is considered and the deconvolution is casted as a deconvolution plus an inpainting problem. Let us retake the notations $\bbf_i$, $\by_i$ and $\bw_i$ to denote respectively the patches of size $\sqrt{N} \times \sqrt{N}$ in the original image $\bbf$, the degraded image $\by$, and the noise $\bw$. Let $\Omega$ be their support. Let $\bar{\bbf}_i$, $\bar{\by}_i$ and $\bar{\bw}_i$ be the corresponding larger patches of size $(\sqrt{N}+2r) \times (\sqrt{N}+2r)$, whose support $\bar \Omega$ is centered at the same position as $\Omega$ and with an extended boundary $\bar{\Omega} \backslash \Omega$ of width $r$ (the width of the blurring kernel, see below), as illustrated in Figure~\ref{fig:hierarchy:deblur}-(b). Let $\bar{\bU}$ be an extension of the convolution operator $\bU$ on $\bar{\Omega}$ such that $\bar{\bU} \bbf_i(x)= \bU \bbf_i (x)$ if $x \in \Omega$, and $0$ if $x \in \bar{\Omega} \backslash \Omega$. Let $\bM$ be a masking operator defined on $\bar{\Omega}$ which keeps all the pixels in the central part $\Omega$ and kills the rest, i.e., $\bM \bar{\bbf}_i (x) = \bbf_i (x)$ if $x \in \Omega$, and 0 if $x \in \bar{\Omega} \backslash \Omega$. If the width $r$ of the boundary $\bar{\Omega} \backslash \Omega$ is larger than the radius of the blurring kernel, one can show that the blurring operation can be rewritten locally as an extended convolution on the larger support followed by a masking, $\bM \bar{\by}_i = \bM \bar{\bU}   \bar{\bbf}_i + \bM \bar{\bw}_i$. Estimating $\bbf_i$ from $\by_i$ can thus be calculated by estimating $\bar{\bbf}_i$ from $\bM \bar{\by}_i$, following exactly the same algorithm, now treating the compounded $\bM \bar{\bU}$ as the degradation operator to be inverted. The boundary pixels in the estimate $\tilde{\bar{\bbf}}_i(x)$, $x \in \bar{\Omega} \backslash \Omega$, can be interpreted as an extrapolation from $\by_i$, therefore less reliable. The deblurring estimate $\tilde{\bbf}_i$ is obtained by discarding these boundary pixels from  $\tilde{\bar{\bbf}}_i$ (which are outside of $\Omega$ anyway).

Local patch based deconvolution algorithms become less accurate if the blurring kernel support is large relative to the patch size. In the deconvolution experiments reported below, $\Omega$ and $\bar{\Omega}$ are respectively set to $8 \times 8$ and $12 \times 12$. The blurring kernels are restricted to a $5 \times 5$ support. 

\subsection{Deblurring Experiments}
The deblurring experiments are performed on the gray-level images Lena, Barbara, Boat, House, and Cameraman, with different amounts of blur and noise. The PLE deblurring is compared with a number of deconvolution algorithms:  ``ForWaRD'' (Fourier-wavelet regularized deconvolution)~\cite{neelamani2004forward}, ``TVB'' (total variation based)~\cite{bioucas2006total}, ``TwIST'' (two-step iterative shrinkage/thresholding)~\cite{bioucas2007new}, ``SP'' (sparse prior)~\cite{levin2007deconvolution}, ``SA-DCT'' (shape adaptive DCT)~\cite{foi2006shape}, ``BM3D'' (3D transform-domain collaborative filtering)~\cite{dabov2008image}, and ``DSD'' (direction sparse deconvolution)~\cite{lou2009direct}. ForWaRD, SA-DCT and BM3D first calculate the deconvolution with a regularized Wiener filter in Fourier, and then denoise the Wiener estimate with, respectively, a thresholding estimator in wavelet and SA-DCT representations, and with the non-local 3D collaborative filtering~\cite{dabov2007image}. TVB and TwIST deconvolutions regularize the estimate with the image total variation prior. SP assumes a sparse prior on the image gradient.  DSD is a recently developed sparse inverse problem estimator, described in Section~\ref{sec:sparse:l1}. In the previous published works, BM3D and SA-DCT are among the deblurring methods that produce the highest PSNRs, followed by SP. The results of TVB, TwIST, SP, SA-DCT and DSD are generated by the authors' original softwares, with the parameters manually optimized, and those of ForWaRD are calculated with our own implementation.  The proposed algorithm runs for 5 iterations.

Table~\ref{tab:deblurring} gives the ISNRs (improvement in PSNR relative to the input image) of the different algorithms for restoring images blurred with Gaussian kernels of standard deviation $\sigma_b=1$ and $2$ (truncated to a $5 \times 5$ support), and contaminated by a white Gaussian noise of standard deviation $\sigma_n = 5$. BM3D produces the highest ISNRs, followed closely by SA-DCT and PLE, whose ISNRs are comparable and are moderately higher than with SP on average. Let us remark that BM3D and SA-DCT apply an empirical Wiener filtering as a post-processing that boosts the ISNR by near 1 dB. The empirical Wiener technique can be plugged into other sparse transform-based methods such as PLE and ForWaRD as well. Without this post-processing, PLE outperforms BM3D and SA-DCT on average.

\begin{figure}[htbp]
\vspace{-3ex}
\begin{center}
\begin{tabular}{cccc}
\hspace{-1ex}\epsfxsize=3.2cm \epsffile{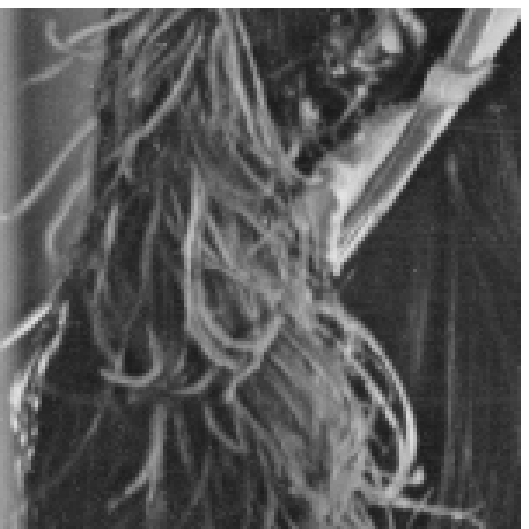}  &
\hspace{-1ex}\epsfxsize=3.2cm \epsffile{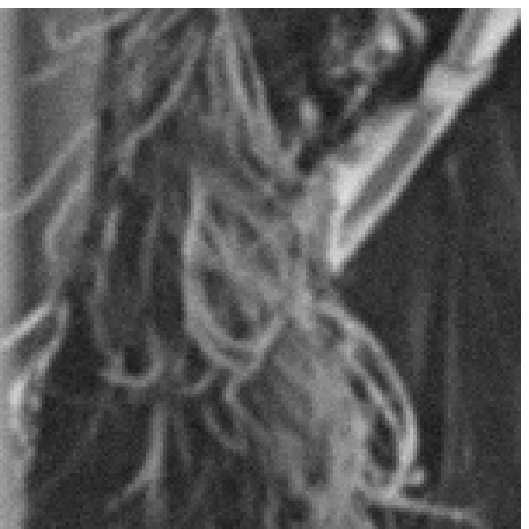}  &
%
\hspace{-1ex}\epsfxsize=3.2cm \epsffile{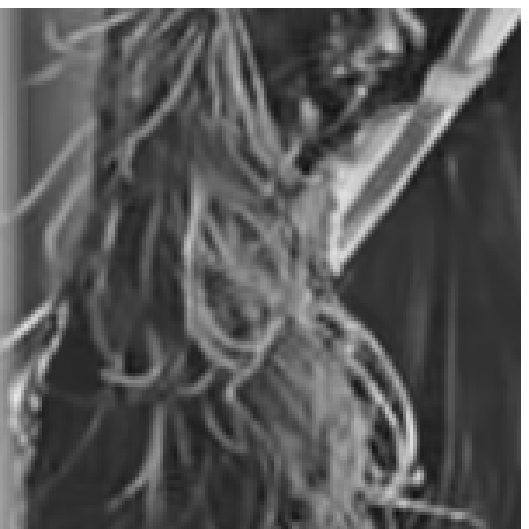}  &
\hspace{-1ex}\epsfxsize=3.2cm \epsffile{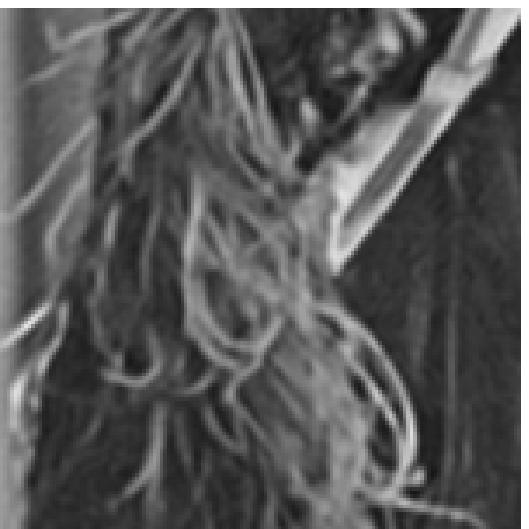}   \\
\hspace{-2ex} \small{\textbf{(a) Original}} & 
\hspace{-2ex} \small{\textbf{(b) Blurred and noisy}} & 
\hspace{-3ex} \small{\textbf{(c) BM3D}} & 
\hspace{-2ex} \small{\textbf{(D) PLE}}
\end{tabular}
\end{center}
\vspace{-3ex}
\caption{\small Gray-level image deblurring. (a) Crop from Lena. (b) Image blurred by a Gaussian kernel of standard deviation $\sigma_b=1$ and contaminated by white Gaussian noise of standard deviation $\sigma_n = 5$ (PSNR=27.10). 
(c) and (d). Images deblurred by BM3D with empirical Wiener post-processing (ISNR 3.40 dB dB)~\cite{dabov2008image}, and the proposed PLE framework (ISNR 2.94 dB). ISNR produced by the other methods under consideration: BM3D without empirical Wiener post-processing (2.65 dB)~\cite{dabov2008image}, TVB (2.72 dB)~\cite{bioucas2006total}, TwIST (2.61dB)~\cite{bioucas2007new}, SP (2.93 dB)~\cite{levin2007deconvolution}, SA-DCT with/without empirical Wiener post-processing (2.95/2.10 dB)~\cite{foi2006shape}, and DSD (1.95 dB)~\cite{lou2009direct}.} 
\label{fig:deblurring:Lena}
\vspace{-6ex}
\end{figure}

Figure~\ref{fig:deblurring:Lena} shows a deblurring example. All the algorithms under consideration reduce the amount of blur and attenuate the noise. BM3D generates the highest ISNR, followed by SA-DCT, PLE and SP, all producing similar visual quality, which are moderately better than the other methods. DSD accurately restores sharp image structures when the atoms are correctly selected, however, some artifacts due to the incorrect atom selection offset its gain in ISNR. The empirical Wiener filtering post-processing in BM3D and SA-DCT efficiently removes some artifacts and significantly improves the visual quality and the ISNR. More examples of PLE deblurring will be shown in the next section.

\begin{table}[htbp]
\vspace{-2ex}
\begin{center}
{\small
\begin{tabular}{|c|c|c||c|c|c|c|c|c|c|c|}
\hline
\multicolumn{3}{|c||}{\textit{\small Kernel size and input PSNR}}  & ForWaRD & TVB & TwIST & SA-DCT & BM3D & SP & DSD* & PLE \\
\hline
\multirow{2}{*} {Lena}  
& $\sigma_b=1$ & 30.62 & 2.51 & 3.03 & 2.87 & 3.56/2.58 & \textbf{4.03}/3.45 & 3.31 & 2.56 & \textit{3.77} \\ \cline{2-11}
& $\sigma_b=2$ & 28.84 & 2.33 & 3.15 & 3.13 & 3.46/3.00 & \textbf{3.91}/3.20 & 3.40 & 2.47 & \textit{3.52}  \\ \hline \hline
\multirow{2}{*} {House}  
& $\sigma_b=1$ & 30.04 & 2.31 & 3.12 & 3.23 & 4.14/3.07 & 4.29/3.80 & 3.52 & 2.27 & \textit{\textbf{4.38}}   \\ \cline{2-11}
& $\sigma_b=2$ & 28.02 & 2.29 & 3.24 & 3.82 & 4.21/3.64 & \textbf{4.73}/4.11 & \textit{3.92}  & 2.97 & 3.90  \\ \hline \hline
\multirow{2}{*} {Boat}  
& $\sigma_b=1$ & 28.29 & 1.69 & 2.45 & 2.44 & 2.93/2.21 & \textbf{3.23}/2.46 & 2.70 & 1.93 & \textit{2.72} \\ \cline{2-11}
& $\sigma_b=2$ & 26.21 & 1.63 & 2.67 & 2.59 & 3.71/\textit{2.63} & \textbf{3.33}/2.44 & 2.60 & 2.02 & 2.48 \\ \hline \hline 
\multirow{2}{*} {\textit{Average}}  
& $\sigma_b=1$ & 29.65 & 2.17 & 2.87 & 2.84 & 3.54/2.62 & \textbf{3.85}/3.23 & 3.17 & 2.25 & \textit{3.62} \\ \cline{2-11}
& $\sigma_b=2$ & 27.69 & 2.08 & 3.02 & 3.18 & 3.79/3.09 & \textbf{3.99}/3.25 & 3.30 & 2.48 & \textit{3.31} \\ \hline 
\end{tabular}
}
\end{center}
\vspace{-2ex}
\caption{\small ISNR (improvement in PSNR with respect to input image) comparison on image deblurring. Images are blurred by a Gaussian kernel of standard deviation $\sigma_b=1$ and $2$, and are then contaminated by white Gaussian noise of standard deviation $\sigma_n = 5$. From left to right: ForWaRD~\cite{neelamani2004forward}, TVB~\cite{bioucas2006total}, TwIST~\cite{bioucas2007new},  SA-DCT (with/without empirical Wiener post-processing)~\cite{foi2006shape}, BM3D (with/without empirical Wiener post-processing)~\cite{dabov2008image}, SP~\cite{levin2007deconvolution}, DSD~\cite{lou2009direct}, and the proposed PLE framework. The bottom box shows the average ISNRs given by each method over all the images with different amounts of blur. The highest ISNR in each row is in boldface, while the highest without post-processing is in italic. The algorithms with * use a training dataset.}  \label{tab:deblurring}
\vspace{-6ex}
\end{table}

\subsection{Zooming deblurring}
When an anti-aliasing filtering is taken into account, image zooming-out can be formulated as $\by = \bS \bU \bbf$,
where $\bbf$ is the high-resolution image, $\bU$ and $\bS$  are respectively an anti-aliasing convolution and a subsampling operator, and $\by$ is the resulting low-resolution image. Image zooming aims at estimating $\bbf$ from $\by$, which amounts to inverting the combination of the two operators $\bS$ and $\bU$. 

Image zooming can be calculated differently under different amounts of blur introduced by $\bU$. Let us distinguish between three cases:  (i) If the anti-aliasing filtering $\bU$ removes enough high-frequencies from $\bbf$ so that $\by = \bS \bU \bbf$ is free of aliasing, then the subsampling operator $\bS$ can be perfectly inverted with a linear interpolation denoted as $\bI$, i.e., $\bI \bS = Id$~\cite{mallat2008wts}. In this case, zooming can can be calculated as a deconvolution problem on $\bI \by = \bU \bbf$, where one seeks to invert the convolution operator $\bU$. In reality, however, camera and television images contain, always a certain amount of aliasing, since it improves the visual perception, i.e., the anti-aliasing filtering $\bU$ does not eliminate all the high-frequencies from $\bbf$. (ii) When $\bU$ removes a small amount of high-frequencies, which is often the case in reality, zooming can be casted as an interpolation problem~\cite{li2001ned, mallat10SME, mueller2007iiu, takeda2007kernel, zhang2006edge, ZhangW08iia}, where one seeks to invert only $\bS$, as addressed in Section~\ref{sec:zooming}. (iii) When $\bU$ removes an intermediate amount of blur from $\bbf$, the optimal zooming solution is inverting $\bS \bU$ together as a compounded operator, as investigated in~\cite{yang2010SR}. 

This section introduces a possible solution for the case (iii) with the PLE deblurring. A linear interpolation $\bI$ is first applied to partially invert the subsampling operator $\bS$. Due to the aliasing, the linear interpolation does not perfectly restore $\bU \bbf$, nevertheless it remains rather accurate, i.e., the interpolated image $\bI \by = \bI \bS \bU \bbf$ is close to the blurred image $\bU \bbf$, as $\bU \bbf$ has limited high-frequencies in the case (iii). The PLE deblurring framework is then applied to deconvolve $\bU$ from $\bI \by$. Inverting the operator $\bU$ is simpler than inverting the compounded operator $\bS \bU$.  As the linear interpolation $\bI$ in the first step is accurate enough in the case (iii), deconvolving $\bI \by$ results in accurate zooming estimates. 

In the experiments below, the anti-aliasing filter $\bU$ is set as a Gaussian convolution of standard deviation $\sigma_G=1.0$ and $\bS$ is an $s \times s = 2 \times 2$ subsampling operator. It has been shown that a pre-filtering with a Gaussian kernel of $\sigma_G=0.8 s$ guarantees that the following $s \times s$ subsampling generates a quasi aliasing-free image~\cite{morel2008consistency}. For a $2 \times 2$ subsampling, the anti-aliasing filtering $\bU$ with $\sigma_G=1.0$ leads to an amount of aliasing and visual quality similar to that in common camera pictures in reality.

\begin{figure}[htbp]
\vspace{-2ex}
\begin{center}
\begin{tabular}{cccccc}
\hspace{-3ex}\epsfxsize=3cm\epsffile{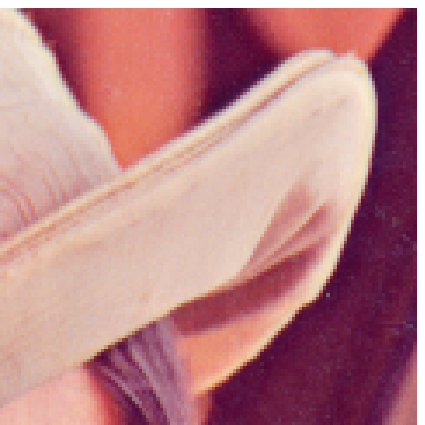}  &
\hspace{-1.5ex}\epsfxsize=3cm\epsffile{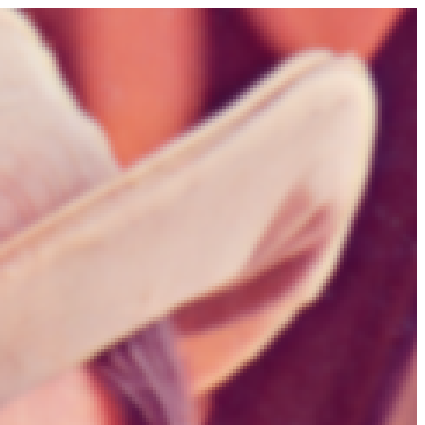}  &
\hspace{-1.5ex}\epsfxsize=1.5cm\epsffile{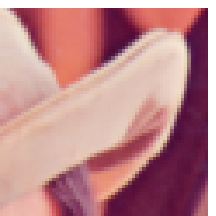}   &
\hspace{-1.5ex}\epsfxsize=3cm \epsffile{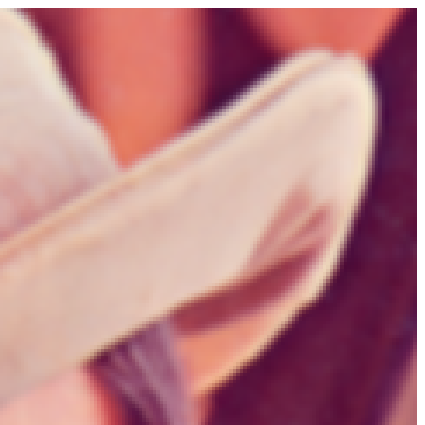}  &
\hspace{-1.5ex}\epsfxsize=3cm \epsffile{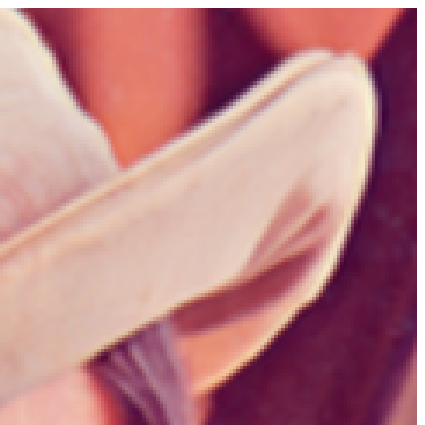}  &
\hspace{-1.5ex}\epsfxsize=3cm\epsffile{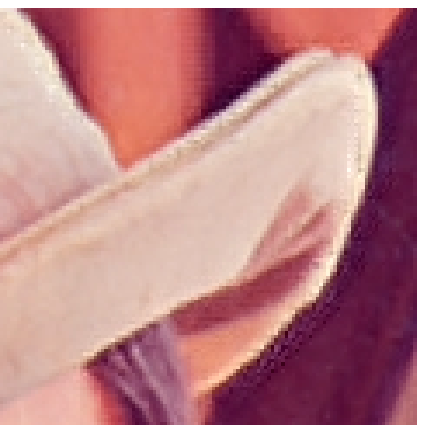}   \\
\hspace{-3ex} \small{\textbf{(a) $\bbf$}} & \hspace{-1.5ex}\small{\textbf{(b) $\bU \bbf$}} & \hspace{-1.5ex}\scriptsize{\textbf{(c) $\by = \bS \bU \bbf$}} & \hspace{-1.5ex}\small{\textbf{(d) $\bI \by$}} & \hspace{-1.5ex}\small{\textbf{(e) PLE}} & \hspace{-1.5ex}\small{\textbf{(f) SR}} 
\end{tabular}
\end{center}
\vspace{-3ex}
\caption{\small Color image zooming deblurring. (a) Crop from Lena: $\bbf$. (b) Image pre-filtered with a Gaussian kernel of standard deviation $\sigma_G=1.0$:  $\bU \bbf$. (c) Image subsampled from $\bU \bbf$ by a factor of $2 \times 2$: $\by = \bS \bU \bbf$. (d) Image interpolated from $\by$ with a cubic spline interpolation: $\bI \by$ (31.03 dB). (d) Image deblurred from $\bI \by$ by the proposed PLE framework (34.27 dB). (e) Image zoomed from $\by$ with~\cite{yang2010SR} (29.66 dB). The PSNRs are calculated on the cropped image between the original $\bbf$ and the one under evaluation.} 
\label{fig:zooming:deblurring:Lena}
\vspace{-4ex}
\end{figure}

Figure~\ref{fig:zooming:deblurring:Lena} illustrates an experiment on the image Lena. Figures~\ref{fig:zooming:deblurring:Lena}-(a) to (c) show, respectively, a crop of the original image $\bbf$, the pre-filtered version $\bU \bbf$, and the low-resolution image after subsampling $\by = \bS \bU \bbf$. As the amount of aliasing is limited in $\by$ thanks to the anti-aliasing filtering, a cubic spline interpolation is more accurate than lower ordered interpolations such as bicubic~\cite{unser1999splines}, and is therefore applied to upsample $\by$, the resulting image $\bI \by$ illustrated in Figure~\ref{fig:zooming:deblurring:Lena}-(d). A visual inspection confirms that $\bI \by$ is very close to $\bU \bbf$, the PSNR between them being as high as 50.02 dB. The PLE deblurring is then applied to calculate the final zooming estimate $\tilde{\bbf}$ by deconvolving $\bU$ from $\bI \by$. (As no noise is added after the anti-aliasing filter, the noise standard deviation is set to a small value $\sigma=1$.) As illustrated in Figure~\ref{fig:zooming:deblurring:Lena}-(e), the resulting image $\bbf$ is much sharper, without noticeable artifacts, and improves by 3.12 dB with respect to $\bI \by$. Figure~\ref{fig:zooming:deblurring:Lena}-(f) shows the result obtained with ``SR'' (sparse representation)~\cite{yang2010SR}. SR implements a sparse inverse problem estimator that tries to invert the compounded operator $\bS \bU$, with a dictionary learned from a natural image dataset. The experiments were performed with the authors' original software and training image set. The dictionaries were retrained with the $\bU \bS$ described above. It can be observed that the resulting image looks sharper and the restoration is accurate when the atoms selection is correct. However, due to the coherence of the dictionaries as explained in Section~\ref{sec:sparse:l1}, some noticeable artifacts along the edges are produced when the atoms are incorrectly selected, which also offset its gain in PSNR. 

\begin{figure}[htbp]
\vspace{-2ex}
\begin{center}
\begin{tabular}{ccccc}
\hspace{-2.5ex} \epsfxsize=3.6cm\epsffile{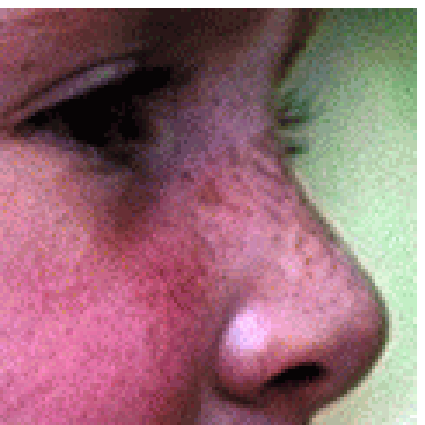}  &
\hspace{-1.5ex}\epsfxsize=1.8cm\epsffile{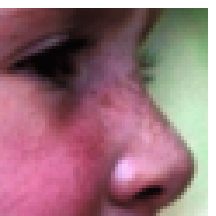}   &
\hspace{-1.5ex}\epsfxsize=3.6cm \epsffile{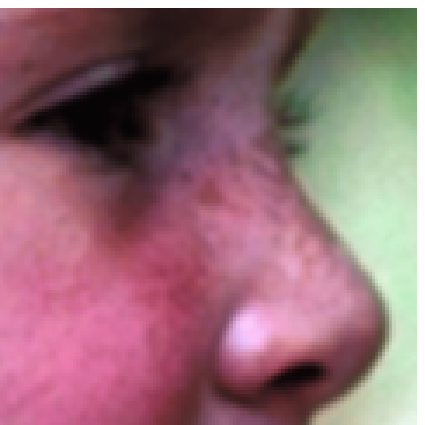}  &
\hspace{-1.5ex}\epsfxsize=3.6cm \epsffile{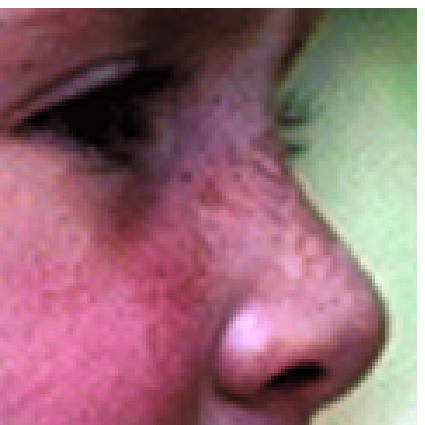}  &
\hspace{-1.5ex}\epsfxsize=3.6cm \epsffile{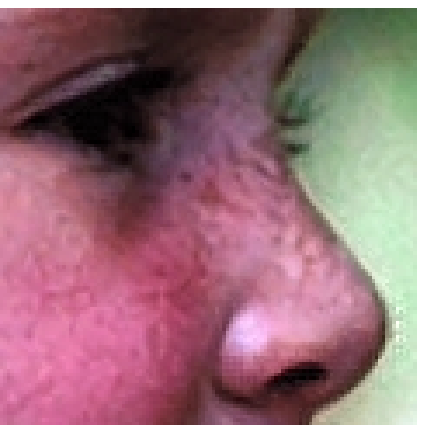}   \\
\hspace{-2.5ex} \small{\textbf{(a) HR}} & 
\hspace{-3ex} \small{\textbf{(b) LR} }\vspace{1ex}&
\hspace{-2.5ex} \small{\textbf{(c) Cubic spline}} & 
\hspace{-2.5ex} \small{\textbf{(c) PLE}} & 
\hspace{-3ex} \small{\textbf{(d) SR}} 
\end{tabular}
\end{center}
\vspace{-3ex}
\caption{\small Color image zooming deblurring. (a) Crop from Girl: $\bbf$. (b) Image pre-filtered with a Gaussian kernel of standard deviation $\sigma_G=1.0$, and subsampled by a factor of $2 \times 2$: $\by = \bS \bU \bbf$. (c) Image interpolated from $\by$ with a cubic spline interpolation: $\bI \by$ (29.40 dB). (d) Image deblurred from $\bI \by$ by the proposed PLE framework (30.49 dB). (e) Image zoomed from $\by$ with~\cite{yang2010SR} (28.93 dB).} 
\label{fig:zooming:deblurring:Girl}
\vspace{-2ex}
\end{figure}

Figure~\ref{fig:zooming:deblurring:Girl} shows another set of experiments on the image Girl. Again PLE efficiently reduces the blur from the interpolated image and leads to a sharp zoomed image without noticeable artifacts. SR produces similarly good visual quality as PLE, however, some slight but noticeable artifacts (near the end of the nose for example) due to the incorrect atom selection offset its gain in PSNR. 

Table~\ref{tab:zooming:deblurring} gives the PSNRs comparison on the color image images Lena, Girl and Flower. PLE deblurring from the cubic spline interpolation improves from 1 to 2 dB PSNR over the interpolated images. Although SR is able to restore sharp images, its gain in PSNR is offset by the noticeable artifacts.

\begin{table}[htbp]
\vspace{-0ex}
\begin{center}
{\small
\begin{tabular}{|c||c|c|c|}
\hline
& Cubic spline & SR* & PLE \\
\hline 
Lena & 31.60  & 30.64 & \textbf{33.78}  \\
\hline 
Girl &  30.62  & 30.43 &  \textbf{31.82}\\
\hline 
Flower &  37.02 &  35.96 & \textbf{39.06}  \\
\hline 
\end{tabular}
}
\end{center}
\vspace{-2ex}
\caption{\small PSNR comparison in image zooming. The high-resolution images are blurred and subsampled to generate the low-resolution images. The first column shows the PSNR produced by cubic spline interpolation. PLE deblurring is applied over the interpolated images and the resulting PSNRs are shown in the third column. The second column gives the PSNR obtained with SR~\cite{yang2010SR}. The highest PSNR in each row is in boldface. The algorithms with * use a training dataset.} \label{tab:zooming:deblurring}
\vspace{-8ex}
\end{table}

\section{Conclusion and future works}
\label{sec:conclusion}
This work has shown that Gaussian mixture models (GMM) and a MAP-EM algorithm provide general and computational efficient solutions for inverse problems, leading to results in the same ballpark as state-of-the-art ones in various image inverse problems. A dual mathematical interpretation of the framework with structured sparse estimation is described, which shows that the resulting piecewise linear estimate stabilizes and improves the traditional sparse inverse problem approach. This connection also suggests an effective dictionary motivated initialization for the MAP-EM algorithm. In a number of image restoration applications, including inpainting, zooming, and deblurring, the same simple and computationally efficient algorithm (its core,~\eqref{eqn:MAP:estimate:gaussian},~\eqref{eqn:MAP:estimate:gaussian:solution},~\eqref{eqn:MAP:model:selection} and~\eqref{eqn:MAP:estimate:gaussian1:best}, can be written in 4 lines Matlab code) produce either equal, often significantly better, or very small margin worse results than the best published ones, with a computational complexity typically one or two magnitude smaller than $l_1$ sparse estimations. Similar results (on average just 0.1 dB lower than BM3D~\cite{dabov2007image}) are obtained for the simpler problem of denoising ($\bU$ the identity matrix). 

As described in Section~\ref{sec:SSMS}, the proposed algorithm is calculated with classic statistical tools of MAP-EM clustering and empirical covariance estimation. As a possible future work, its performance may be further improved with more sophisticated statistical instruments, for example, the stochastic EM algorithms~\cite{celeux1992classification} and more advanced covariance regularization methods~\cite{schafer2005shrinkage}, at a cost of higher computational complexity. 

{\noindent \small \textbf{Acknowledgements:} Work supported by NSF, NGA, ONR, ARO and NSSEFF. We thank St\'ephanie Allassonni\`ere for helpful discussions, in particular about MAP-EM and covariance regularization.}

\vspace{-2ex}

\bibliographystyle{plain}
\bibliography{biblio_SSMS_IEEE}

\end{document}